\pdfoutput=1

\documentclass[11pt]{article}

\usepackage[preprint]{latex/acl}

\usepackage{times}
\usepackage{latexsym}

\usepackage[T1]{fontenc}

\usepackage[utf8]{inputenc}

\usepackage{microtype}

\usepackage{inconsolata}

\usepackage{graphicx}

\usepackage{amsmath,amsfonts,bm}

\def\eqref#1{equation~\ref{#1}}

\def\1{\bm{1}}

\def\vb{{\bm{b}}}

\DeclareMathAlphabet{\mathsfit}{\encodingdefault}{\sfdefault}{m}{sl}
\SetMathAlphabet{\mathsfit}{bold}{\encodingdefault}{\sfdefault}{bx}{n}

\usepackage{cite}
\usepackage{amsmath,amssymb,amsfonts}
\usepackage{algorithmic}
\usepackage{textcomp}
\usepackage{xcolor}
\usepackage{makecell}
\usepackage{natbib}
\usepackage{caption}
\usepackage{multicol}
\usepackage{multirow}
\usepackage{arydshln}
\usepackage{physics}

\newcommand{\eg}{{e.g.}}
\newcommand{\krnorm}[1]{\left\lVert#1\right\rVert}

\newcommand{\lp}{\texttt{{CLP}}}
\newcommand{\wb}{\texttt{{WiCLP}}}

\title{Improving Compositional Attribute Binding in Text-to-Image Generative Models via Enhanced Text Embeddings}

\author{%
  \makecell{Arman Zarei\thanks{Equal contribution. Correspondence
to: azarei@umd.edu, krezaei@umd.edu}, Keivan Rezaei\footnotemark[1], Samyadeep Basu, Mehrdad Saberi, \\ Mazda Moayeri, Priyatham Kattakinda, Soheil Feizi} \\[10px]
  Department of Computer Science\\ University of University of Maryland
}

\begin{document}
\maketitle
\begin{abstract}
Text-to-image diffusion-based generative models have the stunning ability to generate photo-realistic images and achieve state-of-the-art low FID scores on challenging image generation benchmarks. However, one of the primary failure modes of these text-to-image generative models is in composing attributes, objects, and their associated relationships accurately into an image. In our paper, we investigate compositional attribute binding failures, where the model fails to correctly associate descriptive attributes (such as color, shape, or texture) with the corresponding objects in the generated images, and highlight that imperfect text conditioning with CLIP text-encoder is one of the primary reasons behind the inability of these models to generate high-fidelity compositional scenes. In particular, we show that (i) there exists an optimal text-embedding space that can generate highly coherent compositional scenes showing that the output space of the CLIP text-encoder is sub-optimal, and (ii) the final token embeddings in CLIP are erroneous as they often include attention contributions from unrelated tokens in compositional prompts.  Our main finding shows that significant compositional improvements can be achieved (without harming the model's FID score) by fine-tuning {\it only} a simple and parameter-efficient linear projection on CLIP's representation space in Stable-Diffusion variants using a small set of compositional image-text pairs. 
\end{abstract}

\newcommand{\earlystop}{\textsc{Switch-Off}}

\begin{figure*}
\centering
\includegraphics[width=\textwidth]{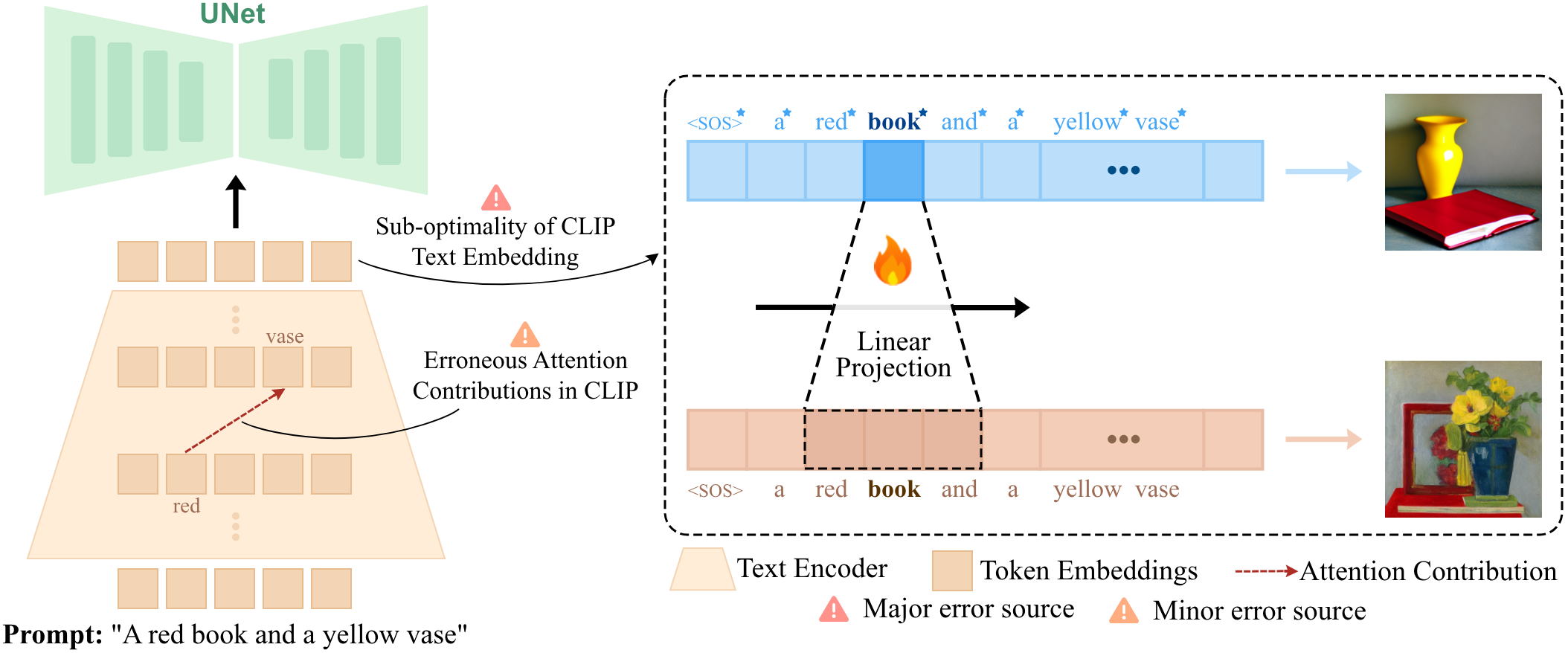}
\vspace{-17pt}
\caption{Overview of our analysis and proposed methods. The figure identifies two sources of errors in Stable Diffusion's inability to generate compositional prompts: (i) erroneous attention contribution in CLIP (minor) and (ii) sub-optimal CLIP text embedding (major). We propose a window-based linear projection (\wb{}), applying linear projection to a token’s surrounding window to enhance embeddings.
}

\label{fig:main_fig}
\end{figure*}

\section{Introduction}
Text-to-image diffusion-based generative models~\citep{DBLP:journals/corr/abs-2112-10752, podell2023sdxl, dalle, saharia2022photorealistic} have achieved photo-realistic image generation capabilities on user-defined text prompts. 
However, recent studies~\citep{huang2023t2icompbench} reveal that text-to-image models struggle with maintaining high fidelity when handling simple compositional prompts, such as those consisting of attributes, objects, and their associated relations (\eg, ``{\it a red book and a yellow vase}'').  This hinders the use of these generative models in various creative scenarios where the end-user wants to generate scenes that accurately reflect the composition and relationships specified in the prompt.

Existing approaches~\citep{chefer2023attendandexcite, feng2023layoutgpt, agarwal2023astar, wang2023compositional} explore various strategies to enhance compositionality in text-to-image models.
These methods primarily focus on modifying cross-attention maps by utilizing bounding box annotations and performing optimizations in the latent space during inference. Recent advancements, such as fine-tuning the UNet~\citep{huang2023t2icompbench}, have also demonstrated improvements in compositionality.
However, the {\it core reasons} behind compositionality failures remain poorly understood. Gaining insights into these root causes is crucial for developing more effective approaches to augment these models with enhanced compositional capabilities.

In our paper, we investigate the potential causes of compositional attribute binding failures in text-to-image generative models, where the model fails to correctly associate descriptive attributes (such as color, shape, or texture) with the corresponding objects in the generated images. We identify two key sources of error:
(i) {\it Erroneous attention contributions in CLIP output token embeddings}: We observe that output token embeddings in CLIP have significant attention contributions from irrelevant tokens, thereby introducing errors in generation. To explore this, we compare the internal attention contributions in CLIP for compositional prompts with the T5 text encoder, known for its stronger compositionality. Quantitative analysis shows that T5 exhibits fewer erroneous attention contributions than CLIP, indicating a potential reason for its superior compositionality.
(ii) {\it Sub-optimality of CLIP output space for compositional prompts}: We find out that there exists an alternative text-embedding space capable of generating highly coherent images from compositional prompts. This indicates that the current CLIP output space is inherently sub-optimal. 
Specifically, optimizing CLIP's text embeddings, while keeping the Stable Diffusion UNet frozen, converges to a more effective embedding space, enabling better compositional image generation.
These findings highlight that refining the output space of the CLIP text encoder could play a critical role in enhancing compositionality.



Building on our observations about the deficiencies of CLIP and identifying its text-embedding space as \textit{a core issue} in compositional attribute binding, we explore augmenting diffusion models with a lightweight module to enhance the text-encoder's output and improve compositionality. Remarkably, a simple linear projection achieves significant improvements, comparable or superior to full fine-tuning of CLIP or training more complex networks on top of it. We demonstrate that this linear projection effectively aligns the CLIP text-encoder’s output with a more optimal embedding space (see Figure~\ref{fig:main_fig}), leading to significantly stronger compositional performances.

\begin{figure}[]
\centering
\setlength{\tabcolsep}{1pt}
\begin{tabular}{cccccccc}
                  \multicolumn{2}{c}{SD v1.4} & \multicolumn{2}{c}{\wb} & \multicolumn{2}{c}{SD v2} & \multicolumn{2}{c}{\wb} \vspace*{1pt} \\
\multicolumn{2}{c}{\includegraphics[width=1.8cm]{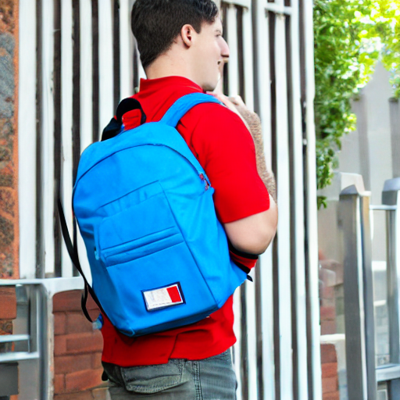}} & 
\multicolumn{2}{c}{\includegraphics[width=1.8cm]{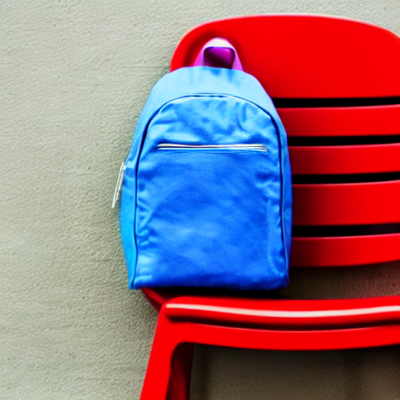}} & 
\multicolumn{2}{c}{\includegraphics[width=1.8cm]{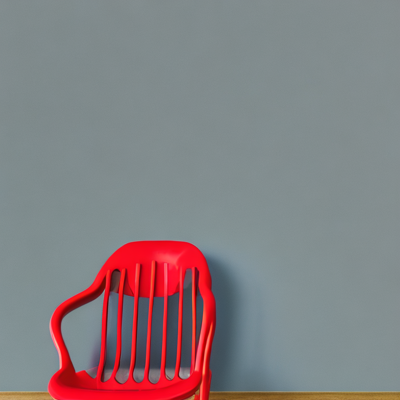}} & 
\multicolumn{2}{c}{\includegraphics[width=1.8cm]{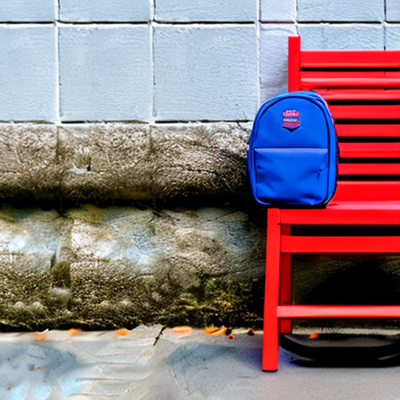}} \vspace*{-3.4pt} \\
\includegraphics[width=0.89cm]{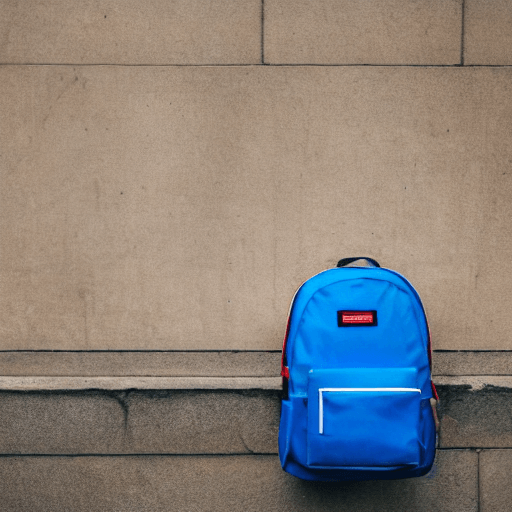} & \hspace*{-4pt}
\includegraphics[width=0.89cm]{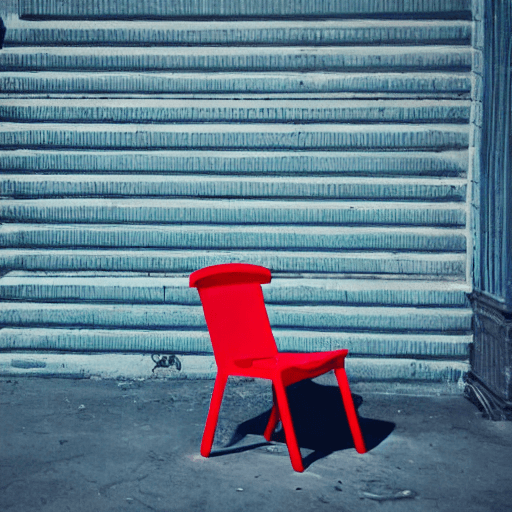} & \includegraphics[width=0.89cm]{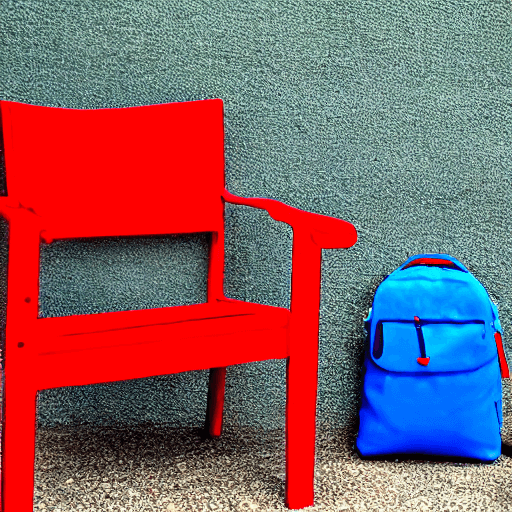} & \hspace*{-4pt} \includegraphics[width=0.89cm]{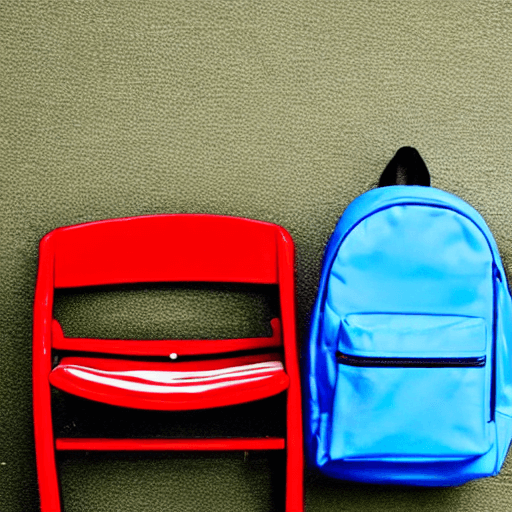} & \includegraphics[width=0.89cm]{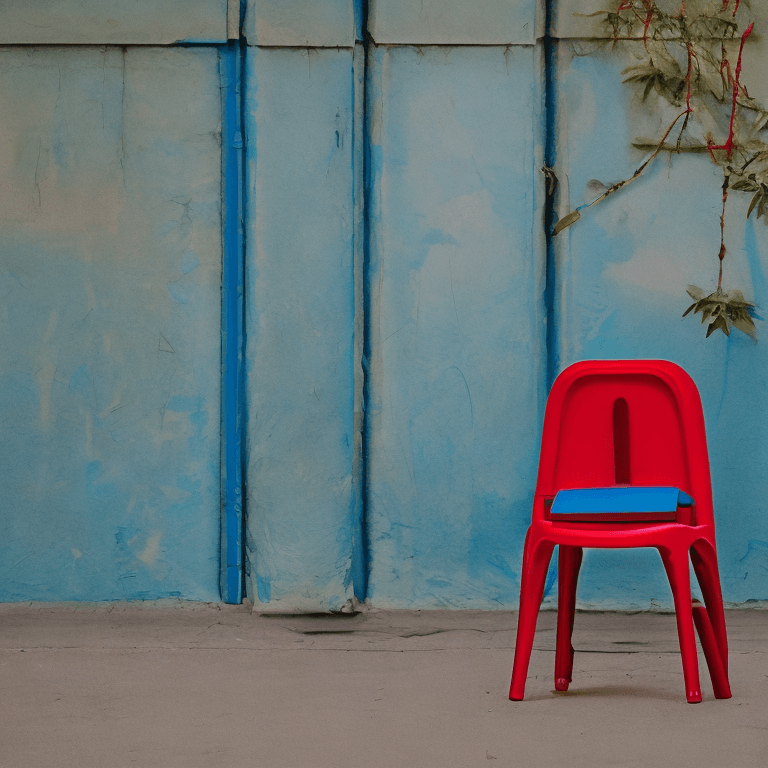} & \hspace*{-4pt} \includegraphics[width=0.89cm]{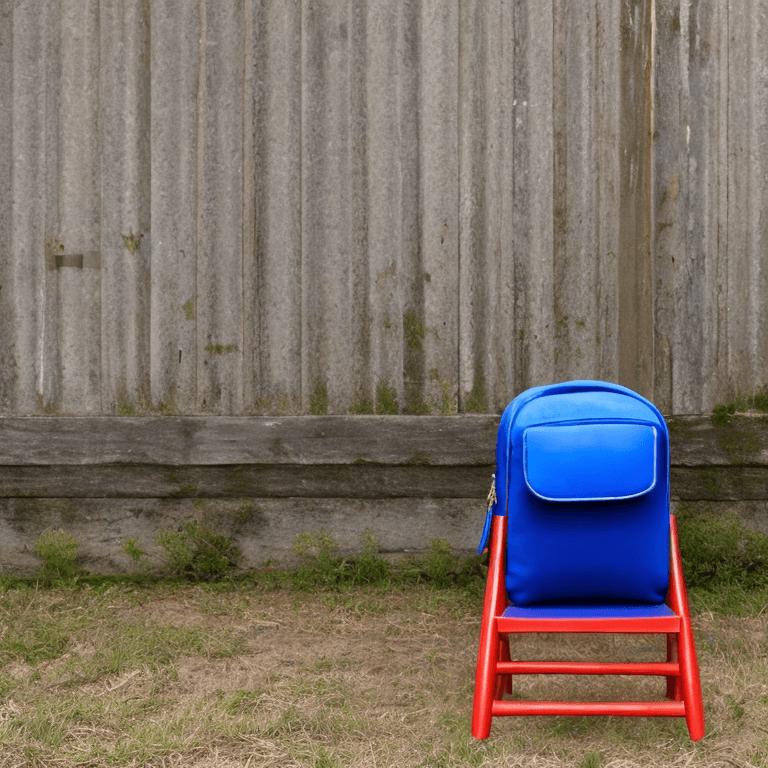} & \includegraphics[width=0.89cm]{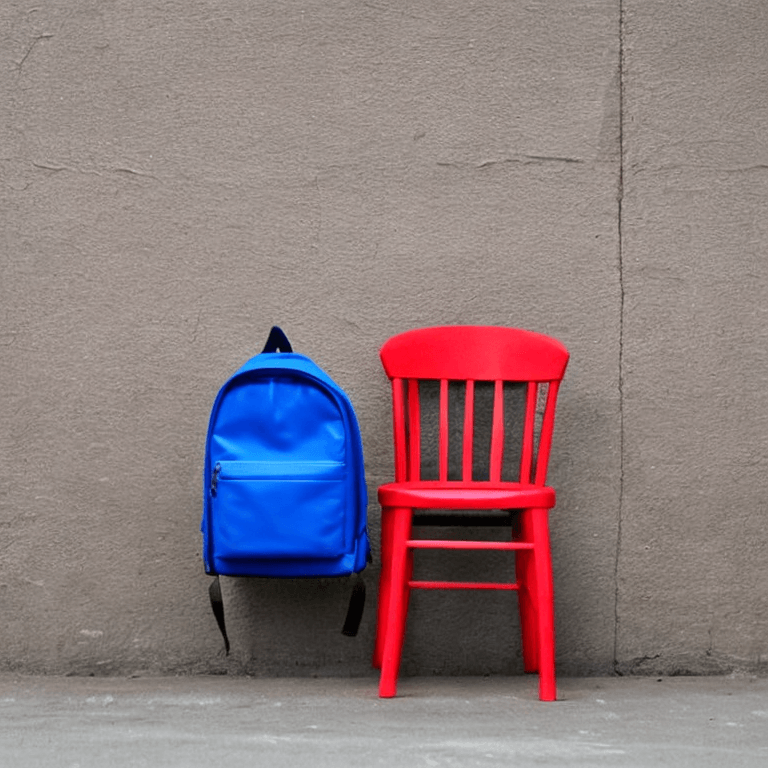} & \hspace*{-6.7pt} \includegraphics[width=0.89cm]{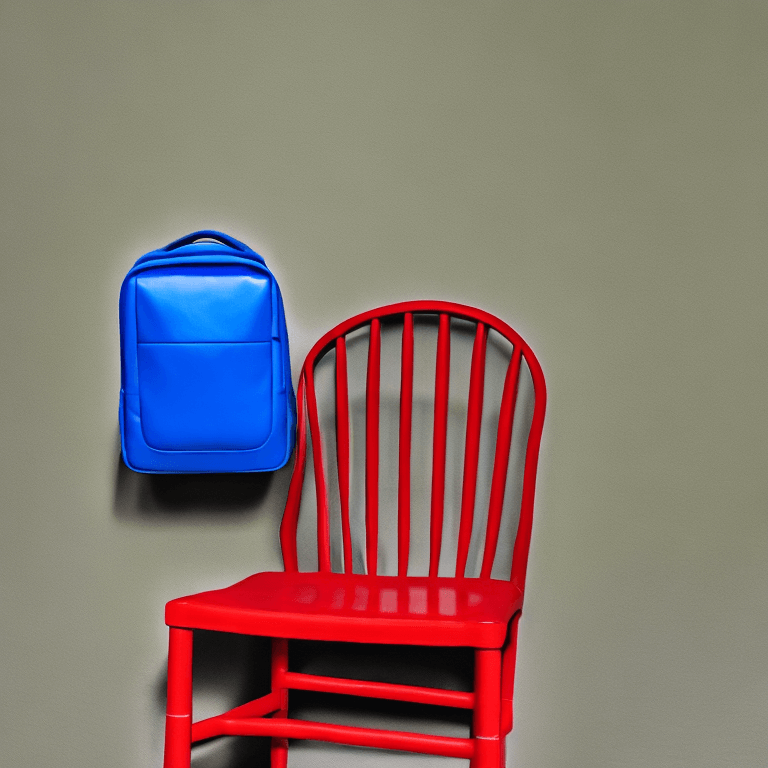} \\

\multicolumn{8}{c}{A blue backpack and a red chair}
\vspace*{4pt}\\ 
\multicolumn{2}{c}{SDXL} & \multicolumn{2}{c}{\wb} & \multicolumn{2}{c}{PixArt-$\alpha$} & \multicolumn{2}{c}{\wb} \vspace*{1pt} \\
\multicolumn{2}{c}{\includegraphics[width=1.8cm]{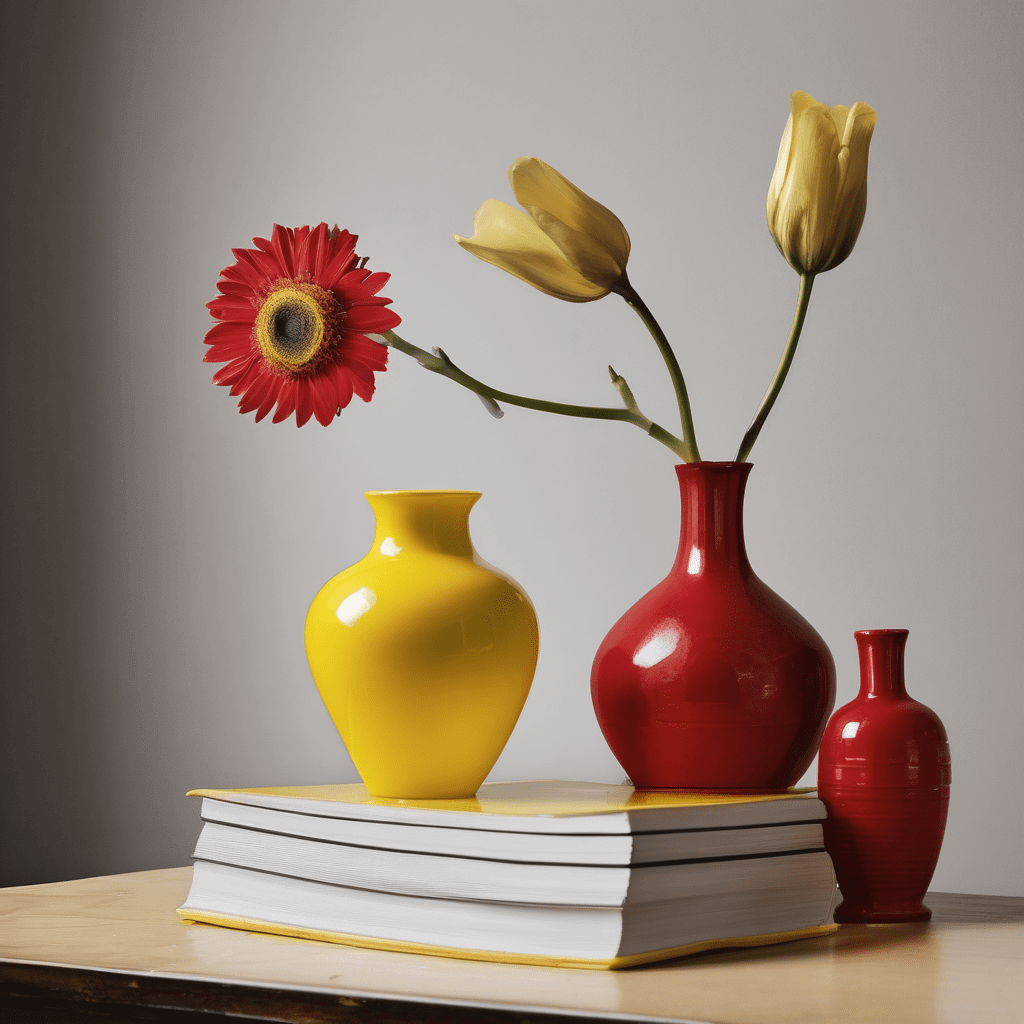}} & 
\multicolumn{2}{c}{\includegraphics[width=1.8cm]{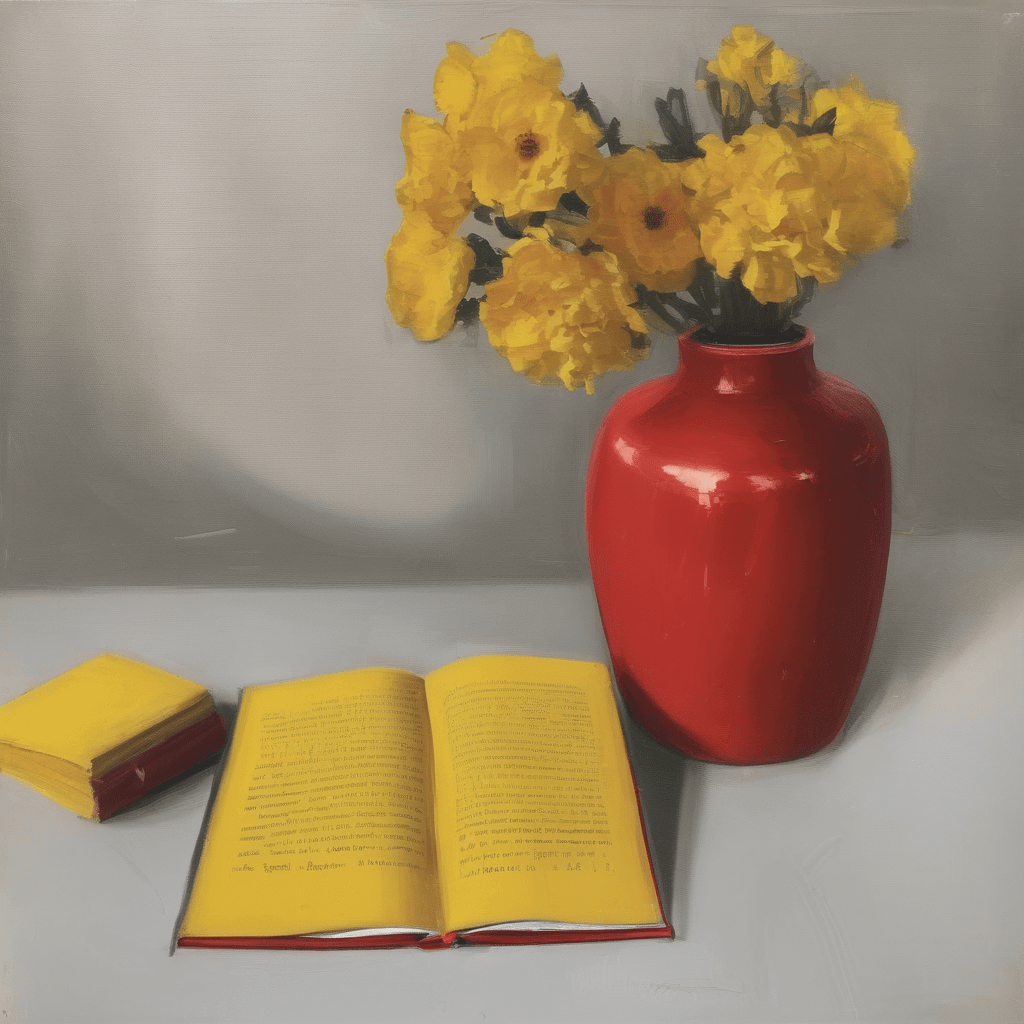}} & 
\multicolumn{2}{c}{\includegraphics[width=1.8cm]{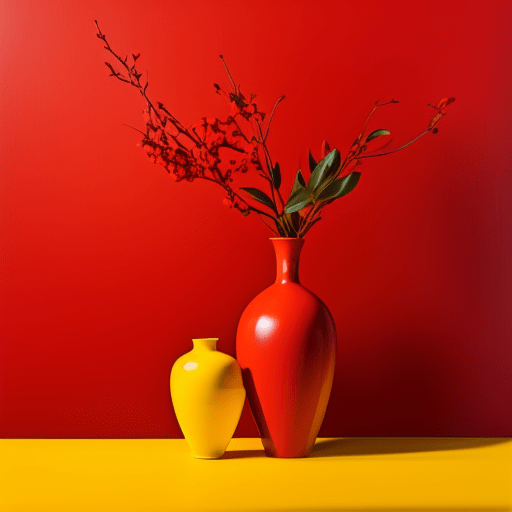}} & 
\multicolumn{2}{c}{\includegraphics[width=1.8cm]{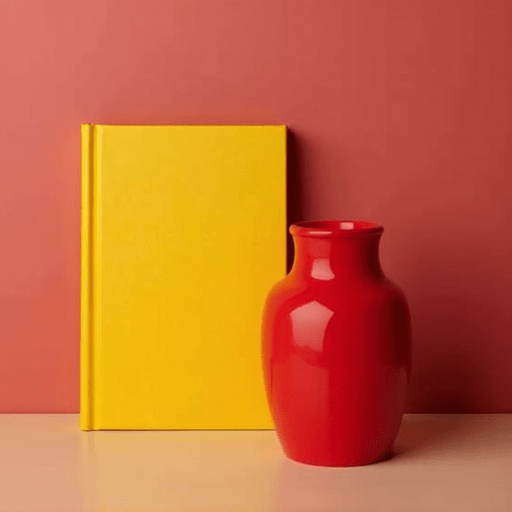}} \vspace*{-3.4pt} \\
\includegraphics[width=0.89cm]{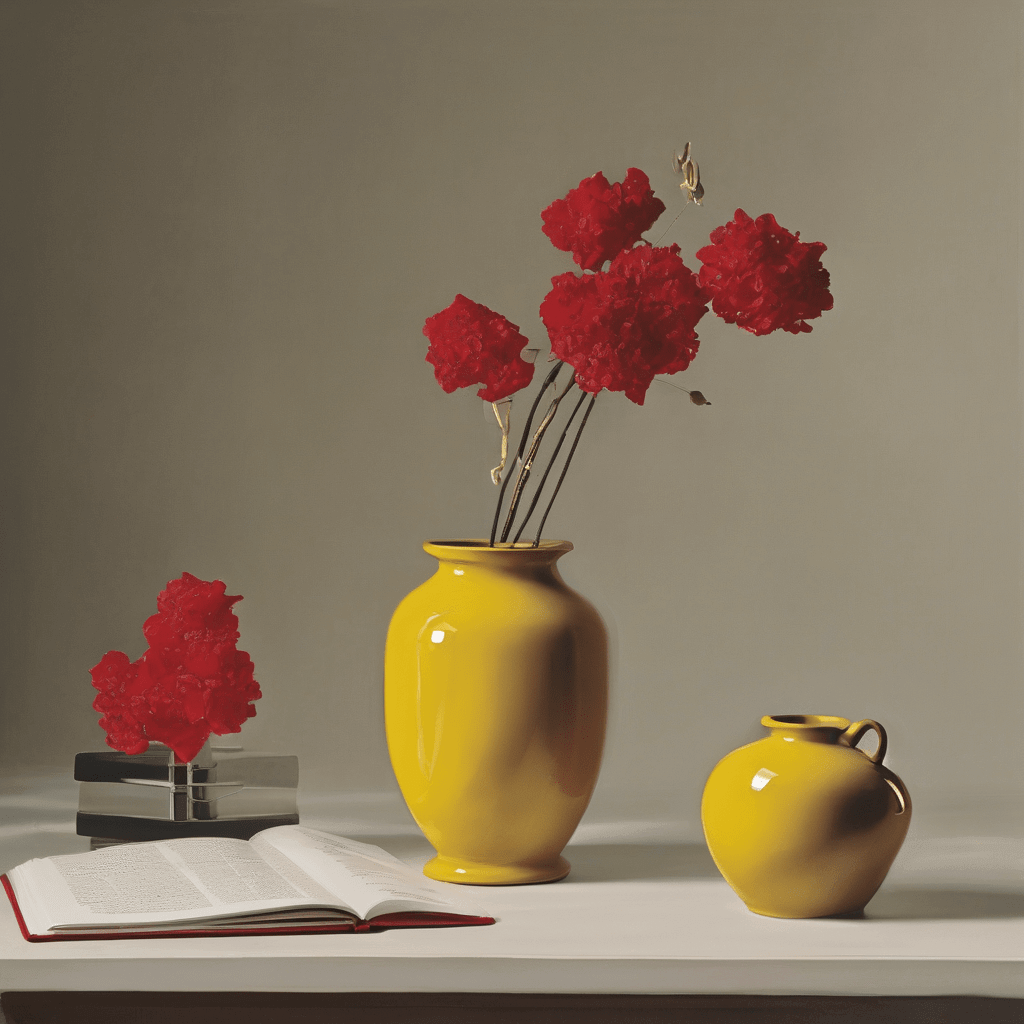} & \hspace*{-4pt}
\includegraphics[width=0.89cm]{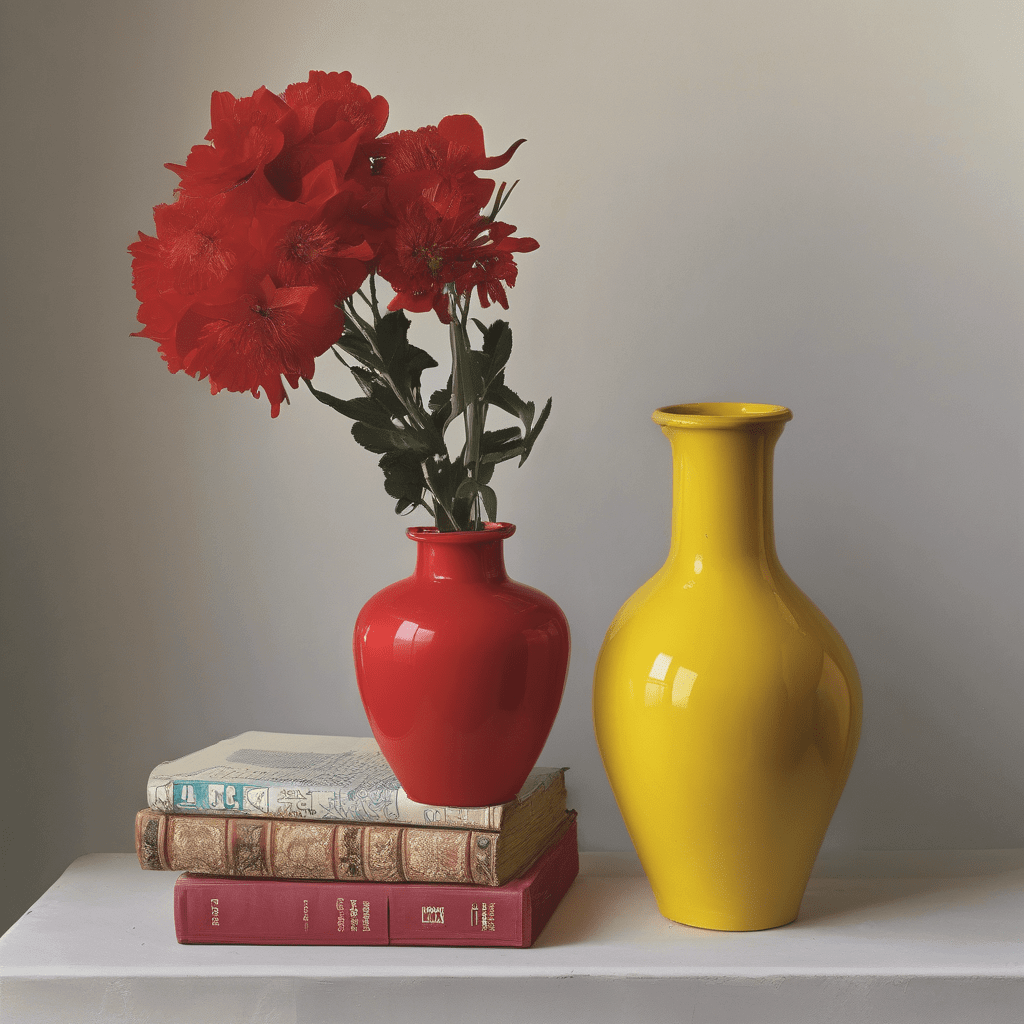} & \includegraphics[width=0.89cm]{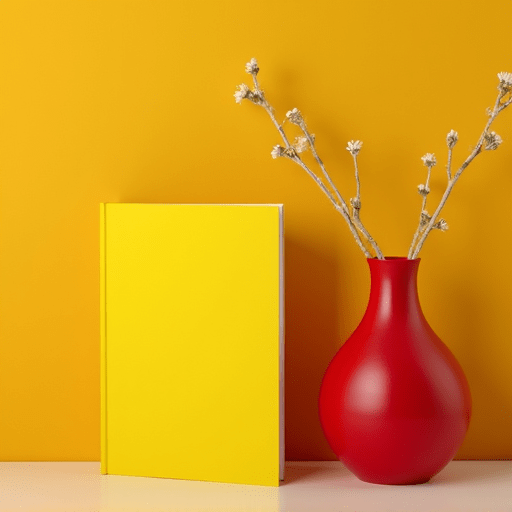} & \hspace*{-4pt} \includegraphics[width=0.89cm]{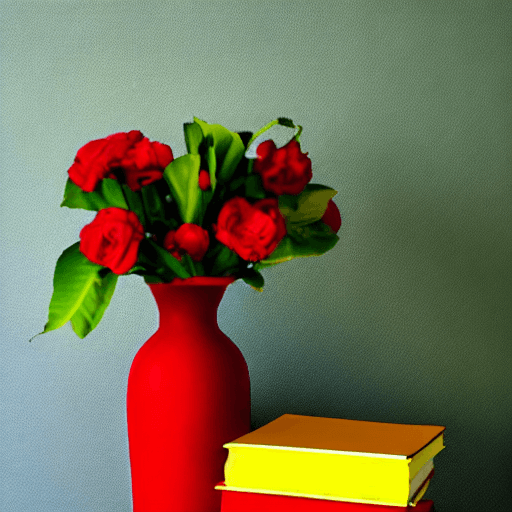} & \includegraphics[width=0.89cm]{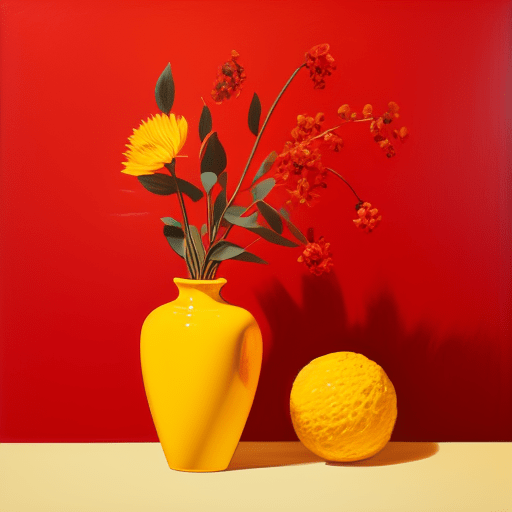} & \hspace*{-4pt} \includegraphics[width=0.89cm]{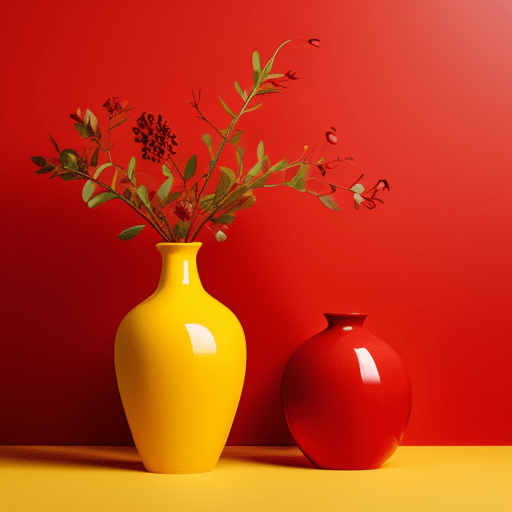} & \includegraphics[width=0.89cm]{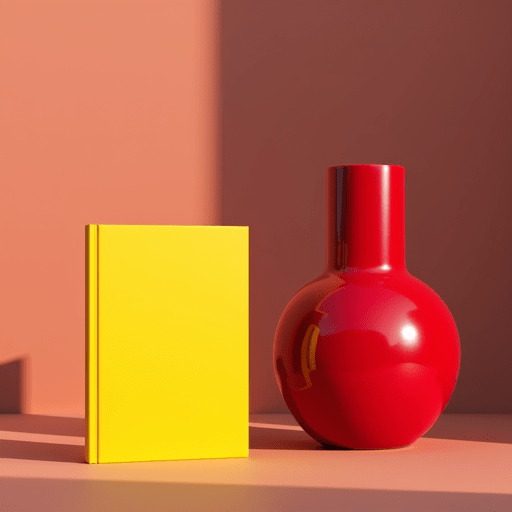} & \hspace*{-6.7pt} \includegraphics[width=0.89cm]{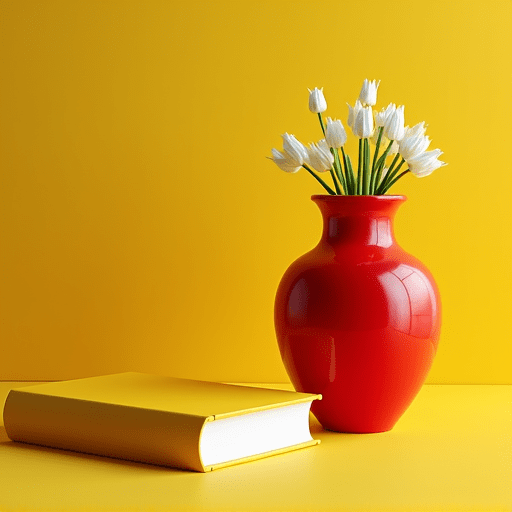} \\
\multicolumn{8}{c}{A yellow book and a red vase}
\end{tabular}

\vspace{-5pt}
\caption{Qualitative comparison of baselines and our projection method (\wb{}). Incorporating \wb{} significantly improves image alignment with the prompts.}
\label{fig:prompts_examples}
\end{figure}

In particular, we introduce Window-based Compositional Linear Projection ($\wb$), a \textit{lightweight} fine-tuning method that significantly improves the model’s performance on compositional prompts (Figure~\ref{fig:prompts_examples}), achieving results comparable to or surpassing existing methods. 
Additionally, $\wb$ preserves the model's overall performance, maintaining high fidelity on clean prompts as evidenced by a low FID score, while offering a solution that is both \textit{parameter efficient} and \textit{speed efficient}. This ensures robust compositional capabilities without compromising the model's general effectiveness.


In summary, our contributions are as follows:
\begin{itemize}
    \item We perform an in-depth analysis of the reasons behind compositionality failures in text-to-image generative models, with a particular focus on investigating the attribute binding aspect of compositionality. We highlight two key reasons contributing to these failures.
    \item Building on our observations, we propose $\wb$ as an enhancement for SD v1.4, SD v2, SDXL, SD v3, DeepFloyd, and PixArt-$\alpha$. This method significantly improves the models' compositional attribute binding, while preserving accuracy on standard prompts.
    We observe improvements of $16.18\%, 15.15\%$, and $9.51\%$ on SD v1.4, $14.35\%, 11.14\%$, and $6\%$ on SD v2, $20.31\%, 13.4\%$, and $5\%$ on SDXL, and $14.16\%$, $9.82\%$, and $2.63\%$ on PixArt-$\alpha$ in VQA scores across color, texture, and shape datasets, respectively.
    Our method outperforms or matches existing baselines in VQA scores, while achieving a superior FID score. It requires \textit{fewer parameters for optimization} and enables \textit{faster inference}, making it both efficient and effective. 
    
\end{itemize}


\section{Background}
\paragraph{Compositionality in Text-to-Image Generative Models.} Compositionality in text-to-image models refers to the ability of a model to accurately capture the correct compositions of objects, their corresponding attributes, and the relationships between objects described in a given prompt.  \citet{huang2023t2icompbench} introduced a benchmark designed to evaluate compositionality in text-to-image models, highlighting the limitations of models when handling compositional prompts. 
The benchmark employs disentangled BLIP-Visual Question Answering (VQA) as a metric for assessing image compositional quality. The VQA score assesses how accurately an image captures the compositional elements described in the prompt by utilizing a vision-language model.
This metrics demonstrates a closer correlation with human judgment compared to metrics like CLIP-Score \citep{hessel2021clipscore}.
The authors also proposed a fine-tuning baseline to enhance compositionality in these models. Alternatively, compositionality issues can be addressed at inference by modifying cross-attention maps using hand-crafted loss functions and bounding boxes derived from a language model~\citep{chefer2023attendandexcite, feng2023layoutgpt, agarwal2023astar, wang2023compositional, nie2024compositional, Lian2023LLMgroundedDE, Liu2022CompositionalVG}. However,~\citet{huang2023t2icompbench} showed that data-driven fine-tuning is more effective for improving compositionality.


\paragraph{Text-to-image Diffusion Models} 
In diffusion models, noise is added to the data following a Markov chain across multiple time-steps $t \in [0, T]$. Starting from an initial random real image $\vb{x}_0$ along with its  caption $c$, $(\vb{x}_{0}, c) \sim \mathcal{D}$, the noisy image at time-step $t$ is defined as $\vb{x}_{t} = \sqrt{\alpha_{t}}\vb{x}_{0} + \sqrt{(1-\alpha_{t})}\vb{\epsilon}$. 
The denoising network denoted by $\epsilon_{\theta}(\vb{x}_{t}, \vb{c}, t)$ is trained to denoise the noisy image $\vb{x}_{t}$ to obtain $\vb{x}_{t-1}$. 
For efficiency, the noising and the denoising operations occur in a latent space defined by $\vb{z} = \mathcal{E}(\vb{x})$, where $\mathcal{E}$ is an encoder such as VQ-VAE~\citep{DBLP:journals/corr/abs-1711-00937}.
Usually, the conditional input $\vb{c}$ to the denoising network $\epsilon_{\theta}(.)$ is a text-embedding of the caption $c$ through a text-encoder $\vb{c} = v_{\gamma}(c)$. The training objective for diffusion models can be defined as follows:
\begin{align*}
    \mathcal{L}(\theta) = \mathbb{E}_{(\vb{x}_0, c) \sim \mathcal{D}, \epsilon, t}
    \left[
    \krnorm{\epsilon - \epsilon_{\theta}(\vb{z}_{t}, \vb{c}, t)}_{2}^{2}\right],
\end{align*}    
where $\theta$ is the set of learnable parameters in the UNet $\epsilon_{\theta}$.  
During inference, given a text-embedding $\vb{c}$, a random Gaussian noise $\vb{z}_{T} \sim \mathcal{N}(0,I)$ is iteratively denoised to produce the final image.  




\section{Sources of Compositionality Failures}

This section conducts an in-depth analysis of compositional attribute binding failures in text-to-image models, focusing on the CLIP text-encoder.

\newcommand{\cont}{\text{cont}}
\newcommand{\attn}{\text{attn}}
\newcommand{\atq}{\text{q}}
\newcommand{\atk}{\text{k}}
\newcommand{\atv}{\text{v}}
\newcommand{\prvtoken}{\bar{\vb{x}}}

\subsection{Source (i) : Erroneous Attention Contributions in CLIP}
\label{sec:cont}

\begin{figure}[t]
\centering
\includegraphics[width=6.5cm]{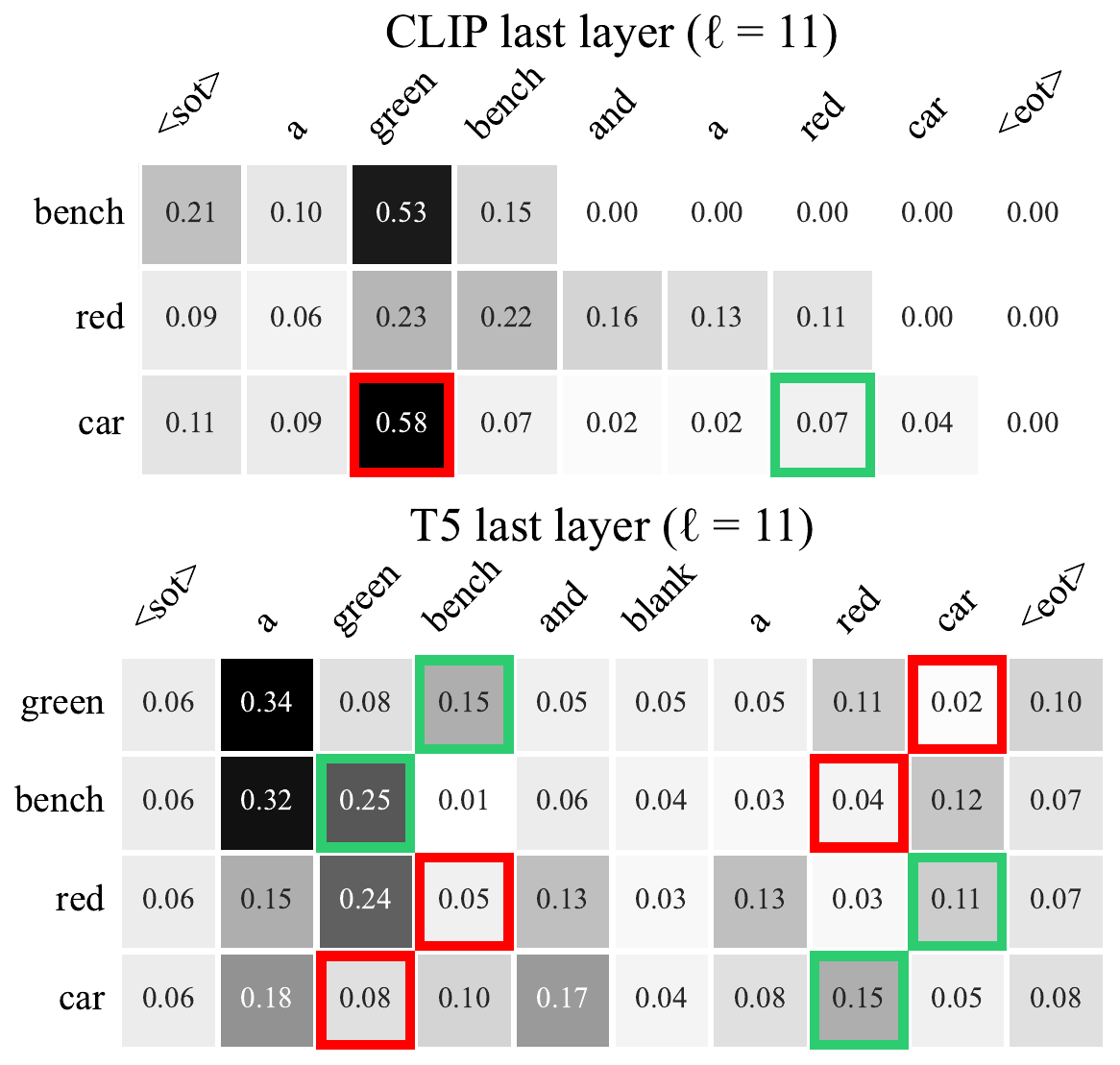}
\vspace{-7pt}
\caption{The heatmap illustrates unintended attention contributions in CLIP, while highlighting the more accurate performance of T5.}
\label{fig:t5-vs-clip}
\vspace{5px}

\includegraphics[width=7cm]{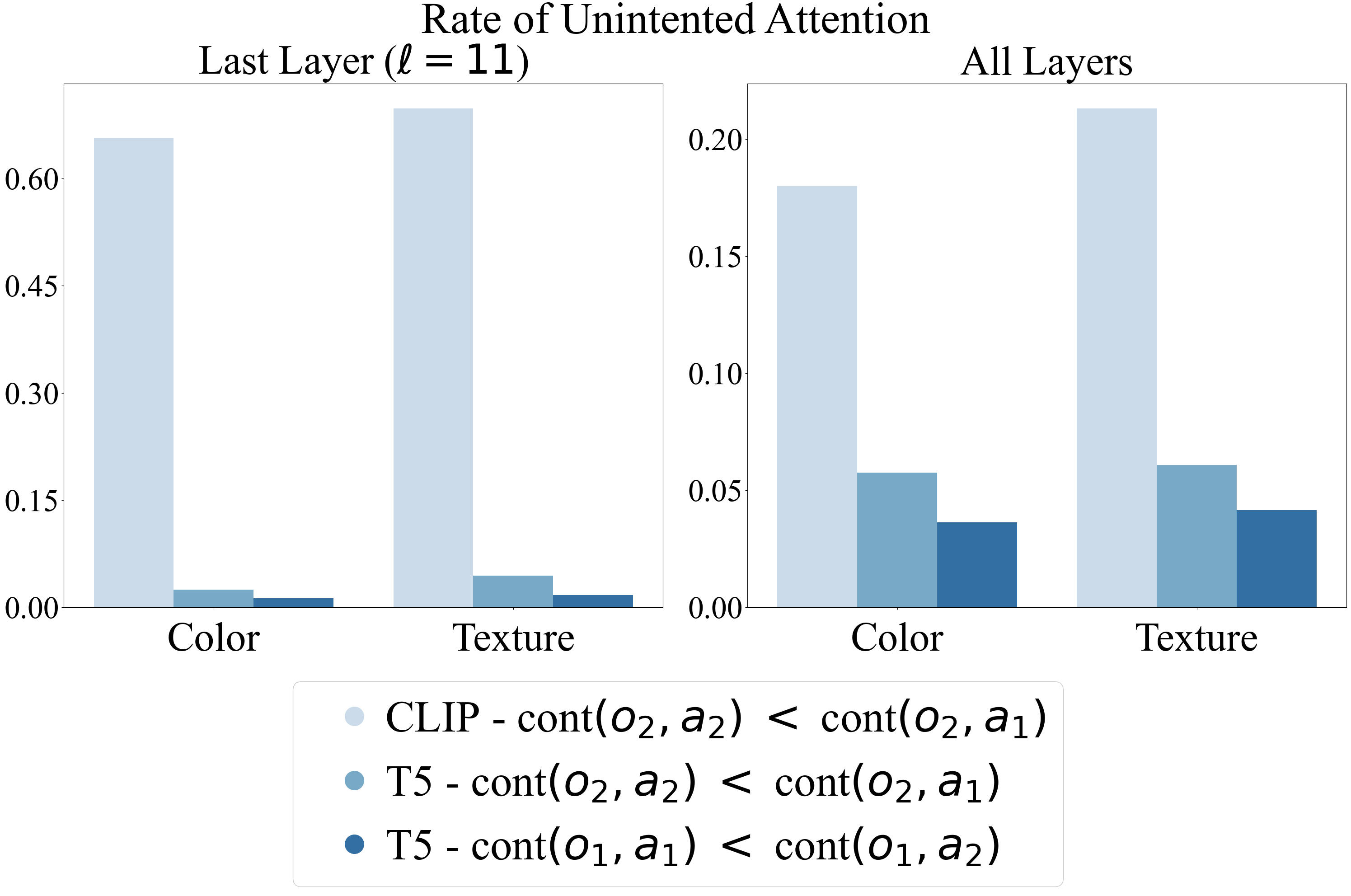}
\vspace{-8pt}
\caption{Quantitatively, we find CLIP to have significantly higher erroneous attention contributions averaged across prompts of color and texture datasets.}
\vspace{-5px}
\label{fig:t5-vs-clip-agg}
\end{figure}


In this section, we leverage attention contributions~\citep{elhage2021mathematical, dar2023analyzing} to analyze how the final text-embeddings of compositional prompts are obtained by the CLIP text-encoder, a widely adopted component in many text-to-image models. We then compare these attention contribution patterns with those produced by the T5 text-encoder which is known for its stronger compositional capabilities.
Many of the compositional prompts from~\citet{huang2023t2icompbench} have a decomposable template of the form $\vb{a}_{i}\ \vb{o}_{j} + \vb{a}_{j}\ \vb{o}_j$, where $\vb{a}_{i}, \vb{a}_{j}$ are attributes (\eg, ``black'', ``matted'') and $\vb{o}_{i}, \vb{o}_{j}$ represent objects (\eg, ``car'', ``bag'').

The attention mechanism in layer $\ell$ of a transformer consists of four weight matrices $W_{\text{q}}, W_{\text{v}}, W_{\text{k}},$ and $W_{\text{o}}$~\citep{vaswanie2021attention}.
Each of these matrices is divided into $H$ heads, denoted by $W_{\text{q}}^{h}, W_{\text{v}}^{h}, W_{\text{k}}^{h} \in \mathbb{R}^{d \times d_{h}}, W_{\text{o}}^{h} \in \mathbb{R}^{d_{h} \times d}$, where $h \in \left[H\right]$.
Here, $d_h$ denotes the dimensionality of the internal token embeddings. For simplicity, we omit $\ell$, but each layer has its own attention matrices.
These matrices operate on the token embeddings produced by the previous layer ($\ell - 1$), denoted as $\prvtoken_j$ for token $j$.
We further denote the projections of $\prvtoken_j$ onto the query, key, and value matrices of the $h$-th attention head in layer $\ell$ as $\atq_{j}^{h}$, $\atk_{j}^{h}$, and $\atv_{j}^{h}$, respectively.
More precisely, 
{\setlength{\abovedisplayskip}{5pt} 
\setlength{\belowdisplayskip}{5pt}
\begin{align*}
    \atq_{j}^{h} = \prvtoken_{j} W_{\text{q}}^{h},\quad
    \atk_{j}^{h} = \prvtoken_{j} W_{\text{k}}^{h},\quad
    \atv_{j}^{h} = \prvtoken_{j} W_{\text{v}}^{h}.
\end{align*}}
The \textit{contribution} of token $j$ to token $i$ in layer $\ell$, denoted by $\cont_{i,j}$, is computed as follows:
{\setlength{\abovedisplayskip}{3pt} 
\setlength{\belowdisplayskip}{3pt}
\begin{align*} \label{eq:attn-cont}
    \cont_{i, j} = \krnorm{\sum_{h=1}^{H} \attn_{i, j}^{h}\ 
    \atv_j^{h}
    \ W_{\text{o}}^{h}}_2 
\end{align*}}
where $\attn_{i, j}^{h}$ is the attention weight of token $i$ to $j$ in the $h$-th head of layer $\ell$. Specifically,
{\setlength{\abovedisplayskip}{5pt} 
\setlength{\belowdisplayskip}{5pt}
\begin{align*}
    \attn_{i, .}^{h} =
    \textsc{Softmax}\left( \left\{
    \frac{
        \langle \atq_i^{h}, \atk_j^{h}\rangle 
    }
    {
    \sqrt{d_h}
    }
    \right\}_{j=1}^{n}
    \right).
\end{align*}}
Notably, $\cont_{i, j}$ is a metric that quantifies the \textit{contribution} of a token $j$ to the norm of a token $i$ at layer $\ell$. We employ this metric to identify layers in which important tokens highly attend to \textit{unintended} tokens, or lowly attend to \textit{intended} ones. See Appendix~\ref{app:sec:cont} for details on attention contribution.



\textbf{Key Finding: T5 has less erroneous attention contributions than CLIP.}
We refer to Figure~\ref{fig:t5-vs-clip} that visualizes attention contribution of both T5 and CLIP text-encoder in the last layer $(\ell = 11)$ for the prompt ``a green bench and a red car". Ideally, the attention mechanism should guide the token ``car'' to focus more on ``red'' than ``green'', but in the last layer of the CLIP text-encoder, ``car'' significantly attends to ``green''. In contrast, T5 shows a more consistent attention pattern, with ``red'' contributing more to the token ``car'' and ``green'' contributing more to the token ``bench''.

We further conduct a comprehensive analysis focusing on specific types of compositional prompts from the T2I-CompBench dataset~\citep{huang2023t2icompbench}. This includes $780$ prompts from the color category and $582$ prompts from the texture category of this dataset, each following the structured format: “$\vb{a}_1$ $\vb{o}_1$ and $\vb{a}_2$ $\vb{o}_2$”. For each prompt, we obtain attention contributions in all layers and count the number of layers where \textit{unintended attention contributions} occur. In the CLIP text-encoder, unintended attention occurs when $\vb{o}_2$ attends more to $\vb{a}_1$ than $\vb{a}_2$. For T5, it occurs when $\vb{o}_2$ attends more to $\vb{a}_1$ than $\vb{a}_2$, or $\vb{o}_1$ attends more to $\vb{a}_2$ than $\vb{a}_1$.
Figure~\ref{fig:t5-vs-clip-agg} provides a quantitative comparison of unintended attention across various prompts between the CLIP text-encoder and T5.
The T5 model demonstrates superior performance on our metric compared to the CLIP text-encoder,
reinforcing the hypothesis that erroneous attention mechanisms in CLIP may contribute to its weaker compositional performance in text-to-image models.
Additional details can be found in Appendix~\ref{app:t5}.
Further experiments with other text-encoders are also reported in Appendix~\ref{app:llama}.



To address the attention shortcomings of the CLIP text-encoder, we explored zero-shot reweighting of attention maps in CLIP to reduce unintended attentions while enhancing meaningful ones. While this improved baseline performance, it fell short of our primary method discussed in the following sections. See Appendix~\ref{app:sec:zero-shot} for more details.

\subsection{Source (ii) : Sub-optimality of CLIP Text-Encoder for Compositional Prompts}
\label{sec:optimized-text-embeddings}

\begin{figure}
\centering
\includegraphics[width=0.48\textwidth]{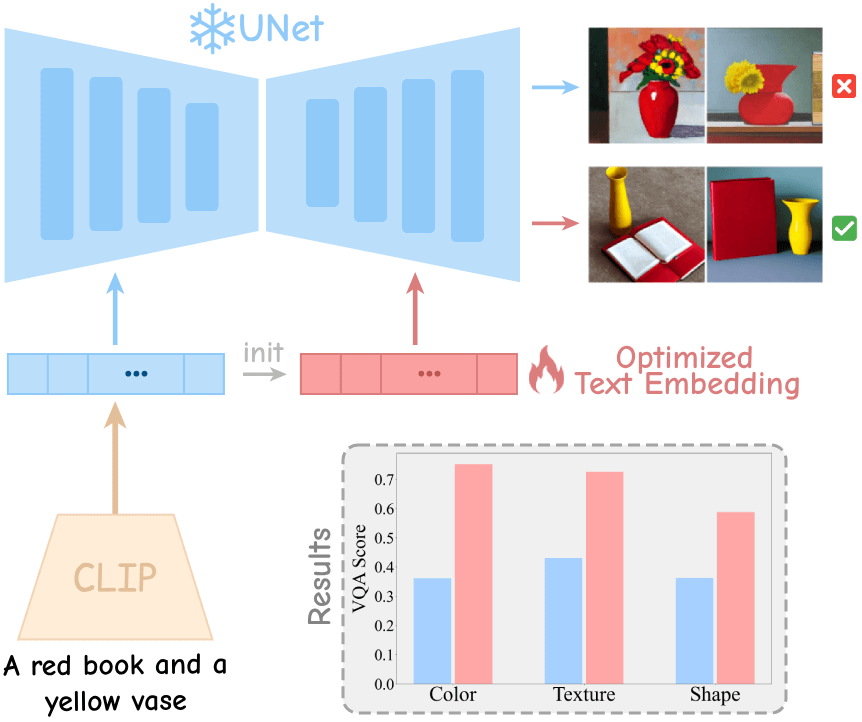}
\caption{Sub-optimality of CLIP Text-Encoder for Compositional Prompts. Optimizing a learnable vector to represent an improved text embedding, while keeping the UNet frozen, enables the generation of more compositionally accurate images.}
\label{fig:opt_emb_pipeline}
\end{figure}

In this section, we investigate whether the UNet is capable of generating better compositional scenes if provided with alternative (\textit{improved}) text embeddings, rather than relying on the output of the CLIP text-encoder. For a given input prompt $p$ with a specific composition (\eg, {\it ``a red book and a yellow table''}), we utilize our dataset (described in Section~\ref{sec:dset}) to obtain $\mathcal{D}_p$, a set of high-quality compositional images corresponding to prompt $p$. Next, we extract the text embedding $\vb{c}$ from the CLIP text-encoder for prompt $p$. Using this embedding as the initialization, we create a learnable vector $\vb{c}^{*}$ of the same dimensionality. Keeping all other components (such as the UNet) frozen, we optimize this learnable vector as follows:
{\setlength{\abovedisplayskip}{7pt} 
\setlength{\belowdisplayskip}{7pt}
\begin{align*}
    \vb{c}^{*} = \arg \min_{\vb{c}} \mathbb{E}_{x_0 \sim \mathcal{D}_p, \epsilon, t}
    \left[
    \krnorm{\epsilon - \epsilon_{\theta}(\vb{z}_{t}, \vb{c}, t)}_{2}^{2}\right].
\end{align*}}
We then use the optimized text embedding $\vb{c}^{*}$ to generate images with the UNet $\epsilon_{\theta}$. Figure~\ref{fig:opt_emb_pipeline} illustrates the complete pipeline.

\textbf{Key Results.} 
Utilizing Stable Diffusion v1.4, we optimize optimize $\vb{c}^{*}$ for all compositional prompts across the color, texture, and shape categories in the T2I-CompBench dataset. By generating samples with $\vb{c}^{*}$ and comparing them to those generated using $\vb{c}$, we observe a significant improvement in the VQA scores.  As shown in Figure~\ref{fig:opt_emb_pipeline}, CLIP text embeddings yield VQA scores of $0.3615$ for color, $0.4306$ for texture, and $0.3619$ for shape. In contrast, the optimized embeddings achieve $0.7513$ for color, $0.7254$ for texture, and $0.5873$ for shape.

These results indicate that CLIP text-encoder does not output the proper text-embedding suitable for generating compositional scenes. However, the existence of an optimized embedding space demonstrates that the UNet can generate coherent compositional outputs when provided with appropriately improved embeddings. This finding motivates the idea  of improving the CLIP output space to mitigate compositionality issues in text-to-image diffusion models. For additional configurations, including results from optimizing a subset of tokens to improve compositionality, refer to Appendix~\ref{app:sec:optimize}.

\section{Projection Layer for Enhancing Compositionality in the CLIP Text Embedding Space}
\label{sec:projection_layer}

Building on our previous findings, we focus on improving the text embedding space utilized in text-to-image generative models. Specifically, we propose learning a projection layer over the CLIP output embedding space to transform its sub-optimal representation into an enhanced space better suited for compositionality. In the following sections, we introduce two methods, \lp{} and \wb{}, which implement linear projections of the CLIP output embedding space to achieve this enhancement.



\subsection{\lp{}: Token-wise Compositional Linear Projection}


Given the text-embedding $\vb{c} \in \mathbb{R}^{n \times d}$ as the output of the text-encoder for prompt $c$, i.e., $\vb{c} = v_{\gamma}(c)$,
we train a linear projection $\lp_{W, b}: \mathbb{R}^{n \times d} \rightarrow \mathbb{R}^{n \times d}$.
This projection includes a matrix $W \in \mathbb{R}^{d \times d}$ and a~bias term $b \in \mathbb{R}^{d}$,
which are applied token-wise to the output text-embeddings of the text-encoder.
More formally, for $\vb{c} \in \mathbb{R}^{n\times d}$ including text-embeddings of $n$ tokens $\vb{c}_1, \vb{c}_2, \dotsm, \vb{c}_n \in \mathbb{R}^{d}$, $\lp_{W, b}(\vb{c})$ is obtained by stacking projected embeddings $\vb{c}'_1, \vb{c}'_2, \dotsm, \vb{c}'_n$ where $\vb{c}'_i = W^T \vb{c}_i + b$.

Finally, we solve the following optimization problem on a dataset $\mathcal{D}$ including image-caption pairs of high-quality compositional images:
{\setlength{\abovedisplayskip}{7pt} 
\setlength{\belowdisplayskip}{7pt}
\begin{align*}
    W^{*}, b^* = \arg \min_{W, b} \mathbb{E}_{(x_0, c) \sim \mathcal{D}, \epsilon, t} \left[
    \Phi_{\lp}
    \right] \\
    \Phi_{\texttt{Proj}} = \krnorm{\epsilon - \epsilon_{\theta}\left(\vb{z}_{t}, {\scriptstyle\texttt{Proj}}_{W, b}\left(\vb{c}\right), t\right)}_{2}^2
\end{align*}}
We then apply $\lp_{W^{*}, b^{*}}$ on CLIP text-encoder to obtain improved embeddings.

\subsection{\wb{}: Window-based Compositional Linear Projection}

In this section, we propose a more advanced linear projection scheme where the new embedding of a token is derived by applying a~linear projection on that token in conjunction with a set of its adjacent tokens within a specified window.
This method not only leverages the benefits of \lp{} but also incorporates the contextual information from neighboring tokens, potentially leading to more precise text-embeddings.

More formally, we train a mapping $\wb_{W, b} : \mathbb{R}^{n \times d} \rightarrow \mathbb{R}^{n \times d}$ including a parameter $s$
(indicating window length), matrix $W \in \mathbb{R}^{(2s+1)d \times d}$, and a~bias term $b \in \mathbb{R}^{d}$.
For text-embeddings $\vb{c} \in \mathbb{R}^{n \times d}$ consisting of $n$ token embeddings of $\vb{c}_1, \vb{c}_2, \dotsm, \vb{c}_n \in \mathbb{R}^d$, we obtain $\wb_{W, b}$ by stacking projected embeddings $\vb{c}'_1, \vb{c}'_2, \dotsm, \vb{c}'_n$ where
{\setlength{\abovedisplayskip}{7pt} 
\setlength{\belowdisplayskip}{7pt}
\begin{align*}
    \vb{c}'_i = W^T \ \textsc{Concatenation}\left(\left(\vb{c}_{j}\right)_{j=i-s}^{i+s}\right) + b
\end{align*}}
Similarly, we solve the following optimization problem to train the projection:
{\setlength{\abovedisplayskip}{7pt} 
\setlength{\belowdisplayskip}{7pt}
\begin{align*}
    W^{*}, b^* = \arg \min_{W, b} \mathbb{E}_{(x_0, c) \sim \mathcal{D}, \epsilon, t} \left[
    \Phi_{\wb}
    \right] 
\end{align*}}

\begin{figure}[t]
\centering
\setlength{\tabcolsep}{1pt}
\begin{tabular}{ccccc}
\multicolumn{5}{c}{\small ''A bathroom with green tile and a red shower curtain''} \\
\includegraphics[width=1.45cm]{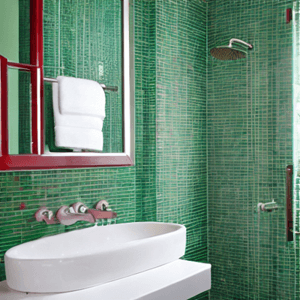}& 
\includegraphics[width=1.45cm]{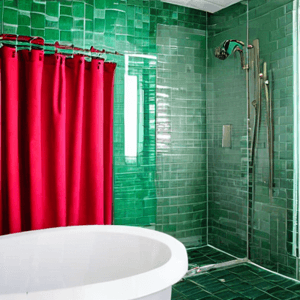} & 
\includegraphics[width=1.45cm]{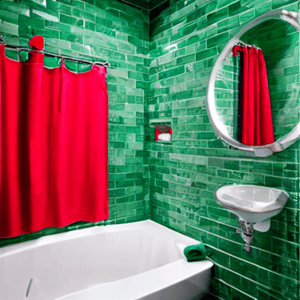} & 
\includegraphics[width=1.45cm]{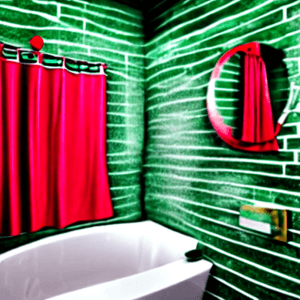} & 
\includegraphics[width=1.45cm]{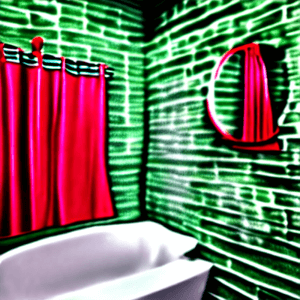} \\
\makecell{\footnotesize $\tau=1000$ \\[-5pt] \scriptsize{(No Guidance)}} & {\small$\tau=900$} & {\small$\tau=800$} & {\small$\tau=600$}  & {\small$\tau=200$} \\

\end{tabular}

\vspace{-7pt}
\caption{Qualitative results showing the impact of \earlystop{} with varying thresholds $\tau$.}
\label{fig:early_guidance_examples}
\vspace{5pt}
\includegraphics[width=7.5cm]{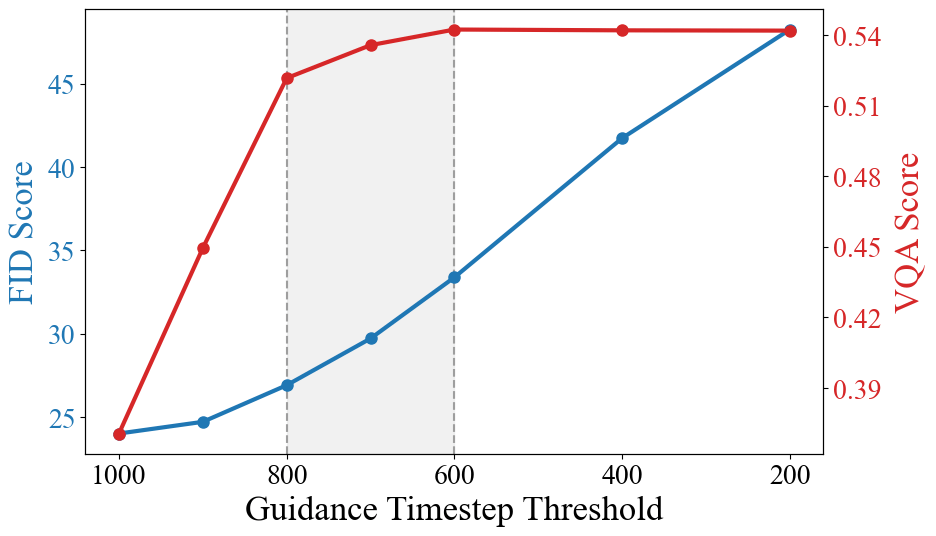}
\vspace{-7pt}
\caption{Trade-off between VQA and FID scores with \earlystop{} at different thresholds.}
\label{fig:early_guidance_vqa_vs_fid}
\end{figure}


We observe that \wb{} improves over \lp{} (special case of \wb{} with $s=0$) by incorporating adjacent tokens along with the token itself. This approach enhances embeddings by reinforcing the contributions of relevant adjacent tokens. See Appendix~\ref{app:set:window-len} for details on window length in $\wb{}$. 


\subsection{\earlystop{}: Trade-off between Compositionality and Clean Accuracy}
Fine-tuning models or adding modules to a base model often results in a degradation of image quality and an increase in the Fréchet Inception Distance (FID) score. To balance the trade-off between improved compositionality and the quality of generated images for clean prompts, nspired by \citet{hertz2022prompttoprompt}, we adopt \earlystop{}, where we apply the linear projection only during the initial steps of inference.
Specifically, given a time-step threshold $\tau$, for $t \geq \tau$,
we use $\wb_{W^{*}, b^{*}}(\vb{c})$, while for $t < \tau$,
we use the unchanged embedding $\vb{c}$ as the input to the cross-attention layers.

Figure~\ref{fig:early_guidance_vqa_vs_fid} illustrates the trade-off between VQA score and FID on a randomly sampled subset of MS-COCO~\citep{coco} for different choices of $\tau$.
As shown, even a large value of $\tau$ suffices for obtaining high-quality compositional scenes as the composition of final generated image is primarily formed at early steps.
Thus, choosing a large $\tau$ preserves the model's improved compositionality while maintaining its clean accuracy. Setting $\tau = 800$ offers a competitive VQA score compared to the model where projection is applied at all time steps, and achieves a competitive FID similar to that of the clean model. Figure~\ref{fig:early_guidance_examples} depicts a few images generated using different choices of $\tau$.
We refer to Appendix~\ref{app:sec:early-guidance} for more visualizations.

\section{Experiments}

\textbf{Existing Baselines.}
We evaluate the performance of multiple methods alongside standard models SD v1.4, SD v2, SDXL \citep{podell2023sdxl}, SD v3 \citep{esser2024scaling}, and PixArt-$\alpha$ \citep{chen2023pixartalpha}. These include Composable Diffusion \citep{liu2022compositional}, which addresses concept conjunction and negation in pretrained diffusion models; Structured Diffusion \citep{feng2022training}, which focuses on attribute binding; Attn-Exct \citep{chefer2023attendandexcite}, which ensures correct attention to all subjects in the prompt using iterative optimizations; GORS \citep{huang2023t2icompbench}, which fine-tunes Stable Diffusion v2 using a reward function; GLIGEN \citep{li2023gligenopensetgroundedtexttoimage}, which utilizes grounding inputs such as bounding boxes; RealCompo \citep{zhang2024realcompobalancingrealismcompositionality}, which integrates spatial-aware diffusion models; and FLUX \citep{flux2024}. 


\textbf{Training Setup.}
All of the models are trained using the objective function of diffusion models on color, texture, and shape datasets described in Section \ref{sec:dset}.
During training, we keep all major components frozen, including the U-Net, CLIP text-encoder, and VAE encoder and decoder, and only the linear projections are trained. We refer to Appendix \ref{app:sec:train_setup} for details on the training procedure.

\begin{figure}[]
\centering
\setlength{\tabcolsep}{1pt}
\begin{tabular}{cccccc}

                  & \multicolumn{5}{c}{prompt: ''A green bench and a yellow dog''} \\
                  & image    & green         & bench         & yellow         & dog         \\
\raisebox{16pt}{\rotatebox[origin=c]{90}{\small{Baseline}}} &
\includegraphics[width=1.4cm]{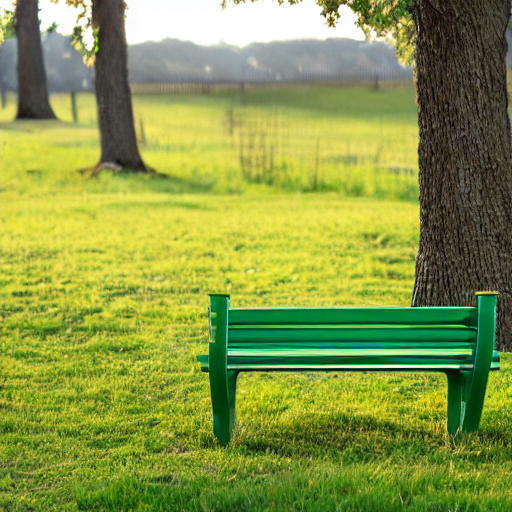} & 
\includegraphics[width=1.4cm]{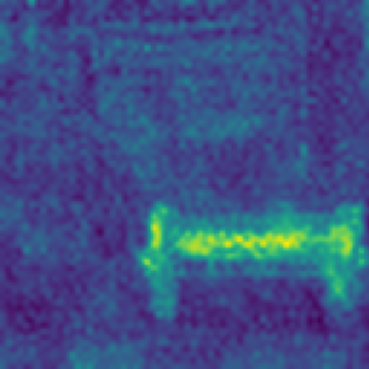} &
\includegraphics[width=1.4cm]{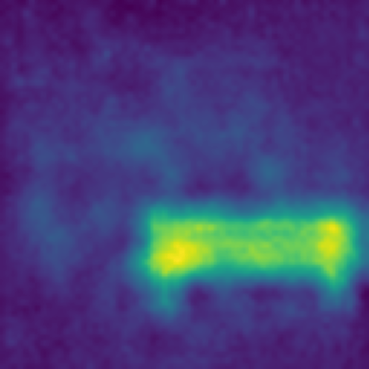} &
\includegraphics[width=1.4cm]{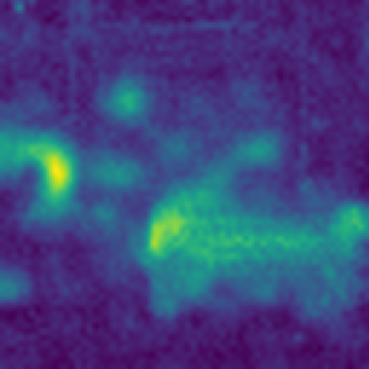} &
\includegraphics[width=1.4cm]{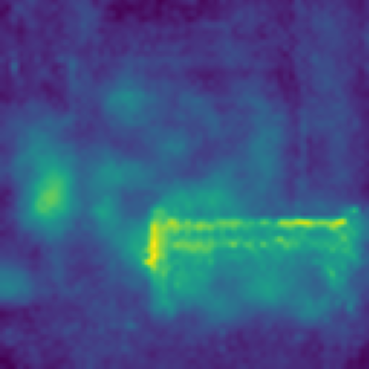} \\
\raisebox{16pt}{\rotatebox[origin=c]{90}{\small{\wb}}} &
\includegraphics[width=1.4cm]{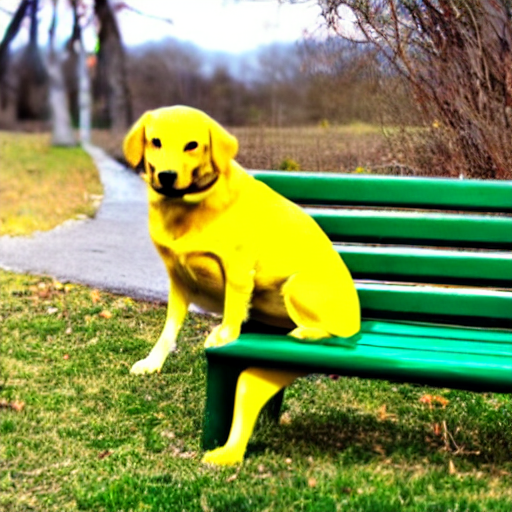} & 
\includegraphics[width=1.4cm]{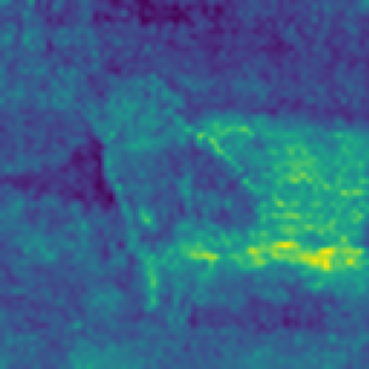} &
\includegraphics[width=1.4cm]{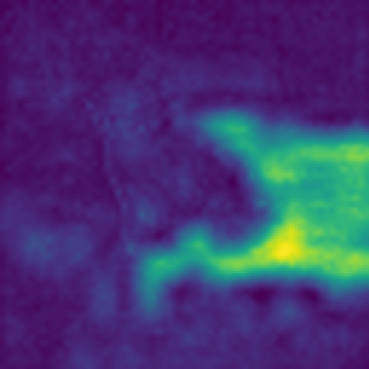} &
\includegraphics[width=1.4cm]{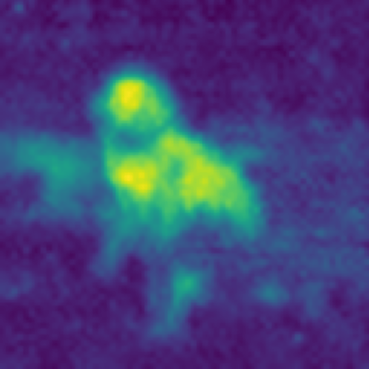} &
\includegraphics[width=1.4cm]{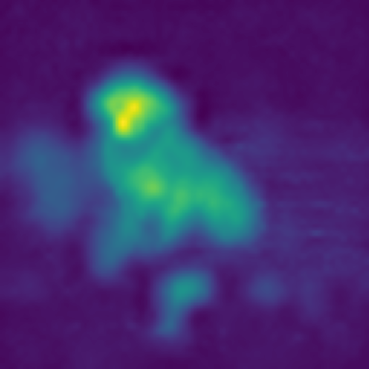} 
\end{tabular}

\vspace{-5pt}

\caption{More accurate cross-attention maps using \lp{}.}

\label{fig:attention_visualization_of_linear_projection}
\end{figure}

\textbf{Dataset Collection.}
\label{sec:dset}
We utilize the T2I-CompBench dataset \citep{huang2023t2icompbench}, a~well-recognized dataset for compositionality, focusing on three categories: color, texture, and shape, with a total of 1,000 prompts across both training and evaluation sets.
To generate high-quality images, we use three generative models: SD v1.4, DeepFloyd, and SynGen \citep{rassin2024linguistic}, creating 210 samples per prompt. This ensures a wide variety of generated images, leveraging each model's strengths. From these, we selected the top 30 with the highest VQA scores to ensure the final dataset consists of images that best reflect the prompts. 

Furthermore, for SDXL, SD v3, and PixArt-$\alpha$, we explored training \wb{} (\wb{}* in Table \ref{table:comparison_with_sota}) on a higher-quality dataset generated by newer text-to-image models, such as SDXL and SD v3. Importantly, leveraging an appropriately curated dataset results in a substantial improvement in VQA scores, highlighting the importance of high-quality training data for compositional understanding.

\subsection{Qualitative and Quantitative Evaluation}

\textbf{Qualitative Evaluation.}
Figure~\ref{fig:prompts_examples} presents images generated when applying \wb{}.
When generating compositional prompts with a baseline model,
objects may be missing or attributes are incorrectly applied.
However, with \wb{}, objects and their corresponding attributes are more accurately generated.
See Appendix~\ref{app:sec:vis-lp-wb} for more visualizations. 

Figure \ref{fig:attention_visualization_of_linear_projection} illustrates cross-attention maps for a sample prompt.
In the base model, attention maps are flawed, with some tokens incorrectly attending to the wrong pixels.
However, with both \lp{} and \wb{}, objects and attributes more accurately attend to their respective pixels.
For more visualizations, see Appendix~\ref{app:sec:cross-attn}.

\begin{table}[t]
\centering
\footnotesize
\begin{tabular}{c|l|c|c|c}
\multicolumn{2}{c|}{{Model}} & \multirow{1}{*}{Color} & \multirow{1}{*}{{Texture}} & \multirow{1}{*}{{Shape}} \\ \hline \hline  &&&&\\[-9pt] 

\multirow{4}{*}{\raisebox{15pt}{\rotatebox[origin=c]{90}{\scriptsize{SD v1.4}}} }  & Baseline    & 0.3765         & 0.4156           & 0.3576         \\ \cdashline{2-5}  &&&&\\[-8pt]
& $\lp$ & 0.4837            & 0.5312              & 0.4307            \\
& $\wb$ & \textbf{0.5383}            & \textbf{0.5671}              & \textbf{0.4527}            \\ \hline   &&&&\\[-9pt] 
\multirow{8}{*}{\raisebox{13pt}{\rotatebox[origin=c]{90}{\scriptsize{SD v2}}}} & Baseline        & 0.5065         & 0.4922           & 0.4221         \\
& Composable                & 0.4063         & 0.3645           & 0.3299         \\ 

& Structured               & 0.4990         & 0.4900           & 0.4218         \\ 
& Attn-Exct                & 0.6400         & 0.5963           & 0.4517         \\ 
& GORS              & 0.6414         & 0.6025           & 0.4546         \\ 
\cdashline{2-5}  &&&&\\[-8pt] 
& $\lp$   & 0.6075 & 0.5707 & 0.4567 \\ 
& $\wb$   & \textbf{0.6500}            & \textbf{0.6036}              & \textbf{0.4821}            \\  \hline  &&&&\\[-9pt] 
\multirow{4}{*}{\raisebox{11pt}{\rotatebox[origin=c]{90}{\scriptsize{SDXL}}} }  & Baseline    & 0.5770         & 0.5217           & 0.4666         \\ \cdashline{2-5} &&&&\\[-8pt] 
& $\wb$ & 0.6930            & 0.6007              & 0.4758            \\
& $\wb$* & \textbf{0.7801}            & \textbf{0.6557}              & \textbf{0.5166}            \\ \hline  &&&&\\[-9pt] 
\multirow{4}{*}{\raisebox{14pt}{\rotatebox[origin=c]{90}{\scriptsize{PixArt-$\alpha$}}} }  & Baseline\textsubscript{\tiny{$512^2$}}    & 0.3877         & 0.4557           & 0.4094         \\ 
 & Baseline\textsubscript{\tiny{$1024^2$}}    & 0.4156         & 0.4594           & 0.3849         \\[1pt] \cdashline{2-5}  &&&&\\[-8pt] 
& $\wb$*\textsubscript{\tiny{$512^2$}} & \textbf{0.5293}            & \textbf{0.5539}              & \textbf{0.4357} \\[1pt] \hline  &&&&\\[-9pt] 
\multirow{3}{*}{\raisebox{12pt}{\rotatebox[origin=c]{90}{\scriptsize{SD v3}}} }  & Baseline    & 0.8164         & 0.7303           & 0.5852         \\ \cdashline{2-5}  &&&&\\[-8pt] 
& $\wb$* & \textbf{0.8213}            & \textbf{0.7488}              & \textbf{0.5963}            \\ \hline &&&&\\[-9pt] 
\multirow{4}{*}{\raisebox{13pt}{\rotatebox[origin=c]{90}{\scriptsize{Others}}} }  & FLUX    & 0.7354         & 0.6016           & 0.4777         \\ 
& GLIGEN & 0.4288            & 0.3904               & 0.3998           \\
& RealCompo & 0.7741            & 0.7427               & 0.6032        
\end{tabular}

\vspace{-5pt}

\caption{Quantitative comparison with state-of-the-art and baseline methods across different categories of the T2I-CompBench dataset}
\label{table:comparison_with_sota}
\end{table}

\textbf{Quantitative Evaluation.}
Table~\ref{table:comparison_with_sota} presents the VQA scores for our methods, \lp{} and \wb{}, alongside the baselines discussed.
VQA scores of our method and other discussed baselines are provided in Table~\ref{table:comparison_with_sota}.
As shown, both \lp{} and \wb{} significantly improve upon the baselines. Both methods demonstrate substantial improvements over the baselines, with \wb{} achieving the highest VQA scores among state-of-the-art approaches that utilize the same baseline model, while being more computationally and parameter-efficient. Additionally, to further validate the performance gains, we evaluated our method using additional metrics, including TIFA \citep{hu2023tifaaccurateinterpretabletexttoimage}. Details of these evaluations are provided in Appendix \ref{app:sec:extended-evaluation}. The results demonstrated consistent improvements over the baselines, reinforcing the effectiveness of our approach.
For analysis of \wb{}'s robustness and generalizability—both when trained across all categories and when applied to models using T5 text encoders—see Appendix~\ref{app:sec:generalizability-of-wiclp} and \ref{app:sec:wiclp-on-t5}.

Notably, our methods maintain the model’s general utility, introducing only a slight increase in the FID score; for example, experiments on MS-COCO prompts show that while our methods slightly increase FID compared to base models, this increase is smaller than that of other baselines—for instance, $\wb{}$ achieves an FID score of $27.40$, outperforming GORS at $30.54$. Additional details on FID performance can be found in Appendix~\ref{app:sec:fid-score}.



\textbf{Human Experiments.}
We conducted a human evaluation where participants compared images generated by SD v1.4 and SD v1.4 + \wb{}, selecting the image that best matched the given prompt. The results showed that in $34.625\%$ of cases, evaluators chose the base model; in $51.875\%$, they preferred the \wb{}; and in $13.50\%$, they rated both equally. See Appendix~\ref{app:sec:human-evaluation} for further details.

\subsection{Impact of \wb{} on Subsets of Tokens}
\begin{figure}[t]
\centering
\includegraphics[width=7.5cm]{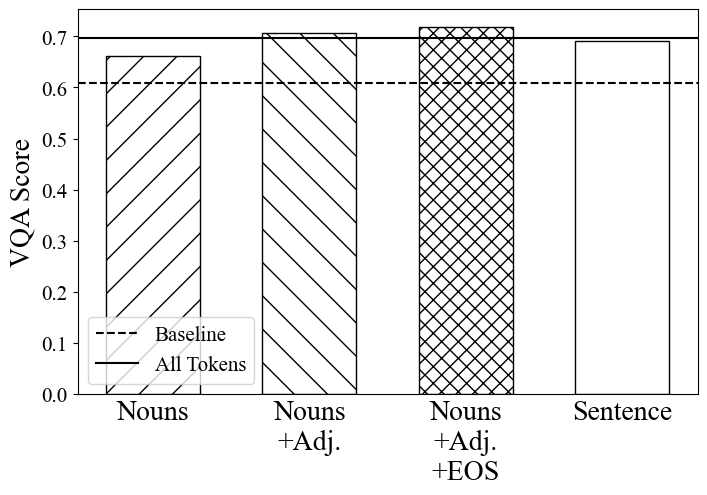}
\vspace{-10pt}
\caption{Effect of applying \wb{} to specific tokens. Applying \wb{} to a subset of tokens is sufficient to enhance compositionality, achieving comparable or superior performance to applying it across all tokens.}
\label{fig:high_impact_tokens_for_wiclp}

\includegraphics[width=\columnwidth]{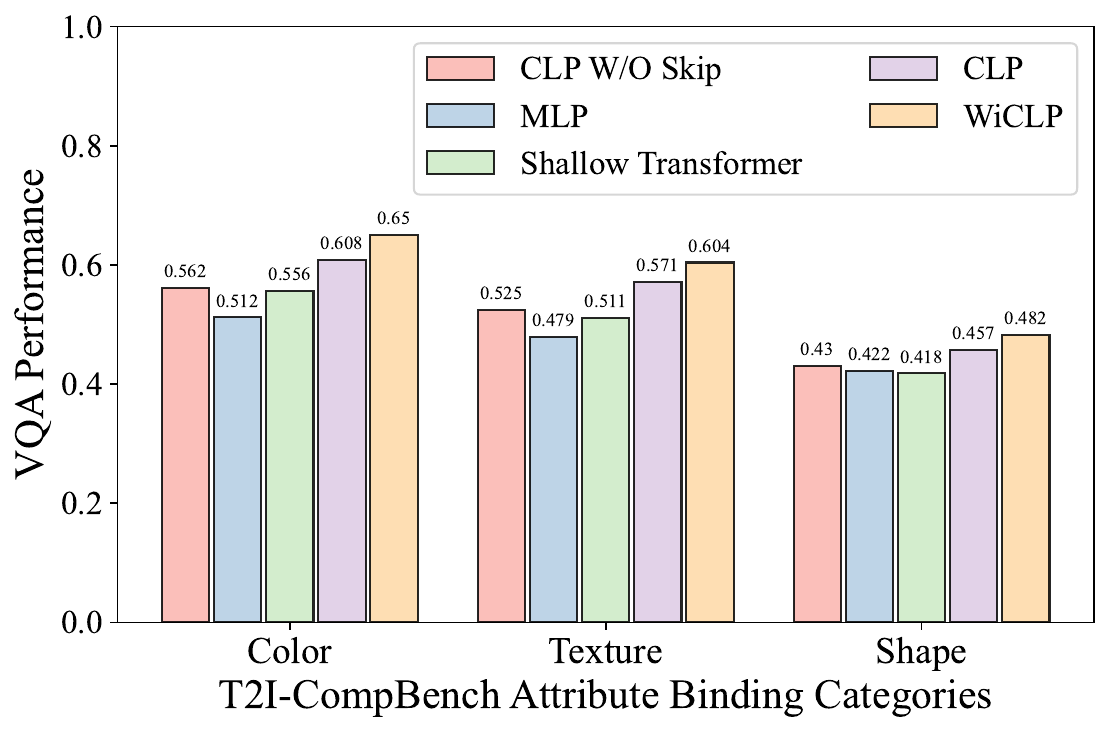}
\vspace{-20pt}
\caption{Performance comparison of different projection architectures on T2I-CompBench.}
\label{fig:projection_comparison_1}
\end{figure}

To better understand the impact of \wb{} on token embeddings, we applied the trained \wb{} to specific subsets of tokens from a sample of dataset sentences. The results, shown in Fig.~\ref{fig:high_impact_tokens_for_wiclp}, compare the following token groups: nouns only; nouns and adjectives; nouns, adjectives, and the EOS (End of Sentence token) token; all sentence tokens; and all tokens outputted by CLIP (sentence tokens plus padding tokens). As can be seen, applying \wb{} only to a small number of tokens is sufficient for improving compositionality. Interestingly, applying \wb{} to the group of nouns, adjectives, and EOS achieves even higher VQA scores than applying \wb{} to all tokens. Despite these findings, we applied \wb{} to all tokens in our work, leaving this targeted approach for future research.

\subsection{Alternatives to \wb{}}
We explored various fine-tuning strategies for improving CLIP, including fine-tuning the entire CLIP, fine-tuning only the last layers of CLIP combined with \wb{}, and using \wb{} alone. Our results show that the original baseline model (SD v1.4) achieves a VQA score of 0.3765 on the color category of the dataset. Fine-tuning the entire CLIP without \wb{} improves the score to 0.5173, fine-tuning the last layers of CLIP combined with \wb{} achieves 0.5497, and \wb{} alone achieves 0.5383.

These findings highlight the effectiveness of \wb{}, which outperforms full fine-tuning of CLIP while being significantly more parameter-efficient. While fine-tuning the last layers of CLIP combined with \wb{} achieves slightly better performance than using \wb{} alone, it requires optimizing a much larger number of parameters. Given this trade-off, we prioritize \wb{} alone to minimize the number of parameters while achieving substantial compositional performance improvements. Additionally, keeping the original CLIP unchanged makes our approach more suitable for \earlystop{} functionality, allowing the module to be easily enabled or disabled as needed.

Additionally, we conducted an ablation study on various projection layer architectures, ranging from simple designs to more advanced, parameter-heavy transformer-based models. For detailed descriptions of these architectures, refer to Appendix \ref{app:ablation_proj}. Our evaluation across all attribute binding categories of T2I-CompBench, as shown in Figure \ref{fig:projection_comparison_1}, reveals that \lp{} and \wb{} consistently achieve the highest performance gains while remaining both parameter- and speed-efficient. For further results and analysis, please refer to Appendix \ref{app:ablation_proj}.




\section{Conclusion}
We analyze error sources in text-to-image models for generating images from compositional prompts, identifying (i) erroneous attention contributions in CLIP token embeddings and (ii) the CLIP text-encoder’s sub-optimal alignment with the UNet. Based on these insights, we propose \wb{}, a simple yet strong baseline that fine-tunes a linear projection on CLIP’s representation space. \wb{} though inherently simple and parameter efficient, outperforms existing methods on compositional image generation benchmarks and maintains a low FID score on a broader range of clean prompts. 

\section*{Limitations} \label{app:sec:limit}
In this paper, we have conducted a comprehensive analysis of one of the primary reasons why Stable Diffusion struggles with generating compositional attribute binding prompts and proposed a lightweight, efficient method to address this challenge. While our approach demonstrates promising results, there remains substantial room for improvement in this area. Our method primarily targets the attribute binding aspect of compositionality, leaving other critical categories, such as spatial relationships (e.g., "a book to the left of a pen"), numeracy (e.g., "four books"), and others, less explored. Investigating the underlying causes of these issues is crucial for advancing the field further.

Moreover, the reliance on CLIP—particularly the CLIP score—as a metric for recognizing and evaluating compositionality poses its own limitations. CLIP, in its current form, does not perform optimally for such tasks. A promising direction for future research would be to first improve CLIP's ability to handle compositionality effectively and then adapt this enhanced version of CLIP for Stable Diffusion. This could pave the way for more robust and accurate text-to-image generation models.

\section*{Acknowledgement}
This project was supported in part by a grant from an NSF CAREER AWARD 1942230, ONR YIP award N00014-22-1-2271, ARO’s Early Career Program Award 310902-00001, Army Grant No. W911NF2120076, the NSF award CCF2212458, NSF Award No. 2229885 (NSF Institute for Trustworthy AI in Law and Society, TRAILS), an Amazon Research Award and an award from Capital One.

\bibliography{ref}

\appendix

\newpage

\section{Optimizing the Text-embeddings of a Subset of Tokens}
\label{app:sec:optimize}

\begin{figure}[t]
\centering
\includegraphics[width=7cm]{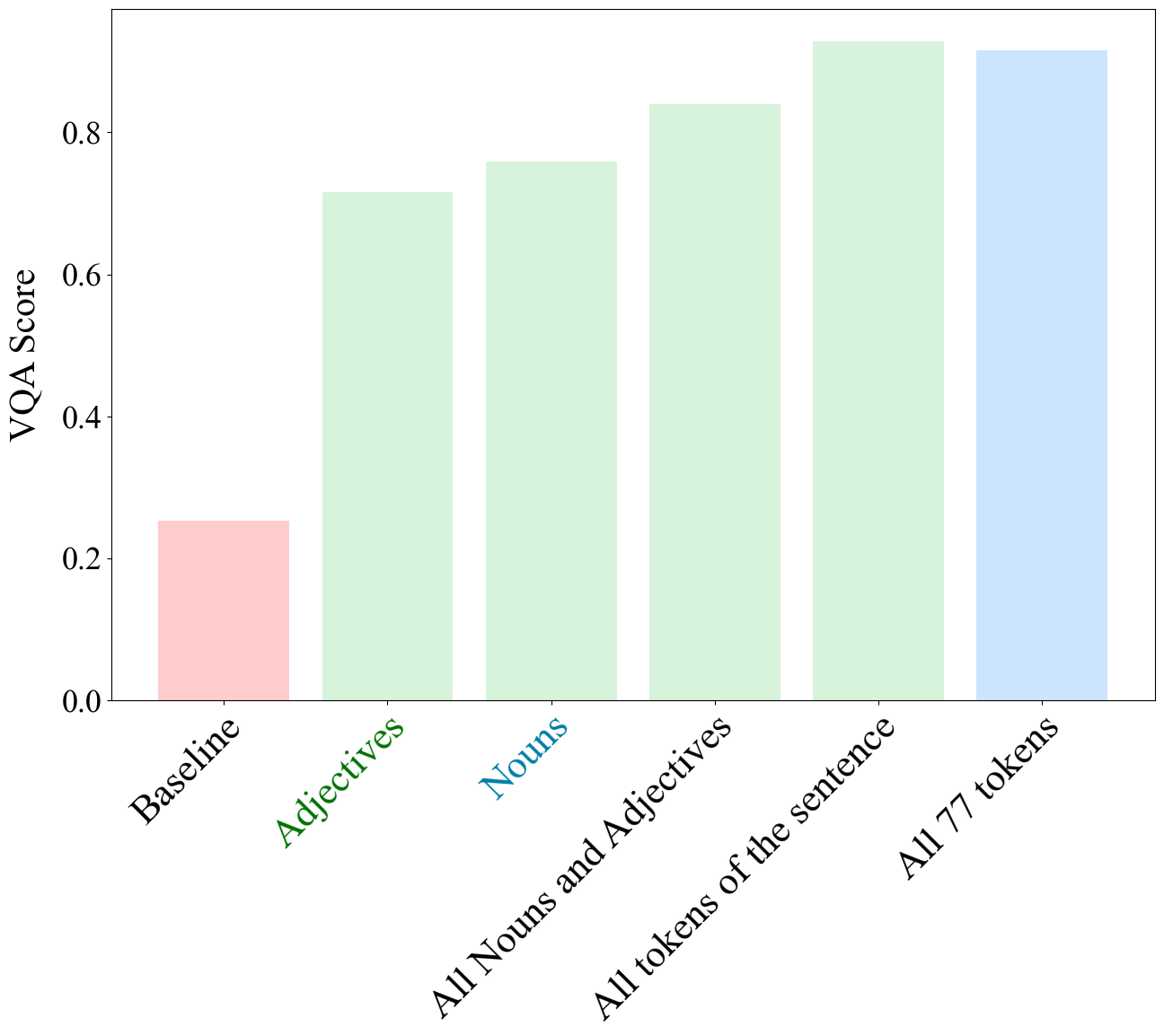}

\definecolor{adj_color}{HTML}{007200}
\definecolor{noun_color}{HTML}{007ea7}

\caption{
Comparison of VQA scores when optimizing different subsets of tokens for the sample prompt: ''A \textcolor{adj_color}{red} \textcolor{noun_color}{book} and a \textcolor{adj_color}{yellow} \textcolor{noun_color}{vase}''}
\label{fig:opt_some_tokens}
\end{figure}

Given $\vb{c} \in \mathbb{R}^{n\times d}$, where $n$ refers to the number of tokens and $d$ refers to the dimensionality of the text-embedding, for the second configuration we only optimize a subset of tokens $n' \in n$. We refer to this subset of tokens as $\vb{c}'$. These tokens correspond to relevant parts of the prompt which govern compositionality (e.g., ``red book'' and ``yellow table'' in ``A red book and an yellow table''). 
\begin{align*}
    \vb{c}'^{*} = \arg \min_{\vb{c}'} \mathbb{E}_{\epsilon, t} || \epsilon - \epsilon_{\theta}(\vb{z}_{t}, \vb{c}', t) ||_{2}^{2},
\end{align*}

\begin{figure*}[t]
    \centering
    \includegraphics[width=\textwidth]{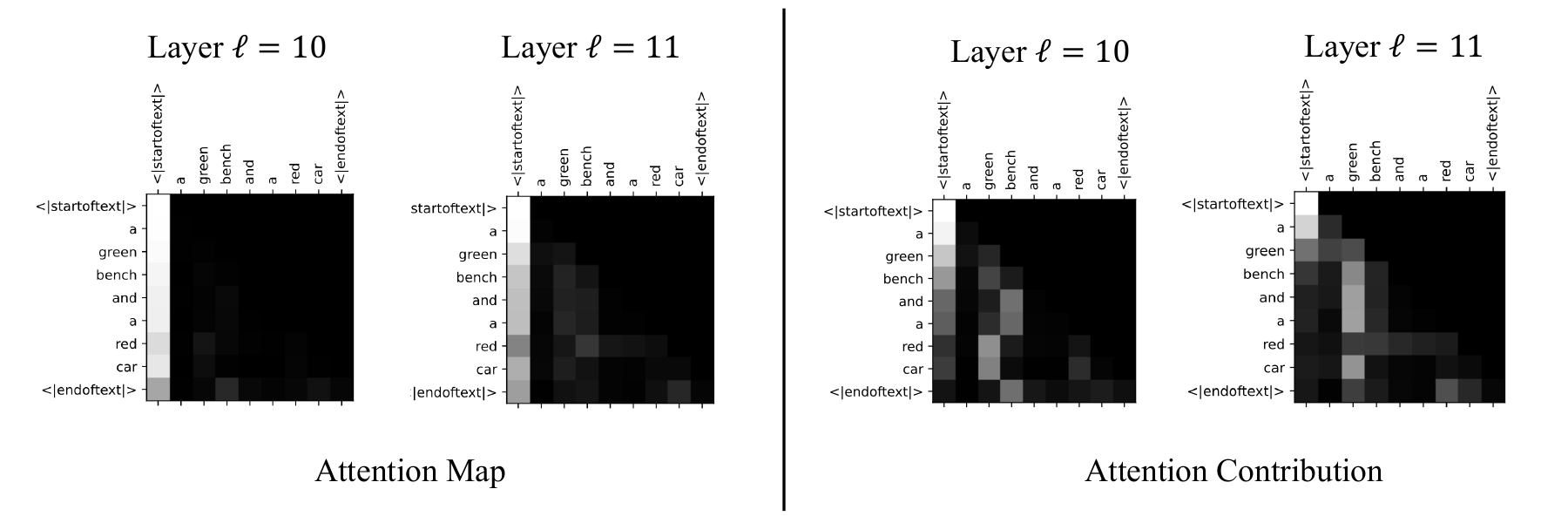}
    \caption{Visualization of attention map and attention contribution for prompt ``a green bench and a red car" over different layers of CLIP. Contribution provides better insight on the attention mechanism.}
    \label{fig:app:attn-map}
\end{figure*}

Figure \ref{fig:opt_some_tokens} shows the results for the sample prompt "a red book and a yellow vase". We considered different subsets of tokens $n'$: adjectives ("red" and "yellow"), nouns ("book" and "vase"), both nouns and adjectives, and all tokens in the sentence. The results indicate that optimizing even a few tokens significantly improves the VQA score. However, optimizing all tokens in the sentence yields the highest score.

\section{Source (i) : Erroneous Attention Contributions}
\label{app:sec:error-one}

\subsection{Attention Contribution}
\label{app:sec:cont}

In this Section, we provide more details on our analysis to quantitatively measure tokens' contribution to each other in a~layer of attention mechanism. One natural way of doing this analysis is to utilize attention maps $\attn_{i, j}^h$ and aggregate them over heads, however, we observe that this map couldn't effectively show the contribution.
Attention map does not consider norm of tokens in the previous layer, thus, does not provide informative knowledge on how each token is formed in the attention mechanism. In fact, as seen in Figure~\ref{fig:app:attn-map}, we cannot obtain much information by looking at these maps while attention contribution clearly shows amount of norm that comes from each of the attended tokens.

\begin{figure}[t]
\centering
\includegraphics[width=0.95\columnwidth]{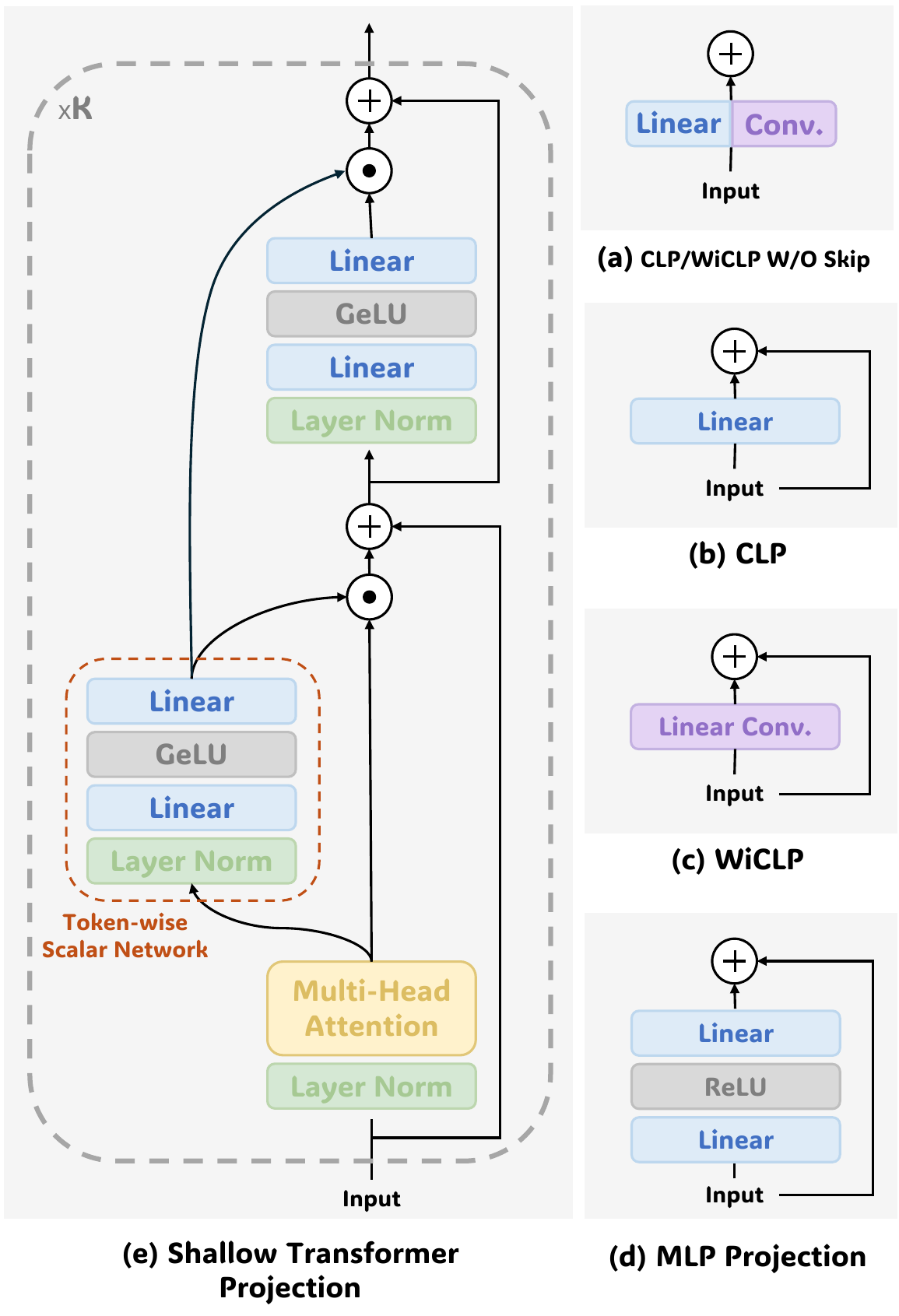}
\caption{Overview of different architectural designs for the projection layer. Each design varies in structure and functionality, influencing how input features are transformed and integrated into the model.}
\label{fig:projection_designs}
\end{figure}

\begin{figure}[t]
\centering
\includegraphics[width=\columnwidth]{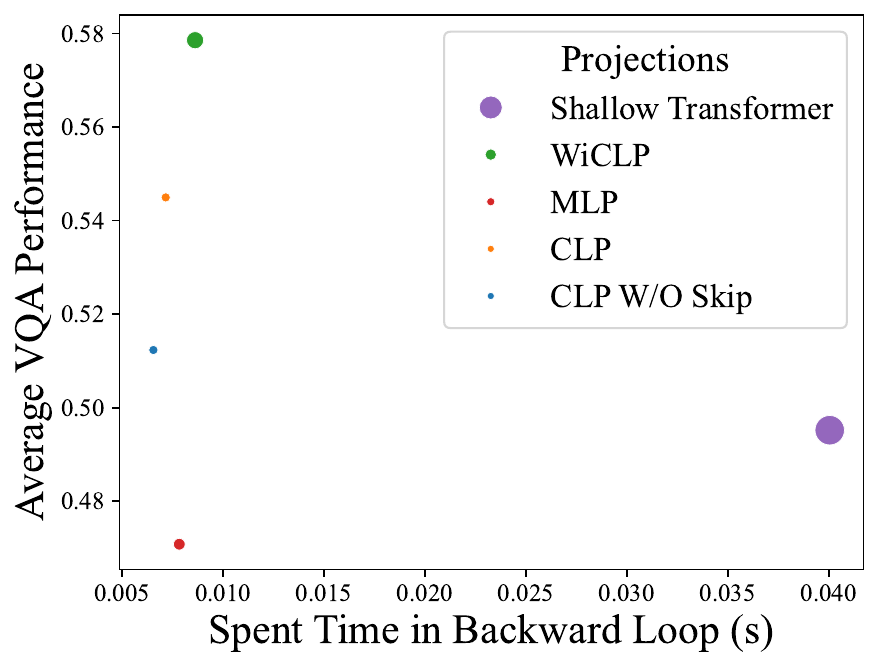}
\caption{Comparison of performance, speed, and parameter efficiency across different projection architectures. Each projection is represented by a different color, with marker size indicating its relative number of parameters. The x-axis represents the time spent during the backward pass, while the y-axis shows the average performance across the three attribute binding categories: color, texture, and shape.}
\label{fig:projection_comparison_2}
\end{figure}

\subsection{Zero-shot Attention Reweighting}
\label{app:sec:zero-shot}

Inspired by attention mechanism shortcomings of CLIP text-encoder, we aim to improve compositionality of CLIP-based diffusion models by zero-shot reweighting of the attention maps. Specifically, we apply a hand-crafted zero-shot manipulation of the attention maps in certain layers of the CLIP text-encoder to effectively reduce unintended attentions while enhancing meaningful ones. This zero-shot reweighting is applied to the logits before the \textsc{Softmax} layer in the last three layers of the text-encoder.
More precisely, we compute a matrix $M \in \mathbb{R}^{n \times n}$ and add it to the attention logits. For each head $h$, the new attention values are computed and then propagated through the subsequent layers of the text encoder:
\begin{align*}
    \attn_{i, .}^{'h} =
    \textsc{Softmax}\left( \left\{
    \frac{
        \langle \atq_i^{h}, \atk_j^{h}\rangle 
    }
    {
    \sqrt{d_h}
    }
    + M_{i, j}
    \right\}_{j=1}^{n}
    \right).
\end{align*}

We set the values in $M$ by considering the ideal case where no incorrect attentions occur in the mechanism.
For example, for prompt ``a green bench and a red car", we ensure that the token "car" does not attend to the token "green" by assigning a sufficiently large negative value to the corresponding entry in matrix $M$.

\begin{figure}
    \centering
    \includegraphics[width=0.5\textwidth]{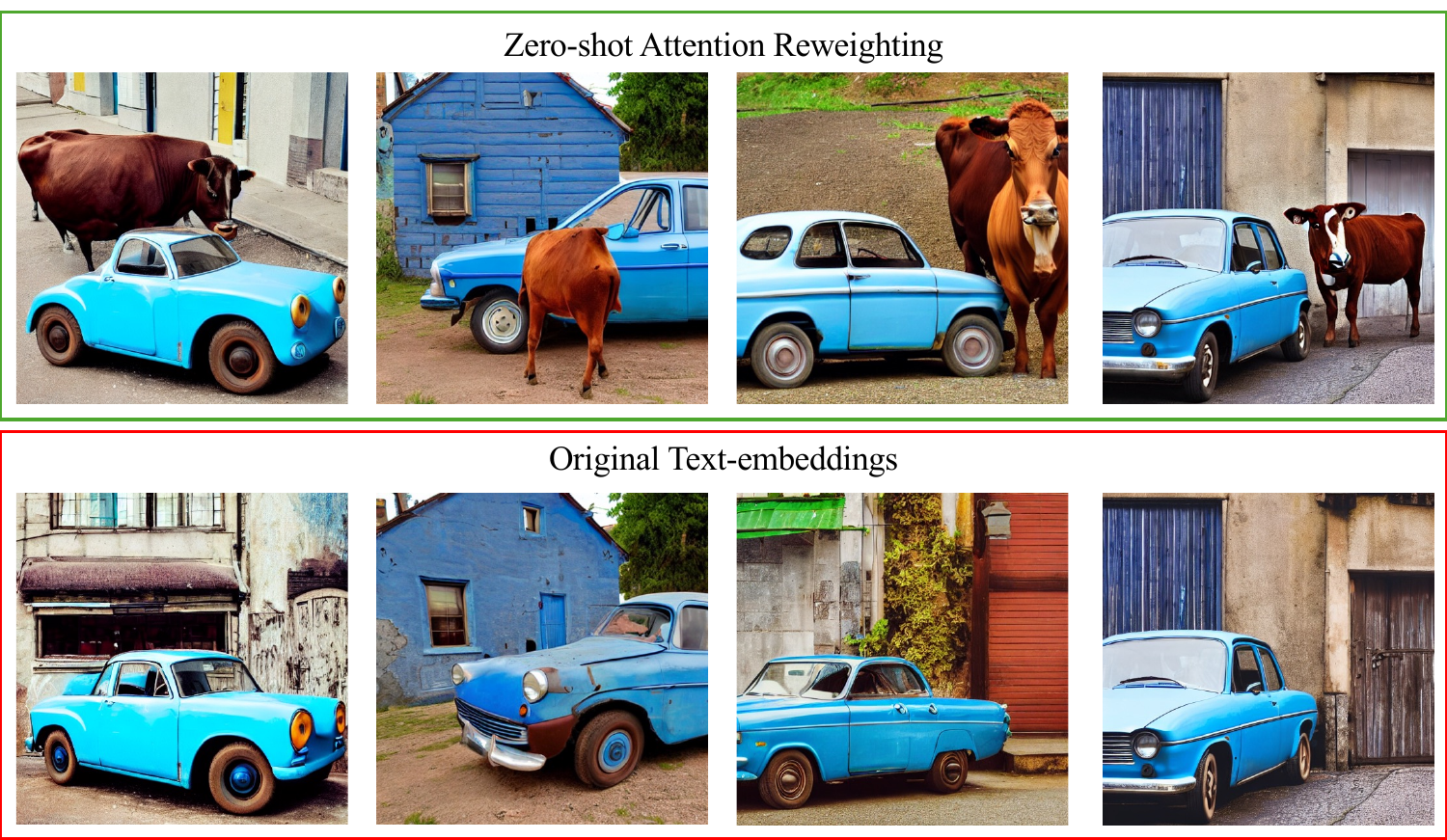}
    \caption{ Visualization of some images generated with same set of seeds using original text-embeddings of prompt ``a blue car and a brown cow" and text-embeddings that are obtained as the result of zero-shot reweighting of attention matrix.}
    \label{fig:zero-shot}
\end{figure}

To fix unintended attentions, we aim to compute a matrix $M$ to be applied across various heads in the last few layers of CLIP, reducing the effect of wrong attention, leading to more accurate text-embeddings that are capable of generating high-quality compositional scenes. To avoid unintended attention for prompts of the form ``$\vb{a}_1 \vb{o}_1 + \vb{a}_2 \vb{o}_2$", we add large negative values to entries $M_{\vb{o}_2, \vb{a}_1}$, $M_{\vb{a}_2, \vb{a}_1}$, and some positive value to $M_{\vb{o}_2, \vb{a}_2}$ and $M_{\vb{o}_1, \vb{a}_1}$, and small negative value to $M_{\vb{o}_2, \vb{o}_1}$. To find what values to assign to those entries, we consider a small set of prompts in color dataset ($5$ prompts in total) and obtain parameters for that matrix to maximize VQA score. Figure~\ref{fig:zero-shot} shows few examples of zero-shot modification.

Applying zero-shot attention reweighting with matrix $M$ on $780$ compositional prompts of the color category of T2I-CompBench dataset, we achieved a $2.93\%$ improvement in VQA scores.

\subsection{Experiments with LLaMa3 8B}
\label{app:llama}


\begin{table}[]
\centering
\begin{tabular}{cc|c|c}
                 & & LLaMa3                & CLIP                 \\ \hline \hline
\multirow{2}{*}{color} & last layer  & 0.015                & 0.657                \\ \cline{2-4}
& all layers  & 0.081                & 0.187                \\ \hline
\multirow{2}{*}{texture} & last layer & {\color[HTML]{333333} 0.033} & 0.696                \\ \cline{2-4}
& all layers & {\color[HTML]{333333} 0.066} & 0.213               
\end{tabular}[t]
\caption{Unintended attention rate in LLaMa3 8B vs CLIP. LLaMa3 shows significant less unintended attentions.}
\label{tab:llama}
\end{table}


\begin{table}[t]
\centering
\renewcommand{\arraystretch}{1.2}
\begin{tabular}{c|c}
                      & FID Score \\ \hline \hline
SD v1.4               & 24.33     \\ \cline{2-2}
SD v1.4 + \wb{}       & 25.40     \\ \hline
SD v2                 & 23.27     \\ \cline{2-2}
SD v2 + \wb{}         & 27.40     \\ \cline{2-2}
GORS                  & 30.54     
\end{tabular}
\caption{Comparison of FID scores between the baseline models and \wb{} using \earlystop{} with $\tau = 800$, as well as the GORS approach.}
\label{tab:fid-comparison}
\end{table}

We explored the analysis of attention contributions to identify unintended attention in LLaMa3 8B, which utilizes a more advanced text encoder specifically designed for language modeling and pretrained on large-scale text corpora. Table~\ref{tab:llama} reports the rate of unintended attention across prompts in the color and texture datasets. The results demonstrate that unintended attention occurs less frequently in more advanced text encoders, further emphasizing the limitations of the CLIP text encoder.

\subsection{Models with T5 text-encoder}
\label{app:t5}
We conducted experiments to measure the VQA score on the color dataset for models that use T5 as their text encoder. DeepFloyd achieved a score of $0.604$, which is significantly higher than that of SD-v1.4. Additionally, DeepFloyd-I-M, which employs a smaller first-stage UNet compared to DeepFloyd, obtained a score of $0.436$, also surpassing the SD-v1.4 score.

\section{Ablation on the Projection Layer Architecture}
\label{app:ablation_proj}
\newcommand{\shallowtransf}{\texttt{{STP}}}

In this section, we discuss the architectural choices we considered for the projection layer. As illustrated in Fig. \ref{fig:projection_designs}, we experimented with five different architectures, ranging from simple linear networks to more advanced attention-based models.

The simplest architecture, shown in Fig. \ref{fig:projection_designs} (a), consists of a single linear or windowed linear convolutional layer that transforms the input into a new embedding. Building on this, we introduced a residual connection from the input to the output, resulting in the architectures depicted in Fig. \ref{fig:projection_designs} (b) and (c). These correspond to \lp{} and \wb{}, which we discuss in detail in Section \ref{sec:projection_layer}. Additionally, we explored the effect of incorporating non-linearity into these projection layers, as shown in Fig. \ref{fig:projection_designs} (d).

To design a more sophisticated architecture, we developed an attention-based projection model, which we refer to as the Shallow Transformer Projection (\shallowtransf{}). \shallowtransf{} consists of stacked attention blocks similar to those in transformer architectures but with a key difference: we introduce a Token-wise Scalar Network that dynamically controls the extent to which each token’s representation is influenced by the projection block (see Fig. \ref{fig:projection_designs} (e)).

For a comprehensive comparison, we evaluated these projection models across various diffusion model variants, ranging from SD v1.4 and SD v2 to more recent models like SDXL. While all projection methods improved upon the baseline, \lp{} and \wb{} demonstrated the most noticeable gains despite their parameter efficiency. This highlights that \lp{} and \wb{} not only offer a lightweight and computationally efficient solution but also achieve superior compositional improvements.

To ensure robust and generalizable quantitative results, we evaluated different projection architectures across all attribute binding categories of T2I-CompBench—color, texture, and shape—on SD v2. The results, presented in Figure \ref{fig:projection_comparison_1}, demonstrate that \lp{} and \wb{} achieve significantly higher performance gains while maintaining parameter and speed efficiency. Additionally, Figure \ref{fig:projection_comparison_2} compares the projections in terms of performance, inference speed, and parameter count. Notably, increasing model complexity, such as with the Shallow Transformer projection, does not necessarily lead to better performance. In contrast, \lp{} and \wb{} strike an optimal balance, offering superior compositional attribute binding while remaining highly efficient in both speed and parameter usage.

\section{Experiments}

\subsection{Training setup}
\label{app:sec:train_setup}
In this section, we present the details of the experiments conducted to evaluate our proposed methods. The training is performed for 25,000 steps with a batch size of 4. An RTX A5000 GPU is used for training models based on Stable Diffusion 1.4, while an RTX A6000 GPU is used for models based on Stable Diffusion 2. We employed the Adam optimizer with a learning rate of $1 \times 10^{-5}$ and utilized a Multi-Step learning rate scheduler with decays ($\alpha=0.1$) at 10,000 and 16,000 steps. For the \wb, a window size of 5 was used. All network parameters were initialized to zero, leveraging the skip connection to ensure that the initial output matched the CLIP text embeddings. Our implementation is based on the Diffusers\footnote{https://github.com/huggingface/diffusers} library, utilizing their modules, models, and checkpoints to build and train our models. This comprehensive setup ensured that our method was rigorously tested under controlled conditions, providing a robust evaluation of its performance.

\subsection{Extended Evaluation}
    \label{app:sec:extended-evaluation}

\begin{figure*}
\centering
\includegraphics[width=11cm]{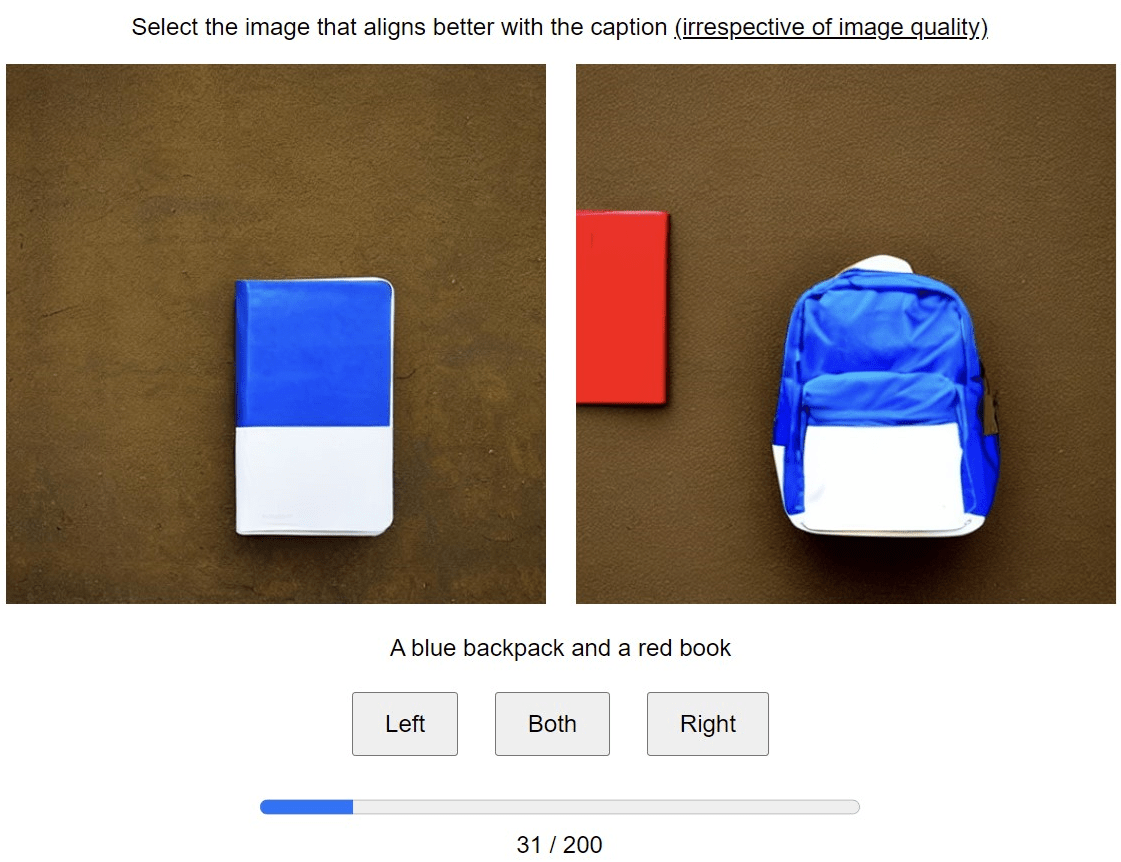}

\definecolor{adj_color}{HTML}{007200}
\definecolor{noun_color}{HTML}{007ea7}

\caption{
A sample from the human evaluation study, where participants were presented with a pair of images and a caption. They were asked to select the image that best represented the caption or choose 'both' if the images equally captured the caption's meaning.
}
\label{fig:human-evaluation}
\end{figure*}

\textbf{Human Evaluation}
    \label{app:sec:human-evaluation}
We conducted a human evaluation in which participants compared images generated by SD v1.4 and SD v1.4 + \wb{}, selecting the image that best matched the given prompt (Figure \ref{fig:human-evaluation}). Five evaluators were presented with 200 randomly selected image pairs, evaluating a total of $1000$ image-caption pairs.

\textbf{TIFA Metric.} To provide a more comprehensive evaluation, in addition to the disentangled BLIP-VQA score proposed by \citep{huang2023t2icompbench}, we also incorporate the TIFA metric \citep{hu2023tifaaccurateinterpretabletexttoimage}. TIFA (Text-to-Image Faithfulness Evaluation with Question Answering) is an automated evaluation method that measures how faithfully a generated image corresponds to its textual input via visual question answering (VQA). It generates multiple question-answer pairs from the text input using a language model, then evaluates image faithfulness by determining whether existing VQA models can accurately answer these questions based on the image. As a reference-free metric, TIFA offers fine-grained and interpretable assessments of image quality.

Using TIFA, we observed that SD v1.4 and SD v2 achieved scores of $0.6598$ and $0.7735$, respectively. Notably, the scores for $\wb{}$ applied on top of SD v1.4 and SD v2 improved to $0.7462$ and $0.8133$, respectively, demonstrating the enhanced performance of our approach.

\textbf{FID Score Comparison}
\label{app:sec:fid-score}
Our method results in a modest increase in FID score on MS-COCO prompts compared to the base models, as shown in Table \ref{tab:fid-comparison}. However, this increase is less pronounced than in other baselines—for example, SD v2 + $\wb{}$ scores 27.40, whereas GORS reaches 30.54.

\subsection{Robustness and Generalizability of \wb{}}
    \label{app:sec:generalizability-of-wiclp}
To validate the robustness and generalizability of our method, we trained \wb{} across all categories simultaneously using the same setup as when training \wb{} on individual categories separately. For SD v2, this resulted in VQA scores of 0.6311 for color, 0.5728 for texture, and 0.4620 for shape—a slight ($\scriptstyle\sim2\%$) decrease compared to training on individual categories. Despite this minor drop, the model still significantly outperforms the baseline, demonstrating its strong generalizability and robustness.

Additionally, to assess the generalizability of \wb{} to other text encoders, we conducted experiments on PixArt-$\alpha$ and DeepFloyd, both of which use a T5 text encoder. The results, presented in Table \ref{table:comparison_with_sota} and Appendix \ref{app:sec:wiclp-on-t5}, demonstrate that \wb{} is highly effective for these models, further highlighting its robustness and broad applicability.

\subsection{Generalization of \wb{} to T5-Based Encoders}
    \label{app:sec:wiclp-on-t5}
    
To assess WiCLP's applicability to text encoders beyond CLIP, we tested it on DeepFloyd-I-M and PixArt-$\alpha$, both of which use T5-XXL as their text encoder. Our results confirm that the linear head tuning approach remains effective, yielding significant improvements across different architectures.

To establish that these models have room for improvement, we conducted interpretability experiments as outlined in Section \ref{sec:optimized-text-embeddings}. For instance, with the prompt “a red book and a yellow vase”, the baseline DeepFloyd-I-M model achieved a BLIP-VQA score of 0.4350. By optimizing the text embedding space with our method, this score significantly increased to 0.9121, demonstrating that the T5 text encoder’s output space can be further refined for better compositional generation in these models.

Table \ref{table:comparison_with_sota} presents the results of applying \wb{} to PixArt-$\alpha$, showing noticeable improvements over the baseline. Similarly, applying WiCLP to DeepFloyd-I-M enhanced BLIP-VQA performance across all 300 evaluation prompts, increasing the score from 0.4636 to 0.5155. These results further reinforce WiCLP’s generalizability and effectiveness across diverse text encoders.

\subsection{\lp{} and \wb{} Visualization}
    \label{app:sec:vis-lp-wb}

In this section, we provide additional visualizations comparing \lp{}, \wb{}, and baseline models in Figures \ref{fig:attention_visualization_of_linear_projection_appendix_1}, \ref{fig:attention_visualization_of_linear_projection_appendix_2}.

\subsection{Visualization of Cross-Attentions}
\label{app:sec:cross-attn}

\begin{figure*}[]
\centering
\setlength{\tabcolsep}{1pt}
\begin{tabular}{cccccc}
                  & \multicolumn{5}{c}{prompt: ''A red book and a yellow vase''} \\
                  & image    & red         & book         & yellow         & vase         \\
\raisebox{35pt}{Baseline} &
\includegraphics[width=2.3cm]{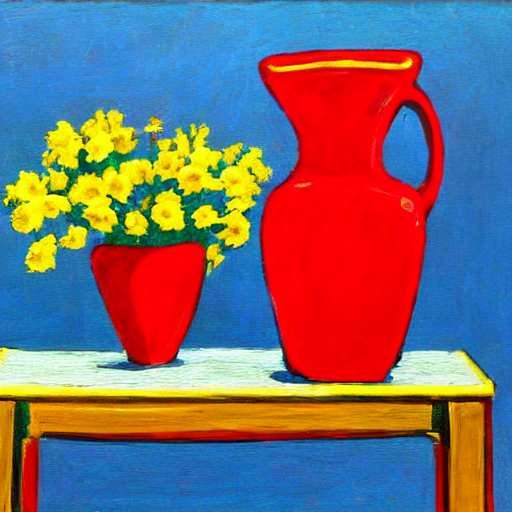} & 
\includegraphics[width=2.3cm]{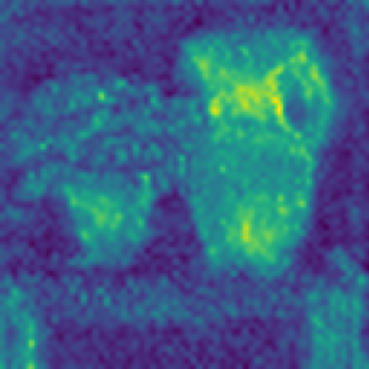} &
\includegraphics[width=2.3cm]{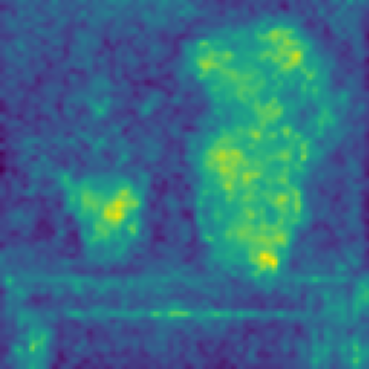} &
\includegraphics[width=2.3cm]{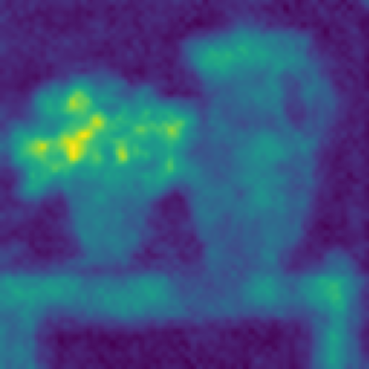} &
\includegraphics[width=2.3cm]{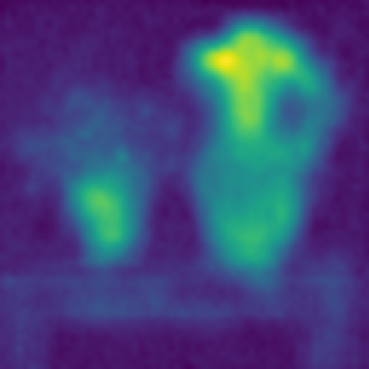} \\
\raisebox{35pt}{\makecell{Linear \\ Projection}} &
\includegraphics[width=2.3cm]{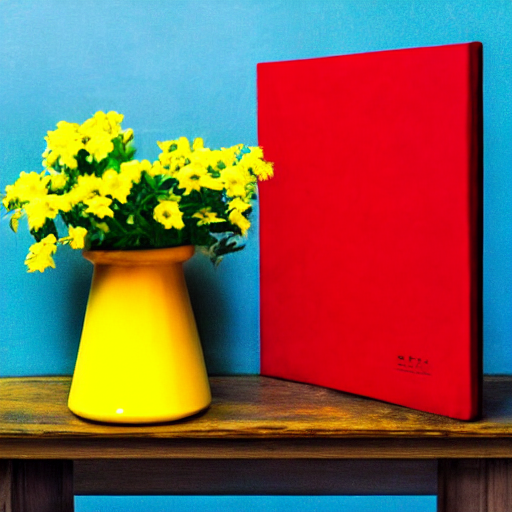} & 
\includegraphics[width=2.3cm]{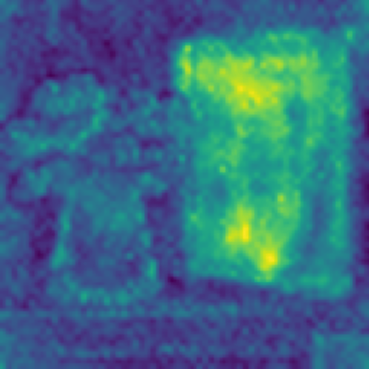} &
\includegraphics[width=2.3cm]{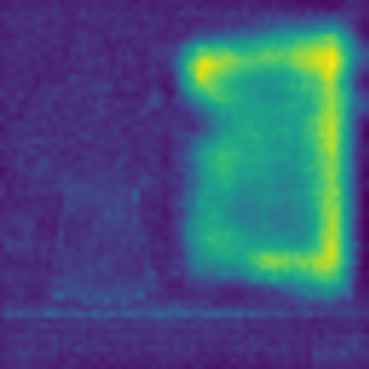} &
\includegraphics[width=2.3cm]{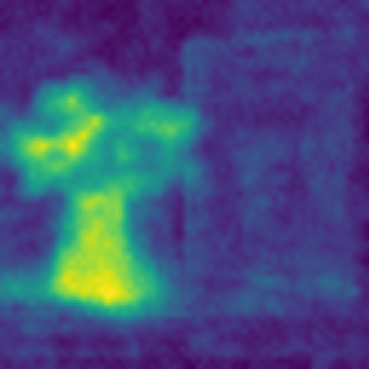} &
\includegraphics[width=2.3cm]{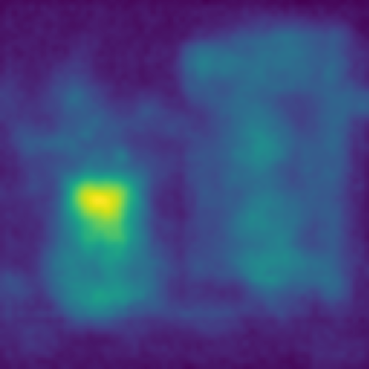} \\ 
\rule[4ex]{0pt}{0pt}
                  & \multicolumn{5}{c}{prompt: ''A blue backpack and a red bench''} \\
                  & image    & blue         & backpack         & red         & bench         \\
\raisebox{35pt}{Baseline} &
\includegraphics[width=2.3cm]{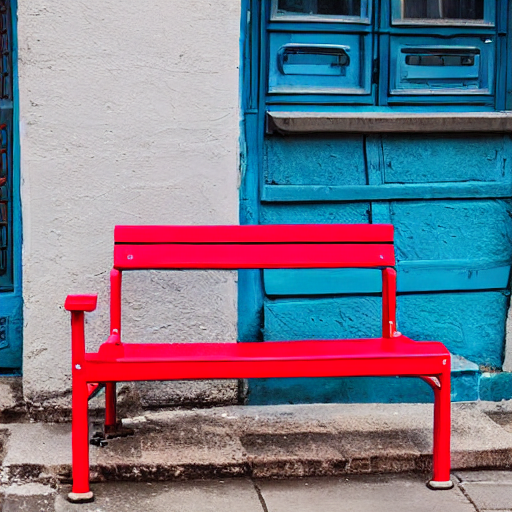} & 
\includegraphics[width=2.3cm]{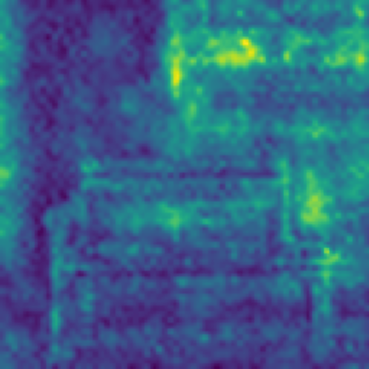} &
\includegraphics[width=2.3cm]{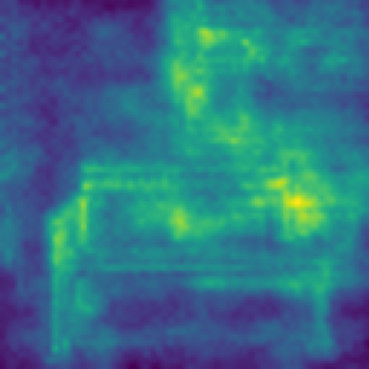} &
\includegraphics[width=2.3cm]{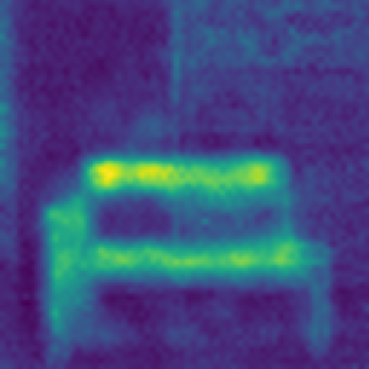} &
\includegraphics[width=2.3cm]{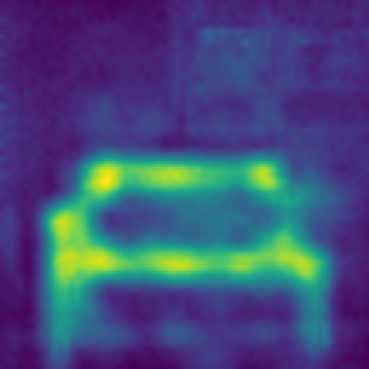} \\
\raisebox{35pt}{\makecell{Linear \\ Projection}} &
\includegraphics[width=2.3cm]{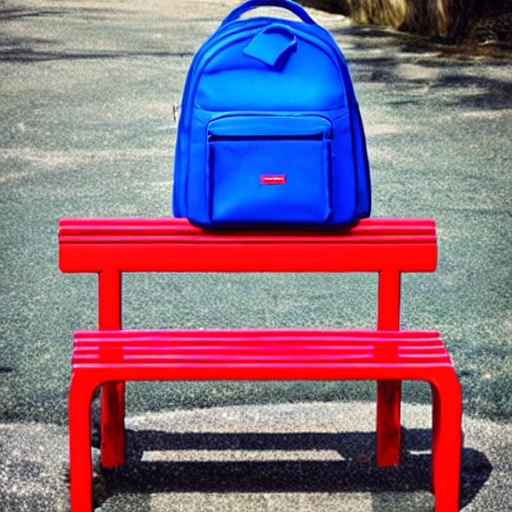} & 
\includegraphics[width=2.3cm]{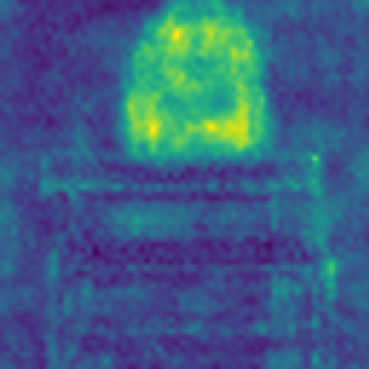} &
\includegraphics[width=2.3cm]{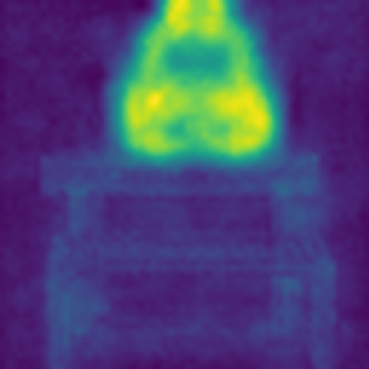} &
\includegraphics[width=2.3cm]{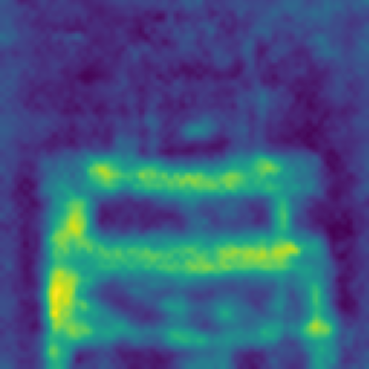} &
\includegraphics[width=2.3cm]{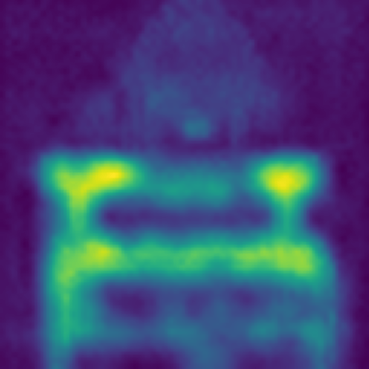} \\
\rule[4ex]{0pt}{0pt}
                  & \multicolumn{5}{c}{prompt: ''A brown boat and a blue cat''} \\
                  & image    & brown         & boat         & blue         & cat         \\
\raisebox{35pt}{Baseline} &
\includegraphics[width=2.3cm]{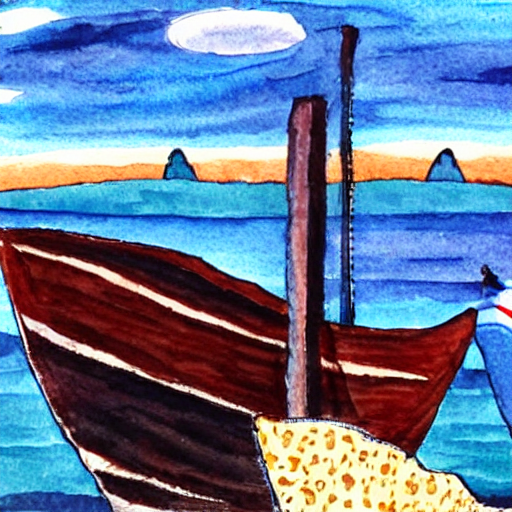} & 
\includegraphics[width=2.3cm]{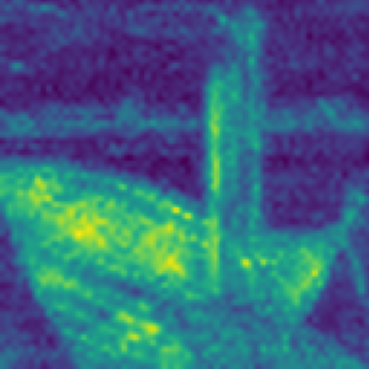} &
\includegraphics[width=2.3cm]{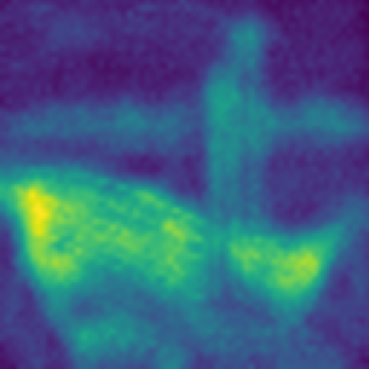} &
\includegraphics[width=2.3cm]{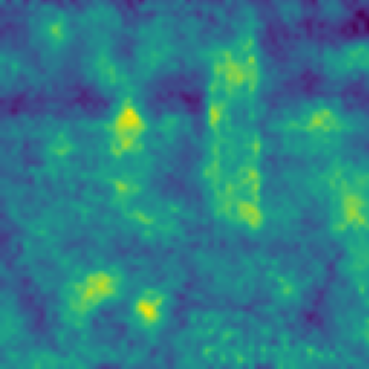} &
\includegraphics[width=2.3cm]{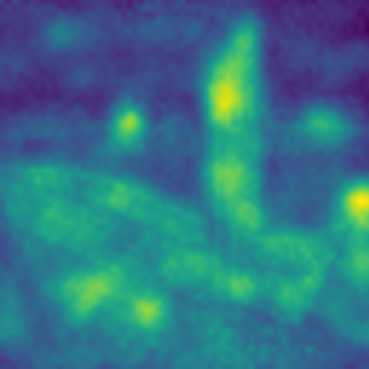} \\
\raisebox{35pt}{\makecell{Linear \\ Projection}} &
\includegraphics[width=2.3cm]{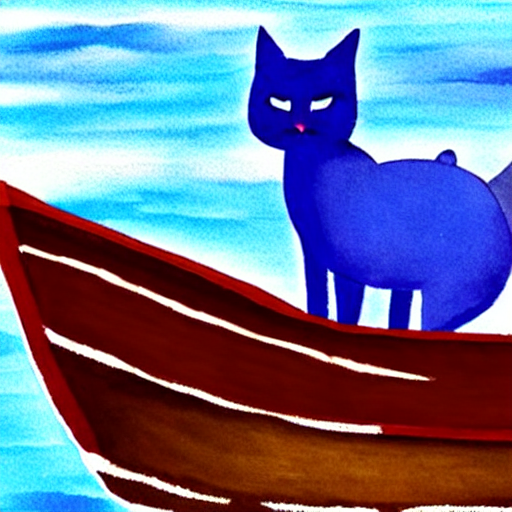} & 
\includegraphics[width=2.3cm]{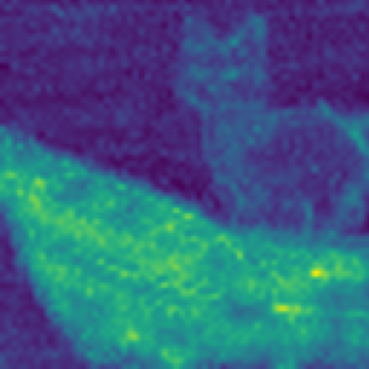} &
\includegraphics[width=2.3cm]{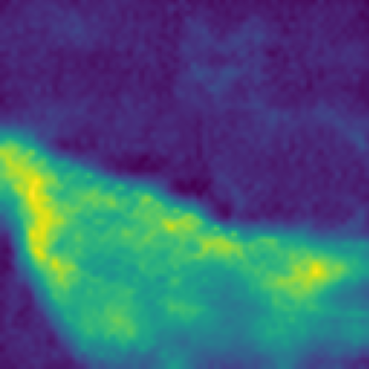} &
\includegraphics[width=2.3cm]{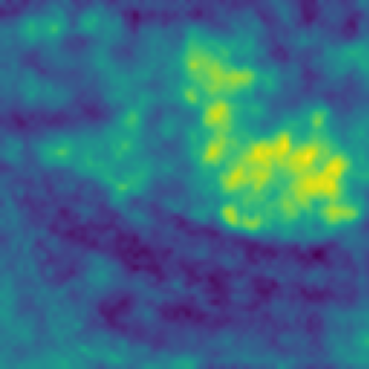} &
\includegraphics[width=2.3cm]{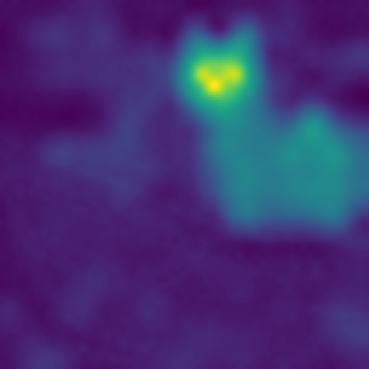} \\
\end{tabular}

\vspace{10pt}

\caption{Comparison of cross-attention maps of the U-Net with and without the \lp}
\label{fig:attention_visualization_of_linear_projection_appendix_1}
\end{figure*}

\begin{figure*}[]
\centering
\setlength{\tabcolsep}{1pt}
\begin{tabular}{cccccc}
                  & \multicolumn{5}{c}{prompt: ''A green blanket and a blue pillow''} \\
                  & image    & green         & blanket         & blue         & pillow         \\
\raisebox{35pt}{Baseline} &
\includegraphics[width=2.3cm]{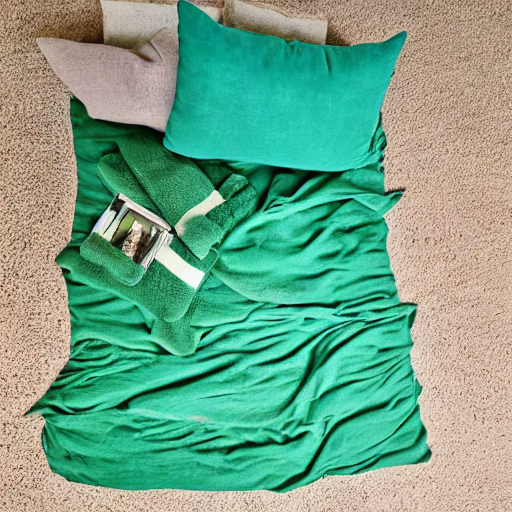} & 
\includegraphics[width=2.3cm]{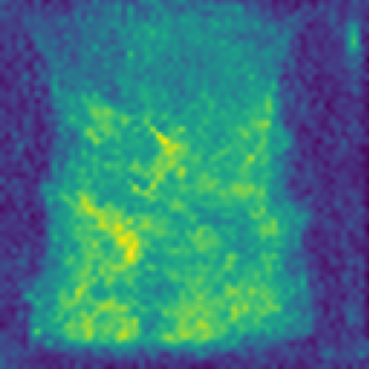} &
\includegraphics[width=2.3cm]{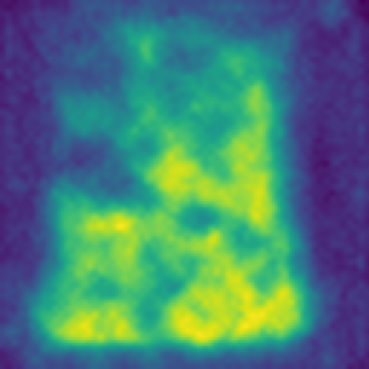} &
\includegraphics[width=2.3cm]{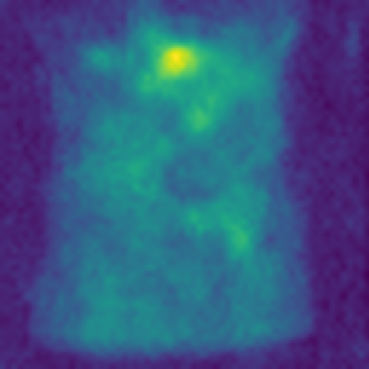} &
\includegraphics[width=2.3cm]{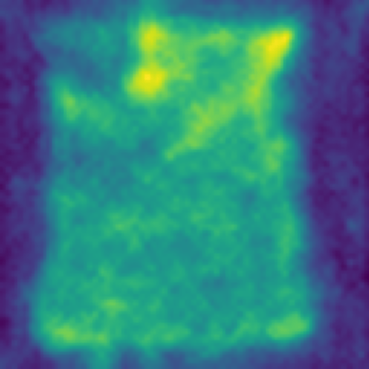} \\
\raisebox{35pt}{\makecell{Linear \\ Projection}} &
\includegraphics[width=2.3cm]{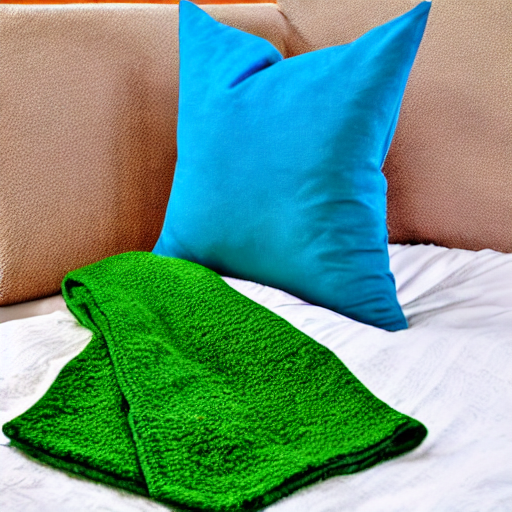} & 
\includegraphics[width=2.3cm]{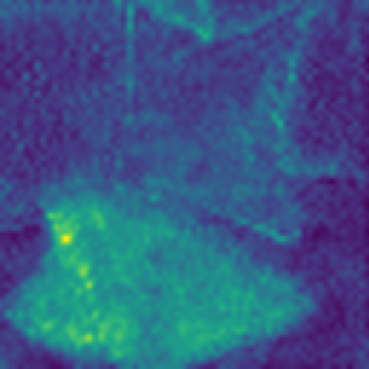} &
\includegraphics[width=2.3cm]{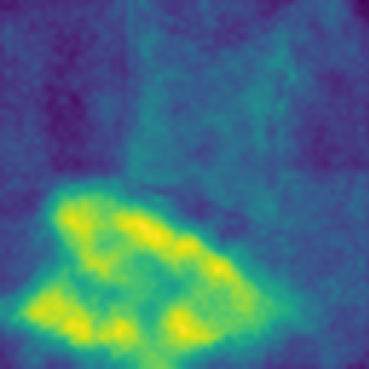} &
\includegraphics[width=2.3cm]{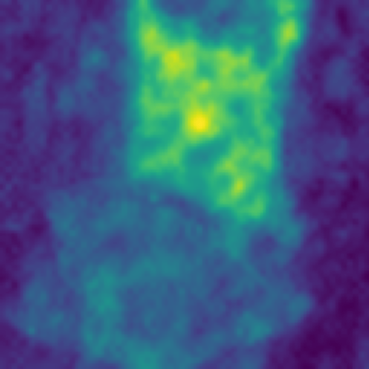} &
\includegraphics[width=2.3cm]{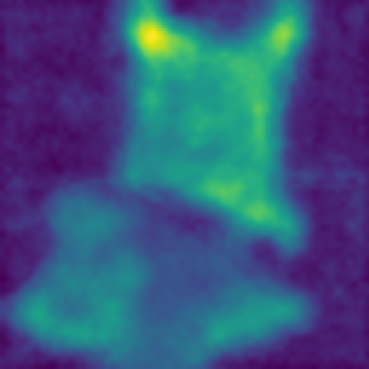} \\ 
\rule[4ex]{0pt}{0pt}
                  & \multicolumn{5}{c}{prompt: ''A black cat sitting in a green bowl''} \\
                  & image    & black         & cat         & green         & bowl         \\
\raisebox{35pt}{Baseline} &
\includegraphics[width=2.3cm]{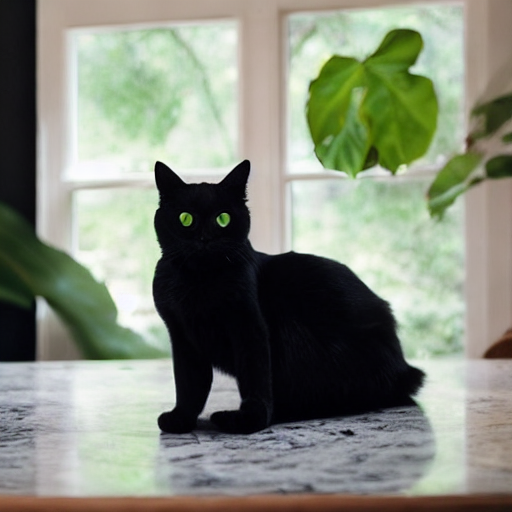} & 
\includegraphics[width=2.3cm]{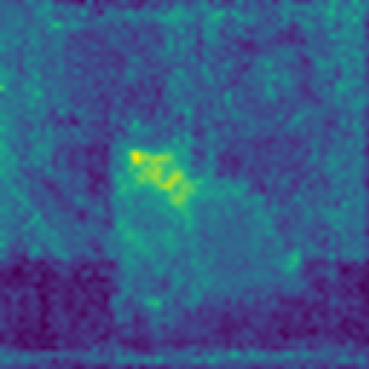} &
\includegraphics[width=2.3cm]{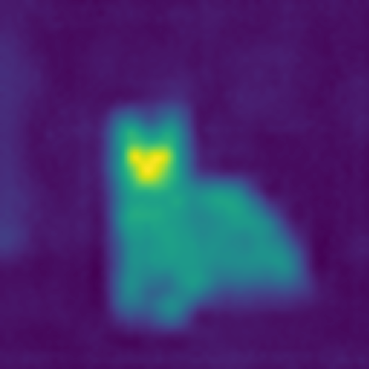} &
\includegraphics[width=2.3cm]{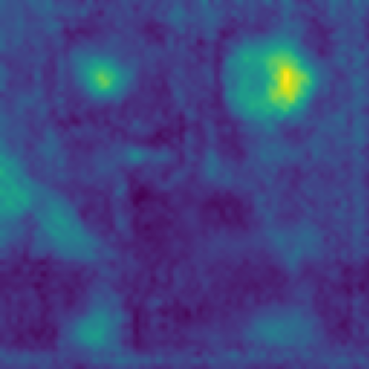} &
\includegraphics[width=2.3cm]{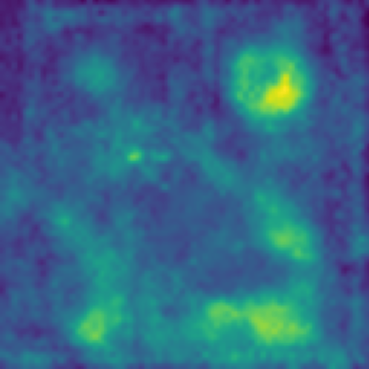} \\
\raisebox{35pt}{\makecell{Linear \\ Projection}} &
\includegraphics[width=2.3cm]{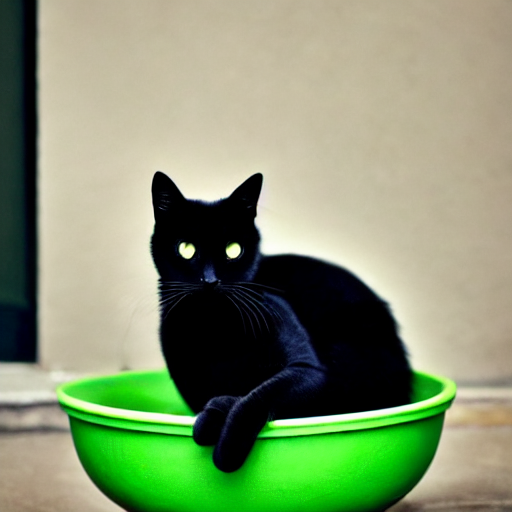} & 
\includegraphics[width=2.3cm]{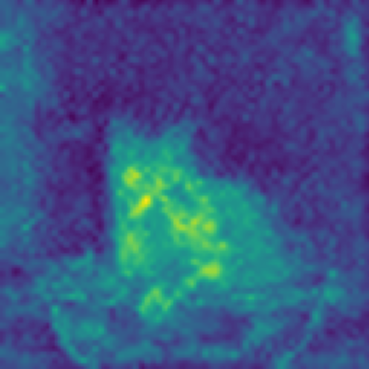} &
\includegraphics[width=2.3cm]{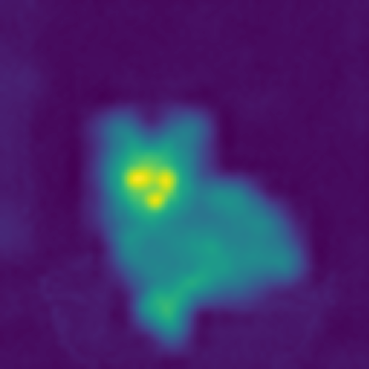} &
\includegraphics[width=2.3cm]{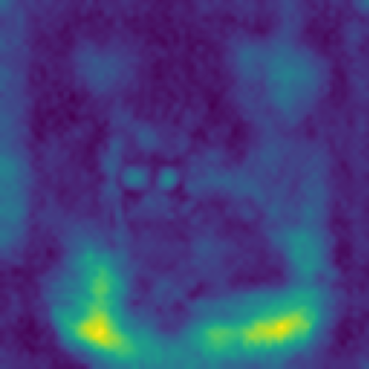} &
\includegraphics[width=2.3cm]{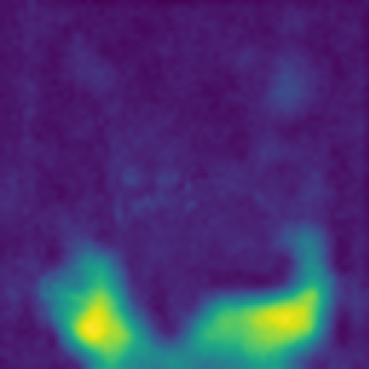} \\
\rule[4ex]{0pt}{0pt}
                  & \multicolumn{5}{c}{prompt: ''A bathroom has brown wall and gold counters''} \\
                  & image    & brown         & wall         & gold         & counters         \\
\raisebox{35pt}{Baseline} &
\includegraphics[width=2.3cm]{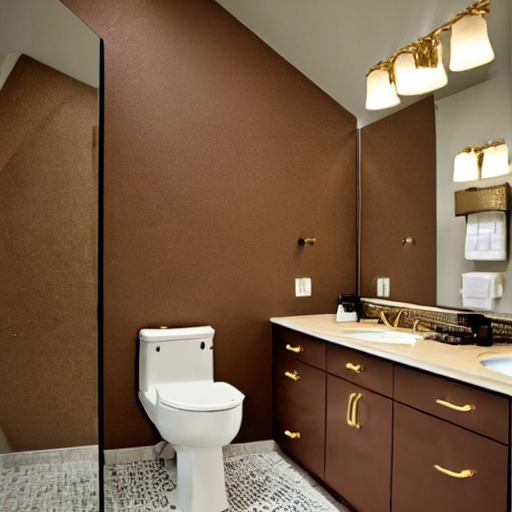} & 
\includegraphics[width=2.3cm]{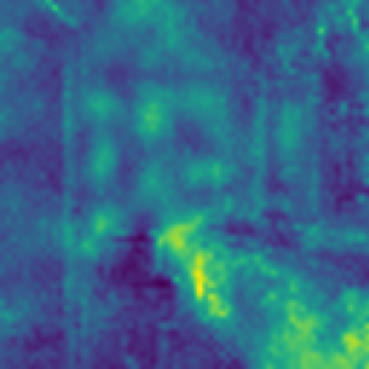} &
\includegraphics[width=2.3cm]{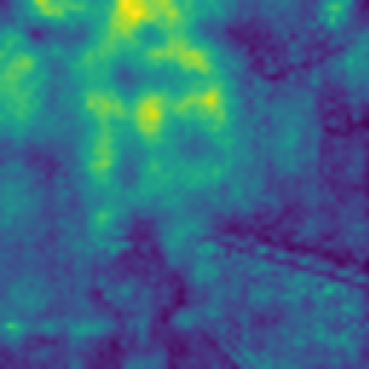} &
\includegraphics[width=2.3cm]{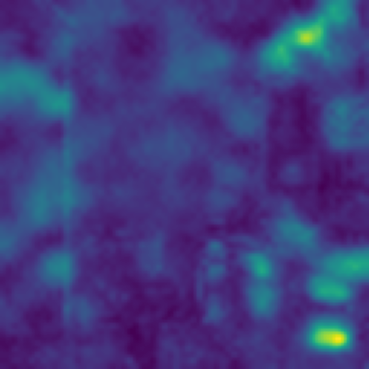} &
\includegraphics[width=2.3cm]{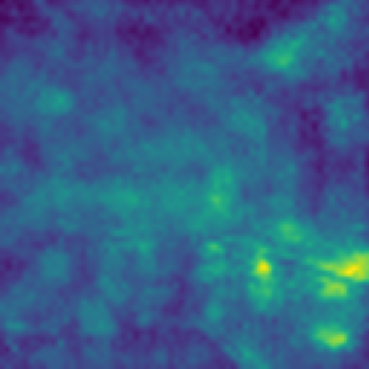} \\
\raisebox{35pt}{\makecell{Linear \\ Projection}} &
\includegraphics[width=2.3cm]{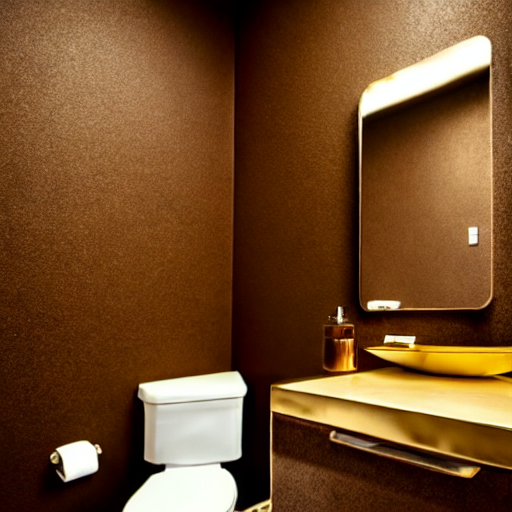} & 
\includegraphics[width=2.3cm]{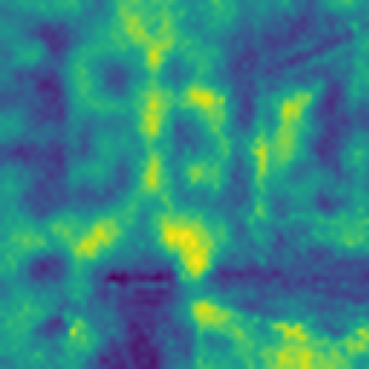} &
\includegraphics[width=2.3cm]{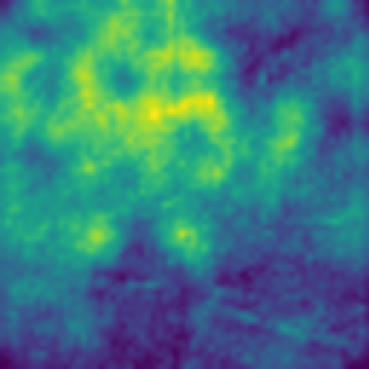} &
\includegraphics[width=2.3cm]{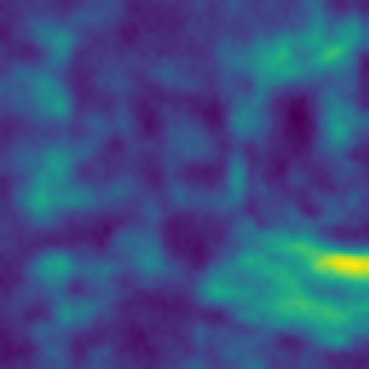} &
\includegraphics[width=2.3cm]{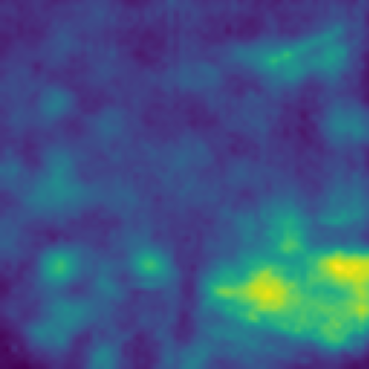} \\
\end{tabular}

\vspace{10pt}

\caption{Comparison of cross-attention maps of the U-Net with and without the \lp}
\label{fig:attention_visualization_of_linear_projection_appendix_2}
\end{figure*}

In this section, we provide additional cross-attention map visualizations in Figures \ref{fig:attention_visualization_of_linear_projection_appendix_1} and \ref{fig:attention_visualization_of_linear_projection_appendix_2}.

\subsection{Visualization of \earlystop{}}
\label{app:sec:early-guidance}

\begin{figure*}[]
\centering
\setlength{\tabcolsep}{1pt}
\begin{tabular}{c@{\hskip 10pt}ccccc}
\multicolumn{6}{c}{prompt: ''A red book and a yellow vase''} \\
\includegraphics[width=2.2cm]{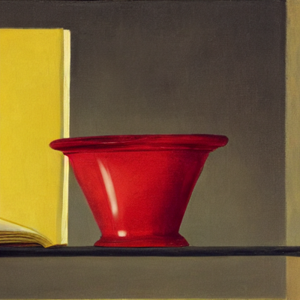} & 
\includegraphics[width=2.2cm]{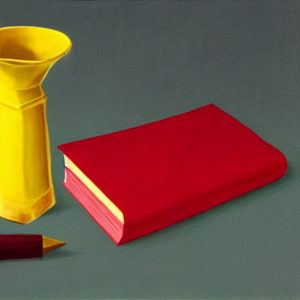} & 
\includegraphics[width=2.2cm]{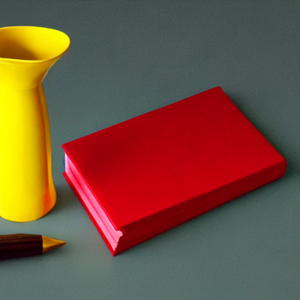} & 
\includegraphics[width=2.2cm]{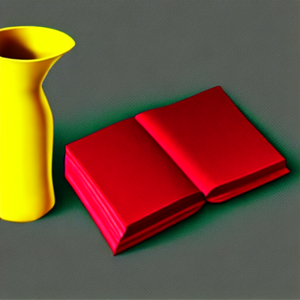} & 
\includegraphics[width=2.2cm]{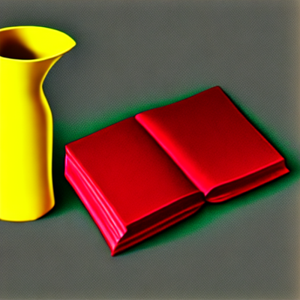} & 
\includegraphics[width=2.2cm]{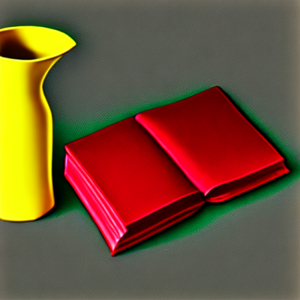} \\
\multicolumn{6}{c}{prompt: ''A bathroom has brown wall and gold counters''} \\
\includegraphics[width=2.2cm]{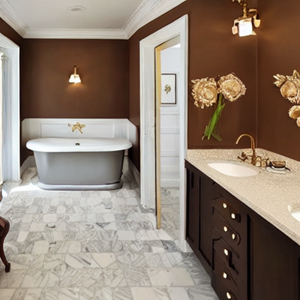} & 
\includegraphics[width=2.2cm]{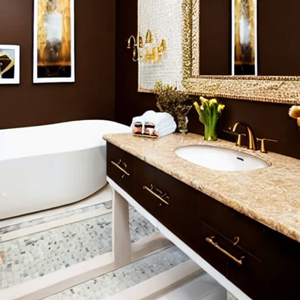} & 
\includegraphics[width=2.2cm]{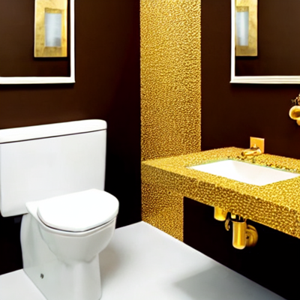} & 
\includegraphics[width=2.2cm]{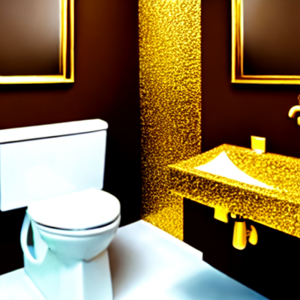} & 
\includegraphics[width=2.2cm]{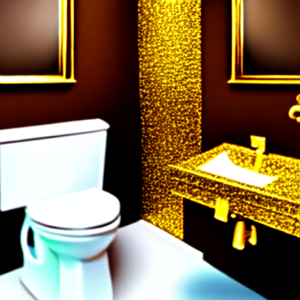} & 
\includegraphics[width=2.2cm]{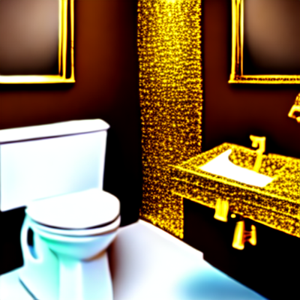} \\
\multicolumn{6}{c}{prompt: ''A blue backpack and a red chair''} \\
\includegraphics[width=2.2cm]{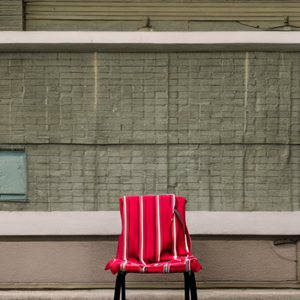} & 
\includegraphics[width=2.2cm]{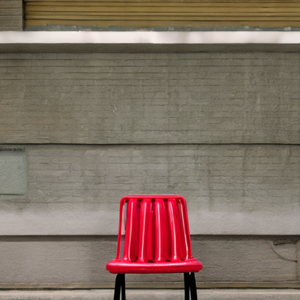} & 
\includegraphics[width=2.2cm]{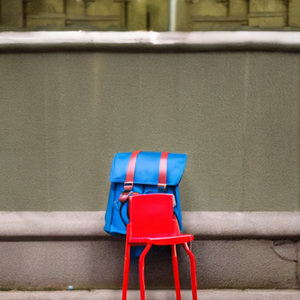} & 
\includegraphics[width=2.2cm]{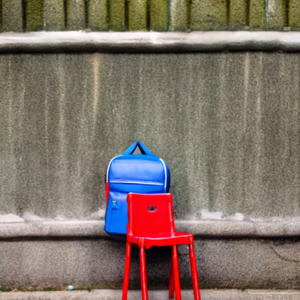} & 
\includegraphics[width=2.2cm]{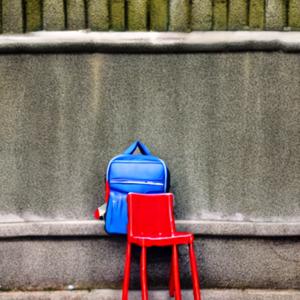} & 
\includegraphics[width=2.2cm]{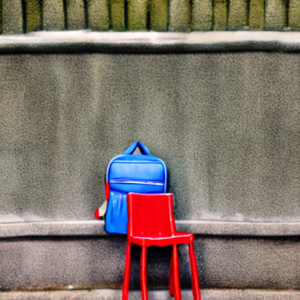} \\
\multicolumn{6}{c}{prompt: ''A blue bear and a brown boat''} \\
\includegraphics[width=2.2cm]{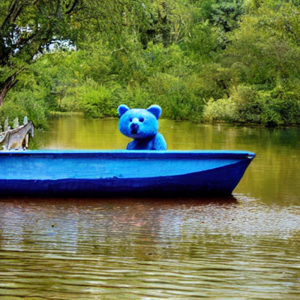} & 
\includegraphics[width=2.2cm]{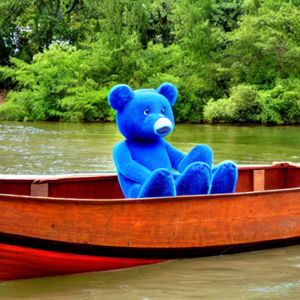} & 
\includegraphics[width=2.2cm]{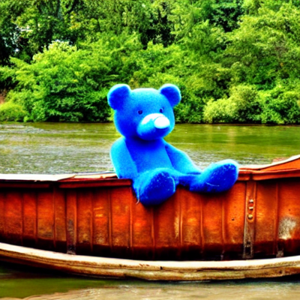} & 
\includegraphics[width=2.2cm]{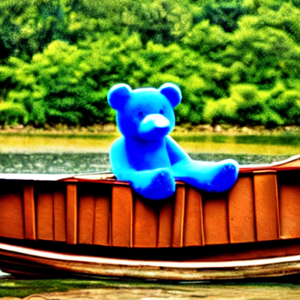} & 
\includegraphics[width=2.2cm]{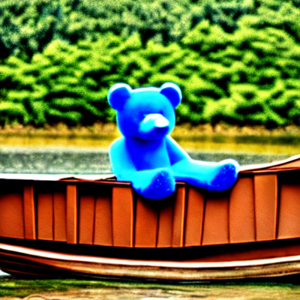} & 
\includegraphics[width=2.2cm]{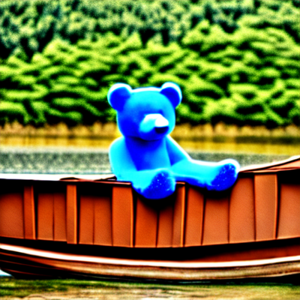} \\
\multicolumn{6}{c}{prompt: ''A brown boat and a blue cat''} \\
\includegraphics[width=2.2cm]{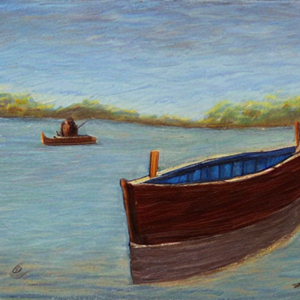} & 
\includegraphics[width=2.2cm]{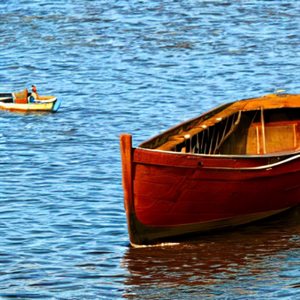} & 
\includegraphics[width=2.2cm]{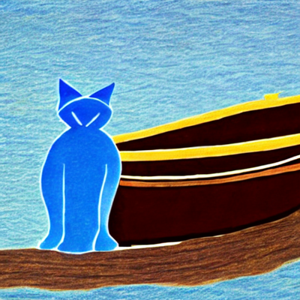} & 
\includegraphics[width=2.2cm]{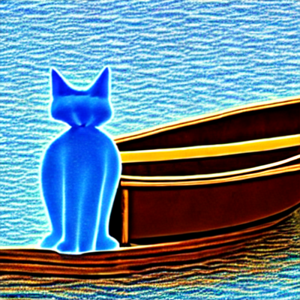} & 
\includegraphics[width=2.2cm]{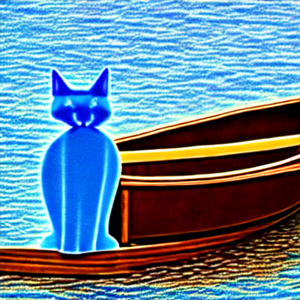} & 
\includegraphics[width=2.2cm]{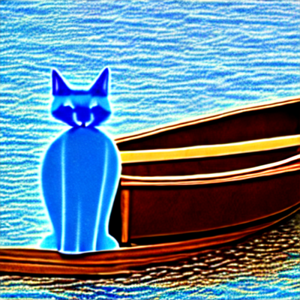} \\
\multicolumn{6}{c}{prompt: ''A green blanket and a blue pillow''} \\
\includegraphics[width=2.2cm]{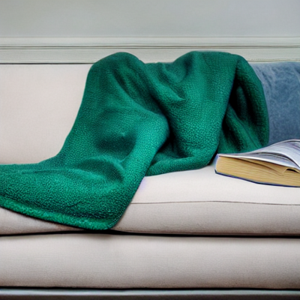} & 
\includegraphics[width=2.2cm]{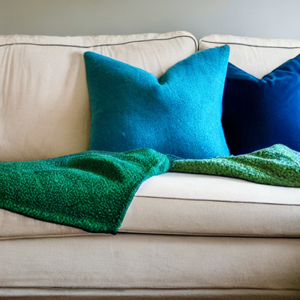} & 
\includegraphics[width=2.2cm]{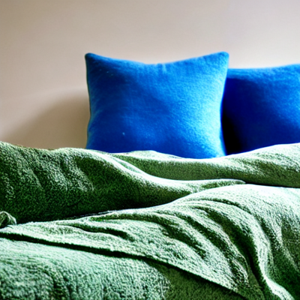} & 
\includegraphics[width=2.2cm]{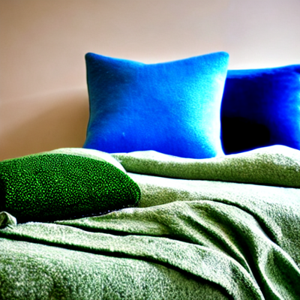} & 
\includegraphics[width=2.2cm]{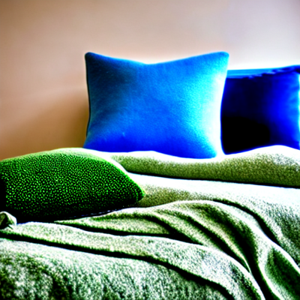} & 
\includegraphics[width=2.2cm]{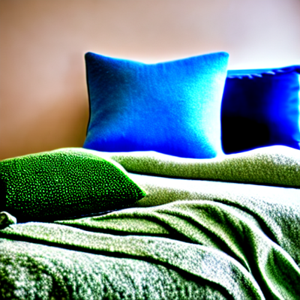} \\
\multicolumn{6}{c}{prompt: ''A green leaf and a yellow butterfly''} \\
\includegraphics[width=2.2cm]{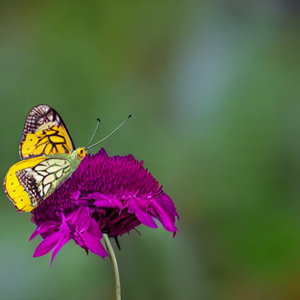} & 
\includegraphics[width=2.2cm]{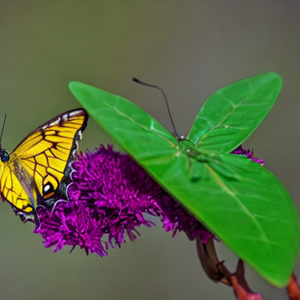} & 
\includegraphics[width=2.2cm]{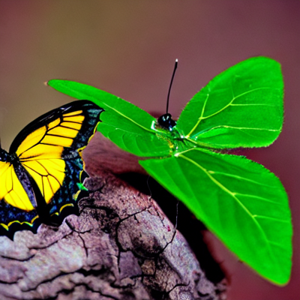} & 
\includegraphics[width=2.2cm]{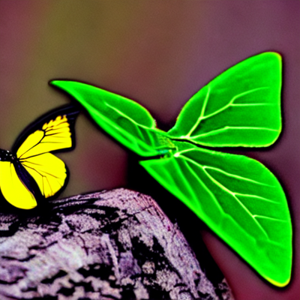} & 
\includegraphics[width=2.2cm]{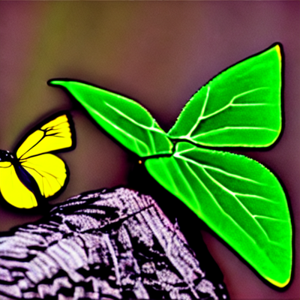} & 
\includegraphics[width=2.2cm]{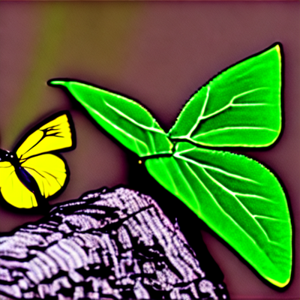} \\
\makecell{$T=1000$ \\ \small{(No Guidance)}} & $T=900$ & $T=800$ & $T=600$ & $T=400$ & $T=200$ \\

\end{tabular}

\vspace{10pt}
\caption{Qualitative results showing the impact of \earlystop{} with varying thresholds $T$}
\label{fig:early_guidance_examples_appendix_1}
\end{figure*}

\begin{figure*}[]
\centering
\setlength{\tabcolsep}{1pt}
\begin{tabular}{c@{\hskip 10pt}ccccc}
\multicolumn{6}{c}{prompt: ''A metallic watch and a fluffy towel''} \\
\includegraphics[width=2.2cm]{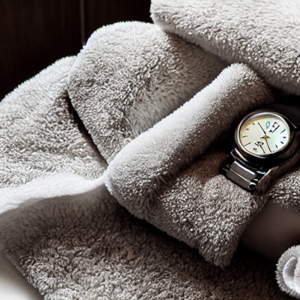} & 
\includegraphics[width=2.2cm]{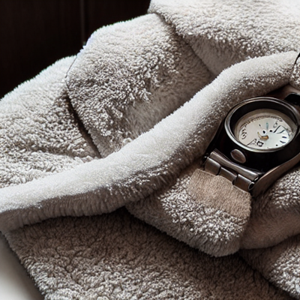} & 
\includegraphics[width=2.2cm]{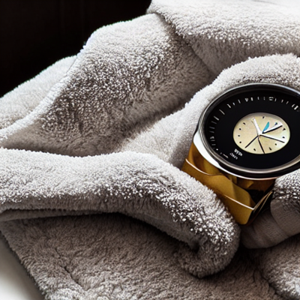} & 
\includegraphics[width=2.2cm]{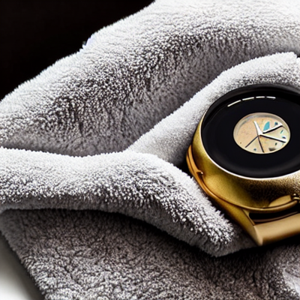} & 
\includegraphics[width=2.2cm]{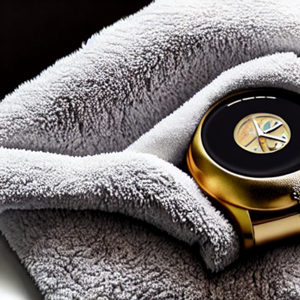} & 
\includegraphics[width=2.2cm]{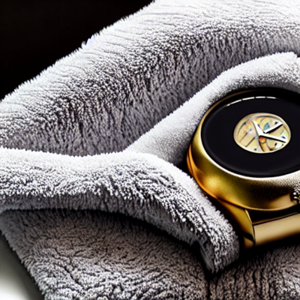} \\
\multicolumn{6}{c}{prompt: ''A pink elephant and a brown giraffe''} \\
\includegraphics[width=2.2cm]{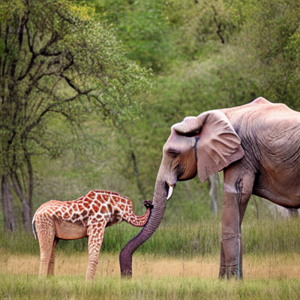} & 
\includegraphics[width=2.2cm]{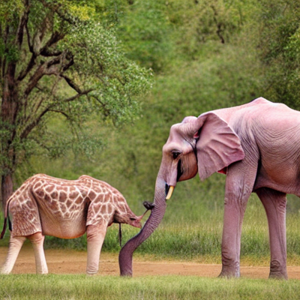} & 
\includegraphics[width=2.2cm]{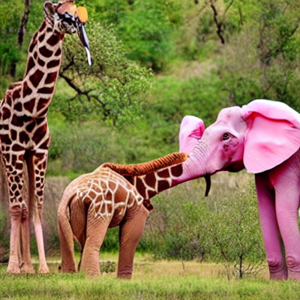} & 
\includegraphics[width=2.2cm]{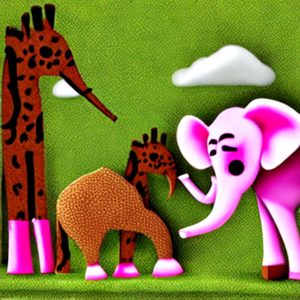} & 
\includegraphics[width=2.2cm]{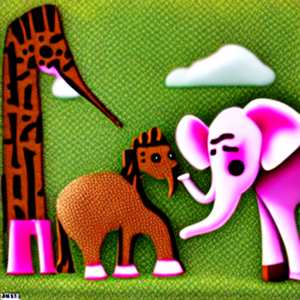} & 
\includegraphics[width=2.2cm]{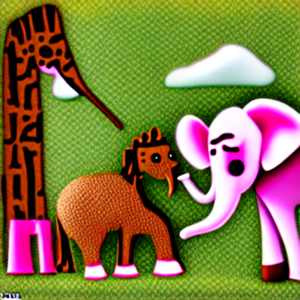} \\
\multicolumn{6}{c}{prompt: ''A plastic bag and a leather chair''} \\
\includegraphics[width=2.2cm]{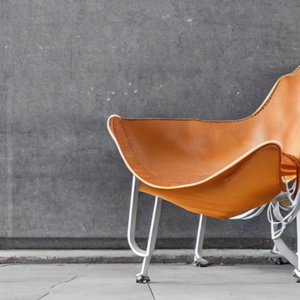} & 
\includegraphics[width=2.2cm]{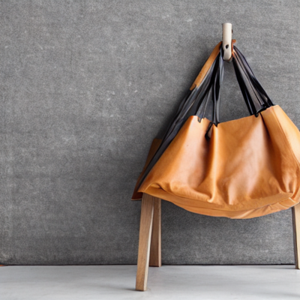} & 
\includegraphics[width=2.2cm]{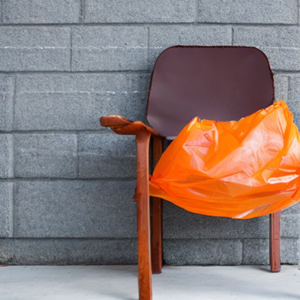} & 
\includegraphics[width=2.2cm]{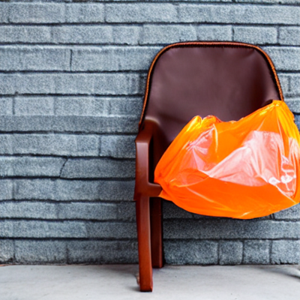} & 
\includegraphics[width=2.2cm]{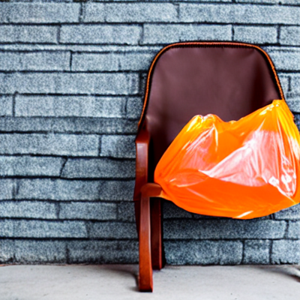} & 
\includegraphics[width=2.2cm]{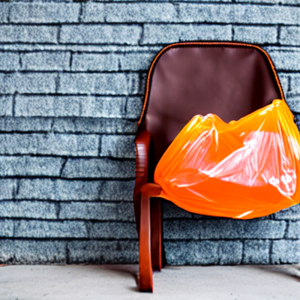} \\
\multicolumn{6}{c}{prompt: ''A red backpack and a blue book''} \\
\includegraphics[width=2.2cm]{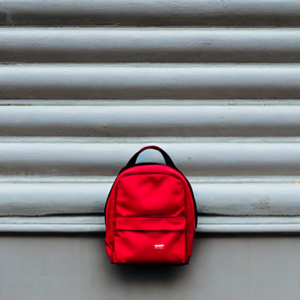} & 
\includegraphics[width=2.2cm]{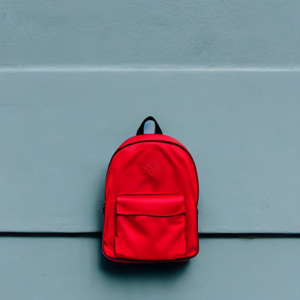} & 
\includegraphics[width=2.2cm]{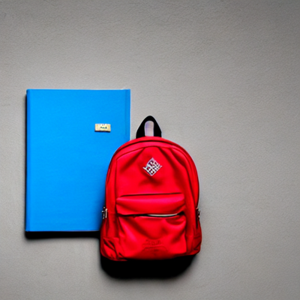} & 
\includegraphics[width=2.2cm]{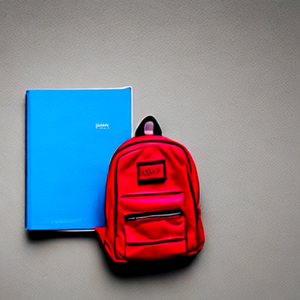} & 
\includegraphics[width=2.2cm]{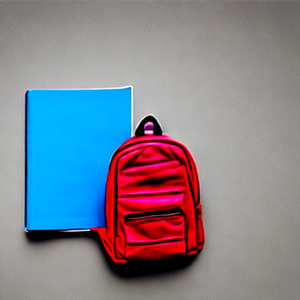} & 
\includegraphics[width=2.2cm]{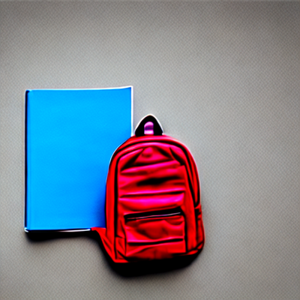} \\
\multicolumn{6}{c}{prompt: ''A red bathroom has a white towel on the bar''} \\
\includegraphics[width=2.2cm]{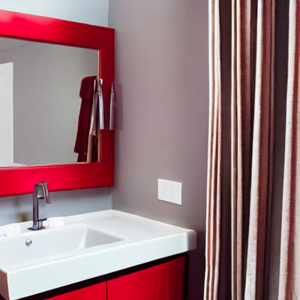} & 
\includegraphics[width=2.2cm]{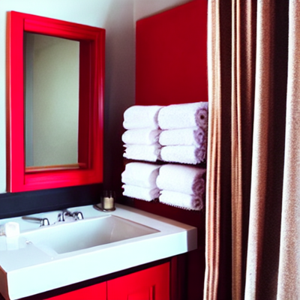} & 
\includegraphics[width=2.2cm]{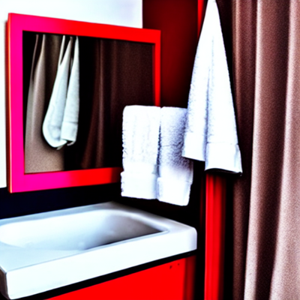} & 
\includegraphics[width=2.2cm]{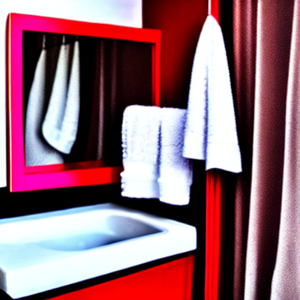} & 
\includegraphics[width=2.2cm]{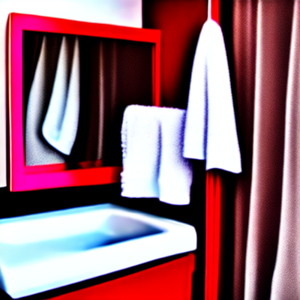} & 
\includegraphics[width=2.2cm]{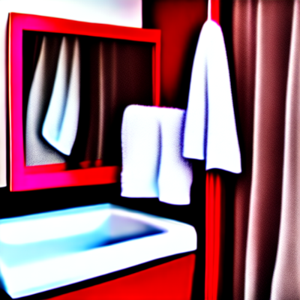} \\
\multicolumn{6}{c}{prompt: ''A red cup and a blue suitcase''} \\
\includegraphics[width=2.2cm]{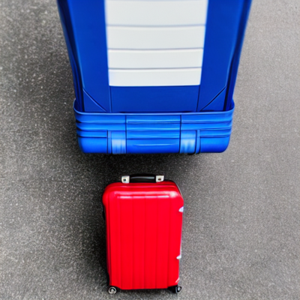} & 
\includegraphics[width=2.2cm]{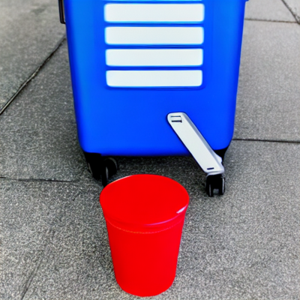} & 
\includegraphics[width=2.2cm]{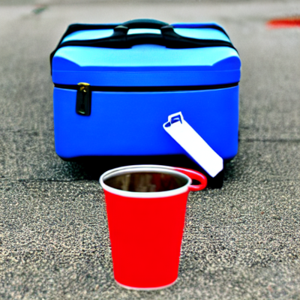} & 
\includegraphics[width=2.2cm]{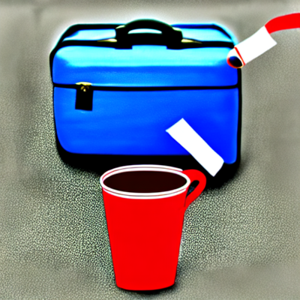} & 
\includegraphics[width=2.2cm]{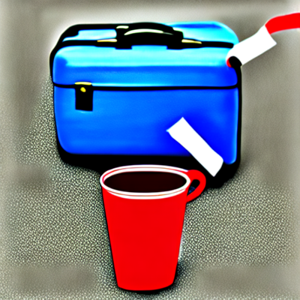} & 
\includegraphics[width=2.2cm]{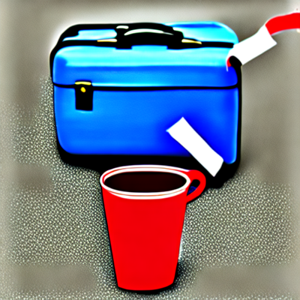} \\
\multicolumn{6}{c}{prompt: ''A white car and a red sheep''} \\
\includegraphics[width=2.2cm]{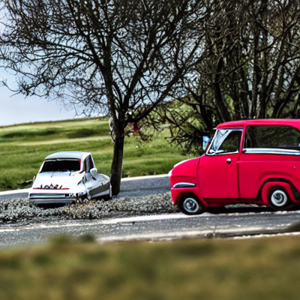} & 
\includegraphics[width=2.2cm]{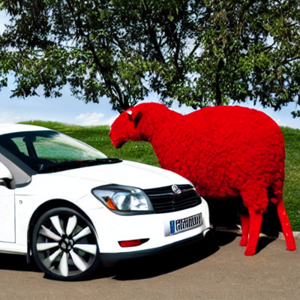} & 
\includegraphics[width=2.2cm]{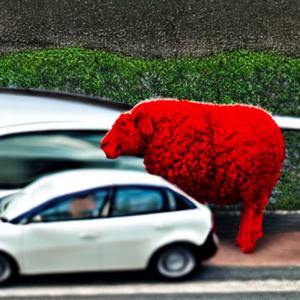} & 
\includegraphics[width=2.2cm]{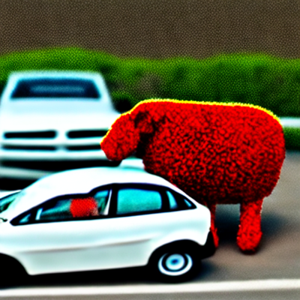} & 
\includegraphics[width=2.2cm]{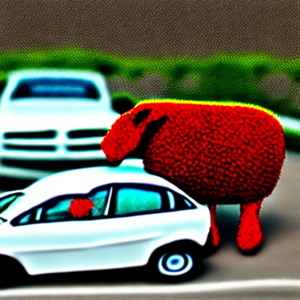} & 
\includegraphics[width=2.2cm]{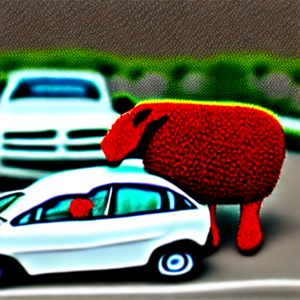} \\
\makecell{$T=1000$ \\ \small{(No Guidance)}} & $T=900$ & $T=800$ & $T=600$ & $T=400$ & $T=200$ \\

\end{tabular}

\vspace{10pt}
\caption{Qualitative results showing the impact of \earlystop{} with varying thresholds $T$}
\label{fig:early_guidance_examples_appendix_2}
\end{figure*}

In this section, we present more qualitative samples illustrating the effect of \earlystop{} at different timestep thresholds for various prompts in Figures \ref{fig:early_guidance_examples_appendix_1} and \ref{fig:early_guidance_examples_appendix_2}.

\subsection{Choice of Window Length in \wb{}}
\label{app:set:window-len}

One might suggest that instead of using token-wise linear projection (\lp{}) or a window-based linear projection with a limited window (\wb{}), employing a linear projection that considers all tokens when finding a better embedding for each token might yield better results. However, our thorough quantitative study and experiments tested various window sizes for \wb{}. We found that using a window size of 5 ($s=2$) achieves the highest performance.

\begin{figure*}[]
\centering
\setlength{\tabcolsep}{1pt}
\begin{tabular}{ccccc}
                  & SD v1.4 & \lp & SD v2 & \wb \\
\raisebox{30pt}{\makecell{A blue bowl and \\ a red train}} &
\includegraphics[width=2.3cm]{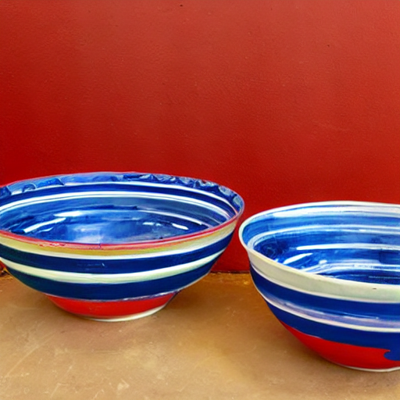} & 
\includegraphics[width=2.3cm]{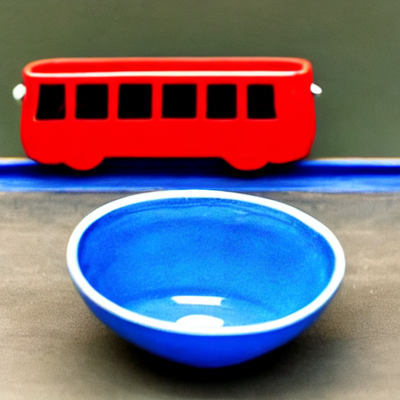} & 
\includegraphics[width=2.3cm]{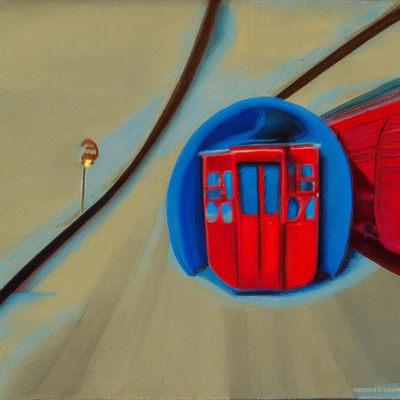} & 
\includegraphics[width=2.3cm]{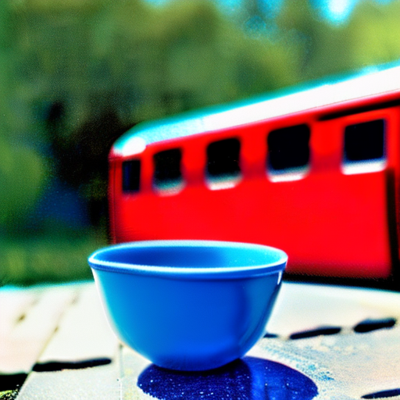} \\
\raisebox{30pt}{\makecell{A blue bench and \\ a green bowl}} &
\includegraphics[width=2.3cm]{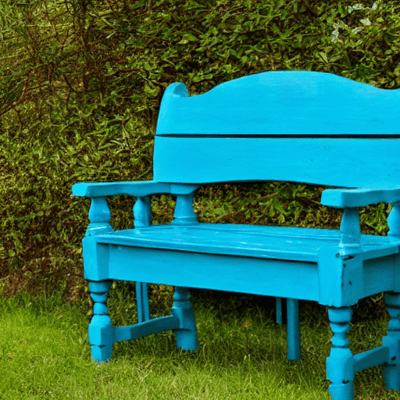} & 
\includegraphics[width=2.3cm]{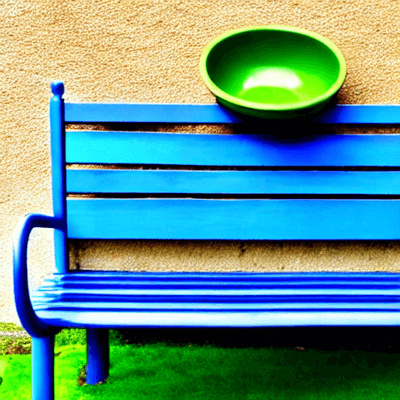} & 
\includegraphics[width=2.3cm]{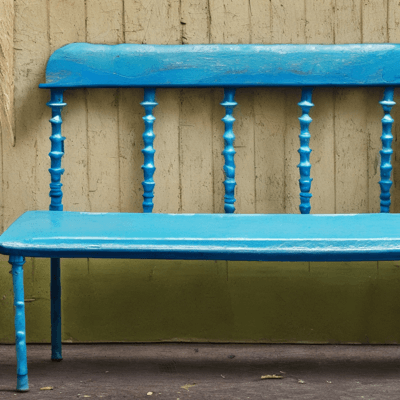} & 
\includegraphics[width=2.3cm]{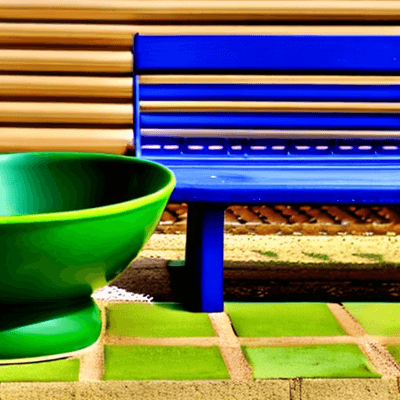} \\
\raisebox{30pt}{\makecell{A blue backpack and \\ a red book}} &
\includegraphics[width=2.3cm]{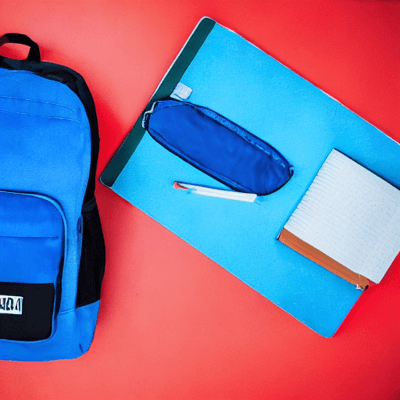} & 
\includegraphics[width=2.3cm]{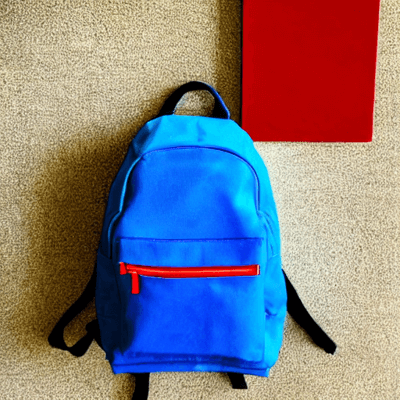} & 
\includegraphics[width=2.3cm]{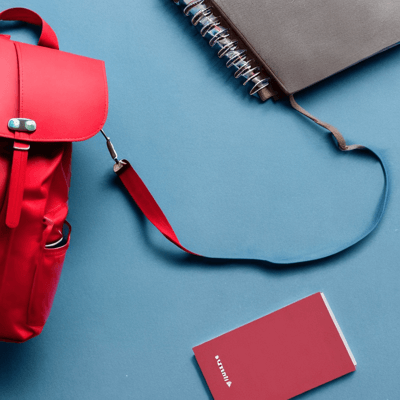} & 
\includegraphics[width=2.3cm]{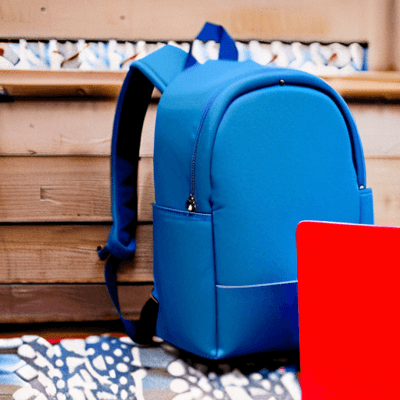} \\
\raisebox{30pt}{\makecell{A black and white cat \\ sitting in a green bowl}} &
\includegraphics[width=2.3cm]{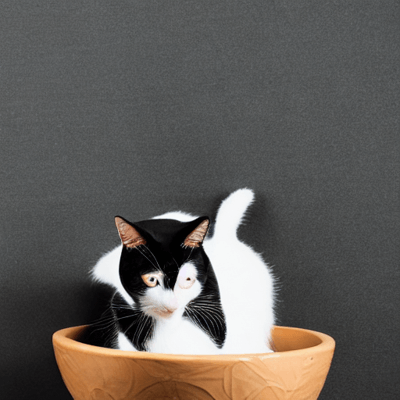} & 
\includegraphics[width=2.3cm]{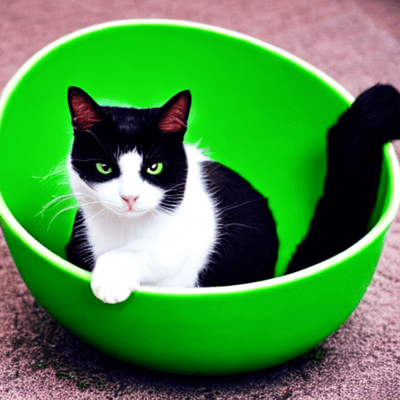} & 
\includegraphics[width=2.3cm]{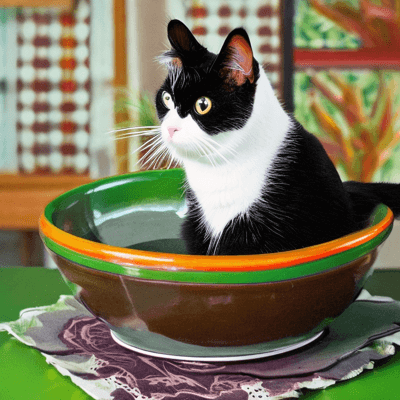} & 
\includegraphics[width=2.3cm]{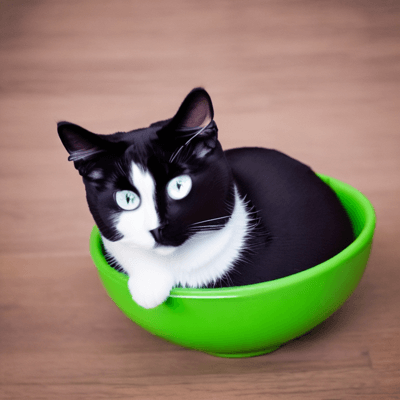} \\
\raisebox{30pt}{\makecell{A brown boat and \\ a blue cat}} &
\includegraphics[width=2.3cm]{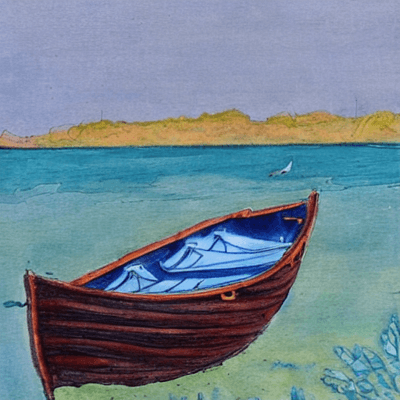} & 
\includegraphics[width=2.3cm]{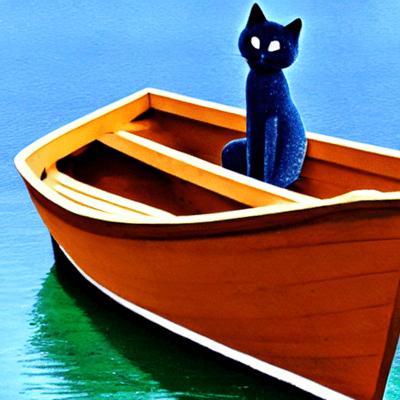} & 
\includegraphics[width=2.3cm]{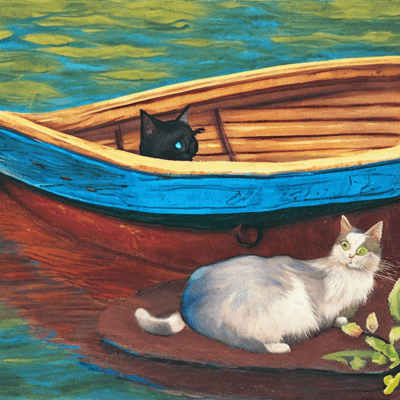} & 
\includegraphics[width=2.3cm]{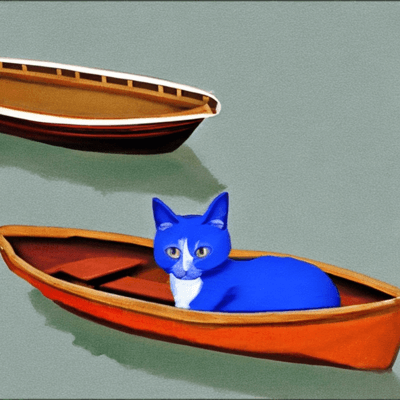} \\
\raisebox{30pt}{\makecell{A brown book and \\ a red sheep}} &
\includegraphics[width=2.3cm]{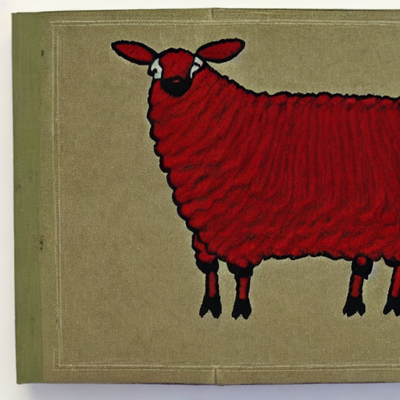} & 
\includegraphics[width=2.3cm]{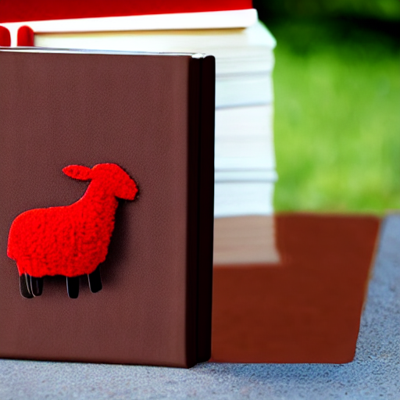} & 
\includegraphics[width=2.3cm]{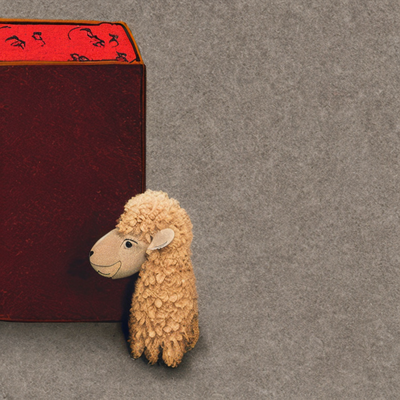} & 
\includegraphics[width=2.3cm]{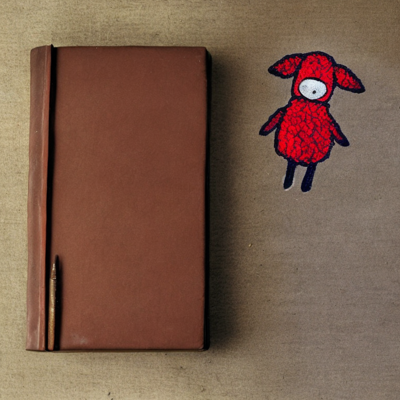} \\
\raisebox{30pt}{\makecell{A fluffy towel and \\ a glass cup}} &
\includegraphics[width=2.3cm]{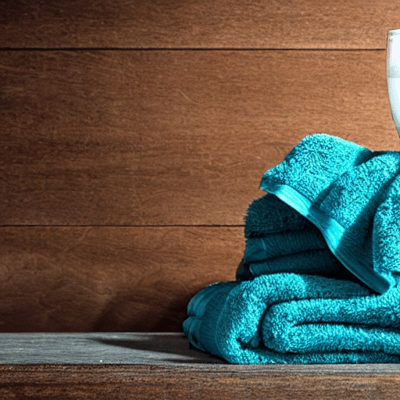} & 
\includegraphics[width=2.3cm]{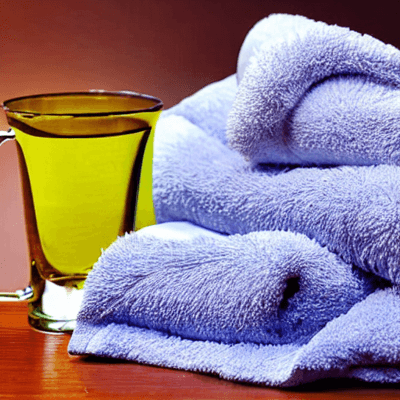} & 
\includegraphics[width=2.3cm]{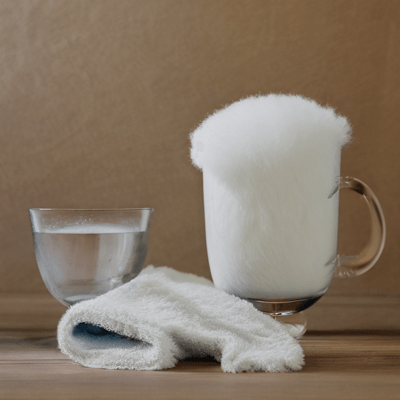} & 
\includegraphics[width=2.3cm]{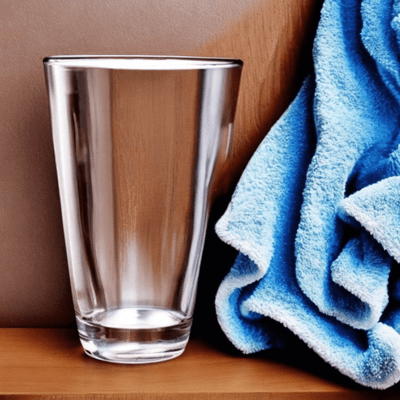} \\
\raisebox{30pt}{\makecell{A plastic container and \\ a fluffy teddy bear}} &
\includegraphics[width=2.3cm]{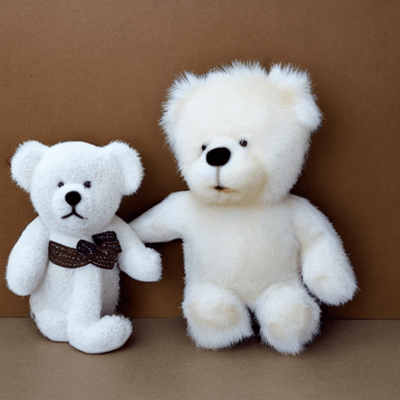} & 
\includegraphics[width=2.3cm]{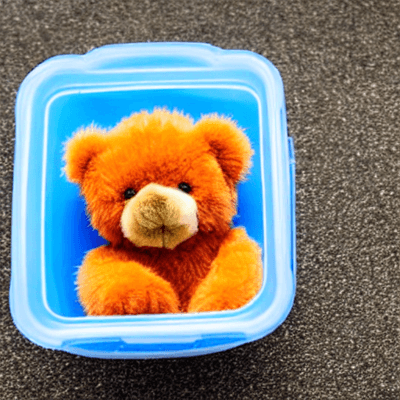} & 
\includegraphics[width=2.3cm]{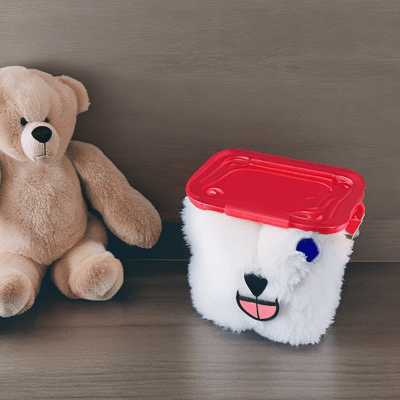} & 
\includegraphics[width=2.3cm]{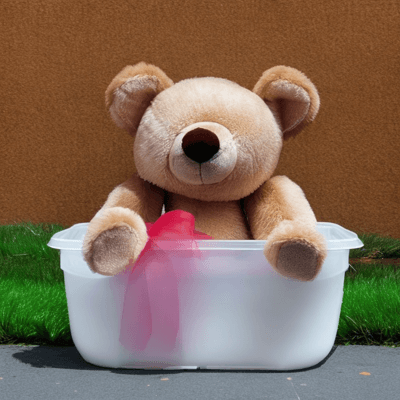} \\
\end{tabular}

\vspace{10pt}

\caption{Qualitative comparison between the baseline and our projection methods (\lp{} and \wb{}).}
\label{fig:prompts_examples_appendix_1}
\end{figure*}

\begin{figure*}[]
\centering
\setlength{\tabcolsep}{1pt}
\begin{tabular}{ccccc}
                  & SD v1.4 & \lp & SD v2 & \wb \\
\raisebox{30pt}{\makecell{A red apple and \\ a green train}} &
\includegraphics[width=2.3cm]{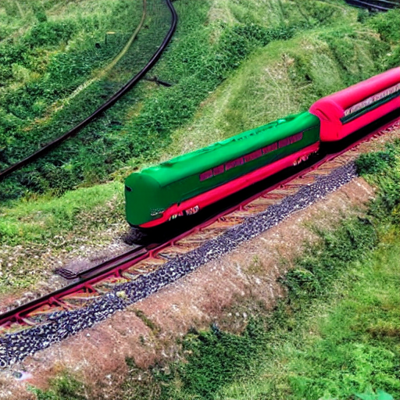} & 
\includegraphics[width=2.3cm]{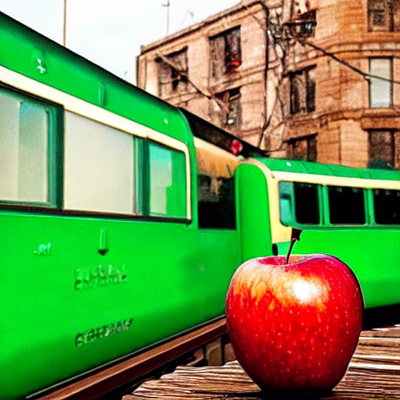} & 
\includegraphics[width=2.3cm]{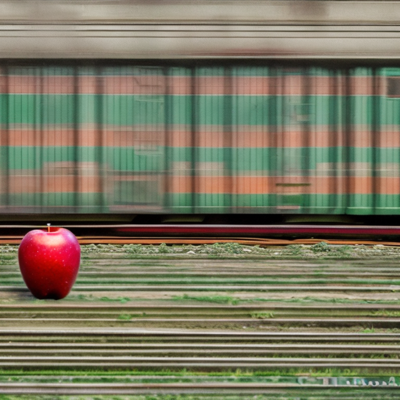} & 
\includegraphics[width=2.3cm]{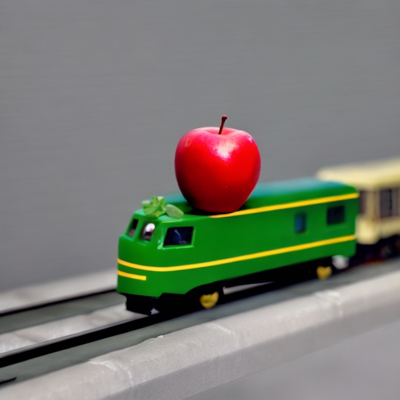} \\
\raisebox{30pt}{\makecell{A red chair and \\ a gold clock}} &
\includegraphics[width=2.3cm]{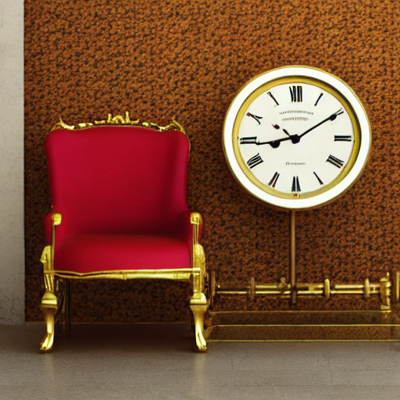} & 
\includegraphics[width=2.3cm]{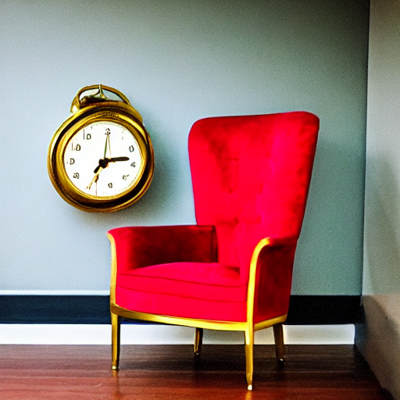} & 
\includegraphics[width=2.3cm]{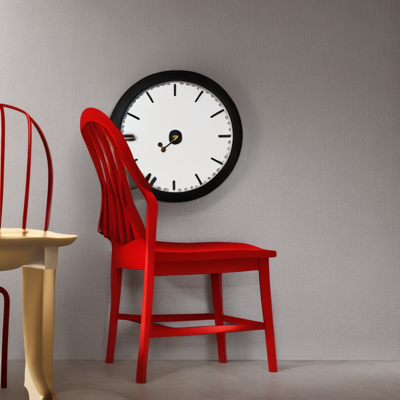} & 
\includegraphics[width=2.3cm]{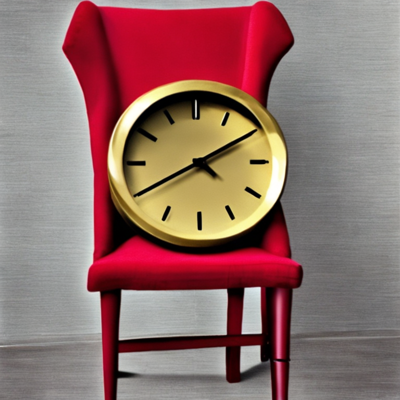} \\
\raisebox{30pt}{\makecell{A red pen and \\ a blue notebook}} &
\includegraphics[width=2.3cm]{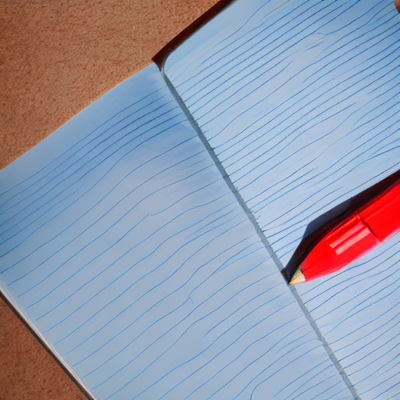} & 
\includegraphics[width=2.3cm]{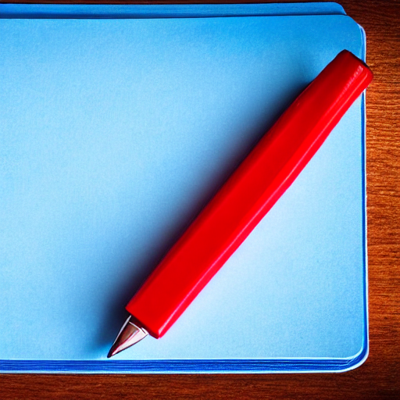} & 
\includegraphics[width=2.3cm]{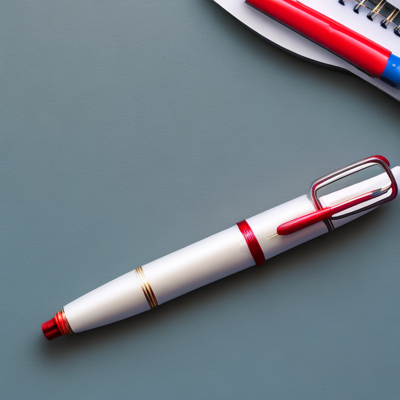} & 
\includegraphics[width=2.3cm]{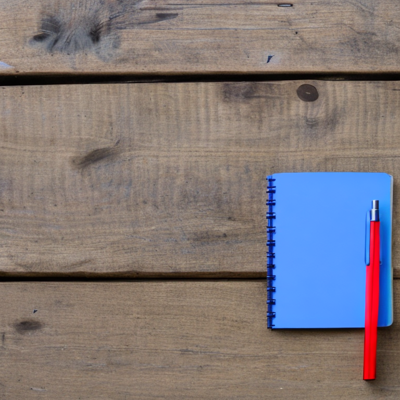} \\
\raisebox{30pt}{\makecell{A round cookie and \\ a square container}} &
\includegraphics[width=2.3cm]{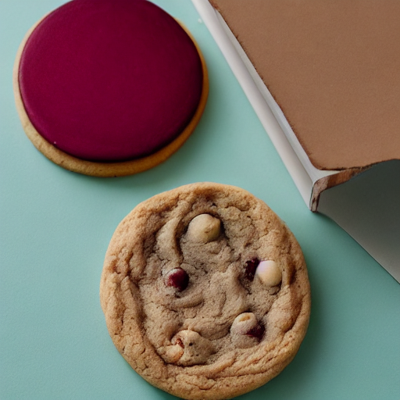} & 
\includegraphics[width=2.3cm]{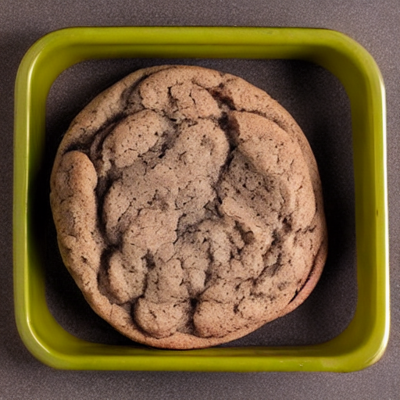} & 
\includegraphics[width=2.3cm]{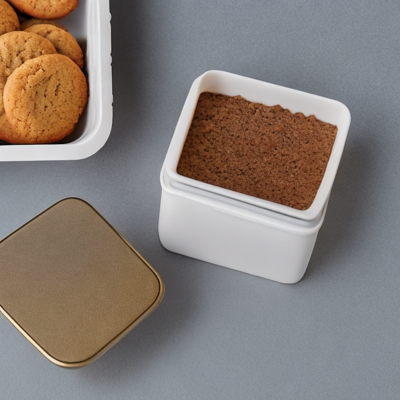} & 
\includegraphics[width=2.3cm]{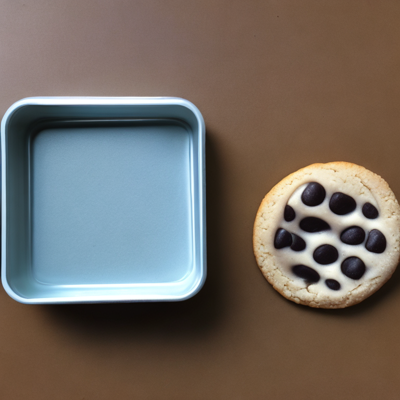} \\
\raisebox{30pt}{\makecell{A wooden floor and \\ a fluffy rug}} &
\includegraphics[width=2.3cm]{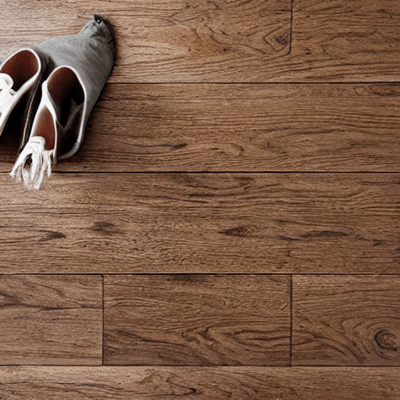} & 
\includegraphics[width=2.3cm]{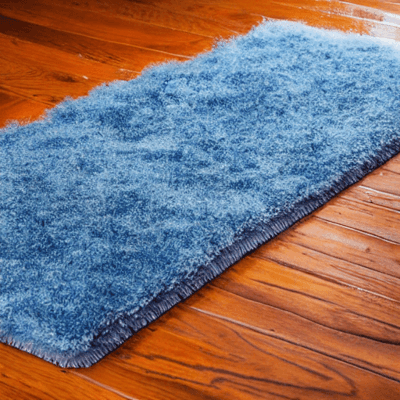} & 
\includegraphics[width=2.3cm]{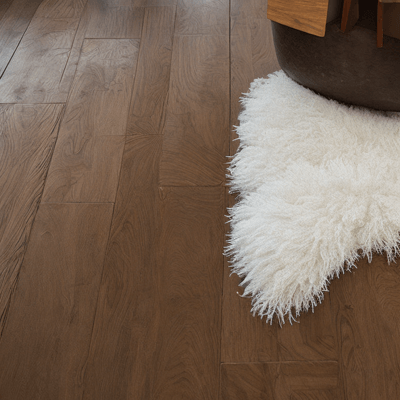} & 
\includegraphics[width=2.3cm]{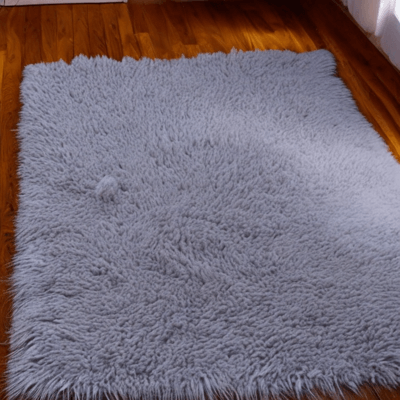} \\
\raisebox{30pt}{\makecell{The leather jacket and fluffy \\ scarf keep the cold at bay}} &
\includegraphics[width=2.3cm]{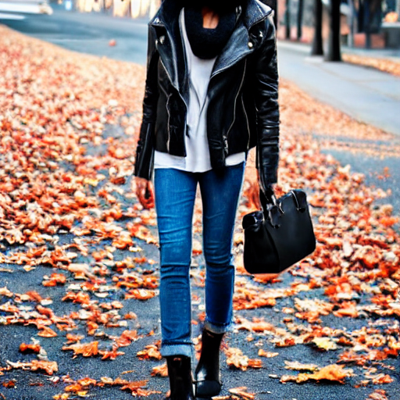} & 
\includegraphics[width=2.3cm]{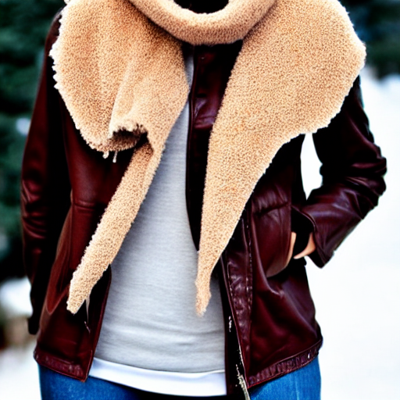} & 
\includegraphics[width=2.3cm]{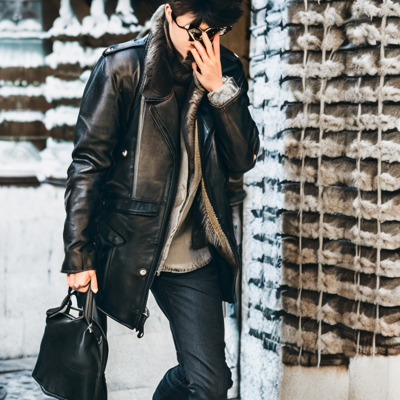} & 
\includegraphics[width=2.3cm]{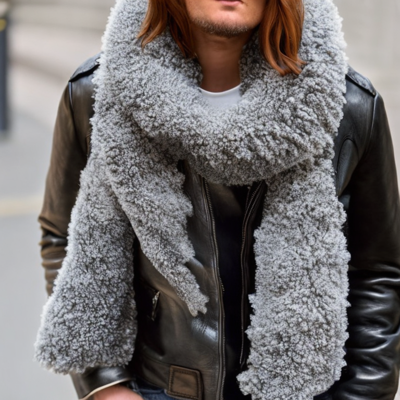} \\
\raisebox{30pt}{\makecell{Wooden pencil and \\ a glass plate}} &
\includegraphics[width=2.3cm]{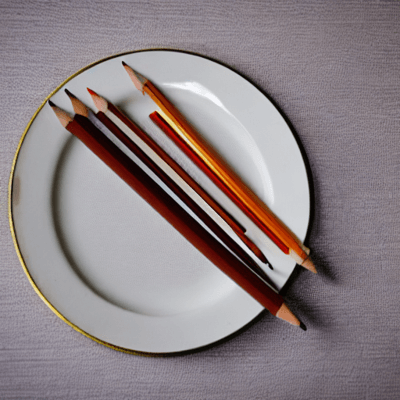} & 
\includegraphics[width=2.3cm]{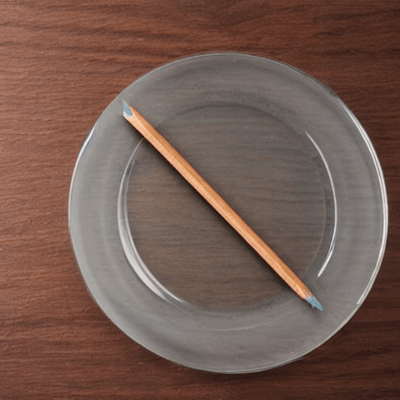} & 
\includegraphics[width=2.3cm]{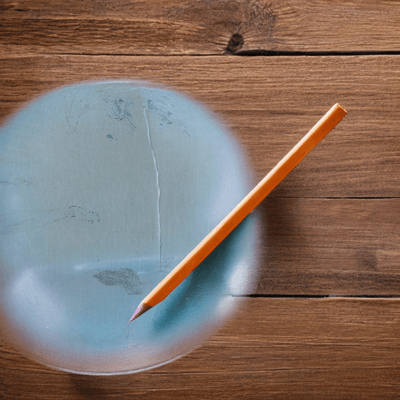} & 
\includegraphics[width=2.3cm]{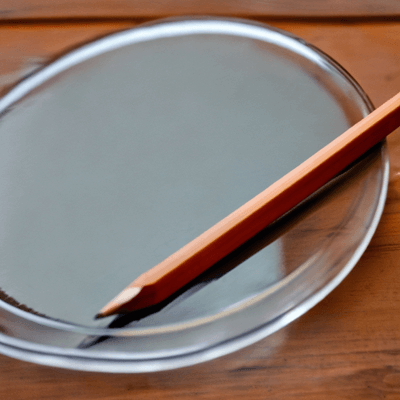} \\
\raisebox{30pt}{\makecell{A green leaf and \\ a yellow butterfly}} &
\includegraphics[width=2.3cm]{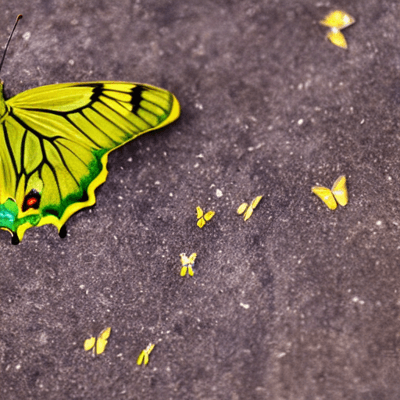} & 
\includegraphics[width=2.3cm]{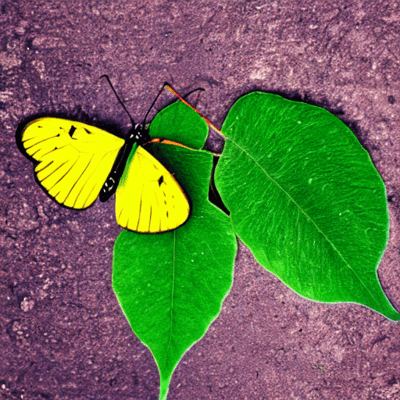} & 
\includegraphics[width=2.3cm]{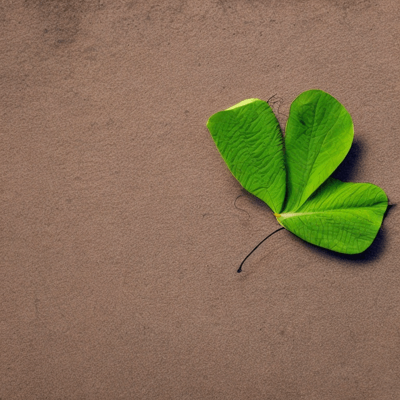} & 
\includegraphics[width=2.3cm]{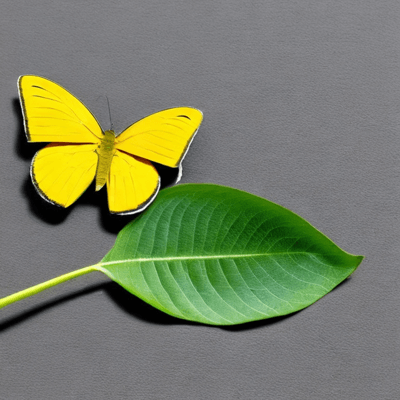} \\
\end{tabular}

\vspace{10pt}

\caption{Qualitative comparison between the baseline and our projection methods (\lp{} and \wb{}).}
\label{fig:prompts_examples_appendix_2}
\end{figure*}

\end{document}